\documentclass[12pt]{article}

\usepackage{times,graphicx,xcolor,amsmath,hyperref}
\usepackage[margin=0.6in]{geometry}
\newenvironment{sciabstract}{%
\begin{quote} \bf}
{\end{quote}}

\title{Climate-Invariant Machine Learning}

\author
{Tom Beucler$^{1,2\ast}$, Pierre Gentine$^{3}$, Janni Yuval$^{4}$, Ankitesh Gupta$^{2}$, \\    
Liran Peng$^{2}$, Jerry Lin$^{2}$, Sungduk Yu$^{2}$, Stephan Rasp$^{5}$, Fiaz Ahmed$^{6}$, \\
Paul A. O'Gorman$^{4}$, J. David Neelin$^{6}$, Nicholas J. Lutsko$^{7}$, Michael Pritchard$^{2,8}$\\
\\
\footnotesize{$^{1}$Faculty of Geosciences and Environment, University of Lausanne, Lausanne, VD 1015, Switzerland}\\
\footnotesize{$^{2}$Department of Earth System Science, University of California, Irvine, CA 92697, USA}\\
\footnotesize{$^{3}$Department of Earth and Environmental Engineering, Columbia University, New York, NY 10027, USA}\\
\footnotesize{$^{4}$Department of Earth, Atmospheric, and Planetary Sciences, Massachusetts Institute of Technology,}\\   
\footnotesize{Cambridge, MA 02139, USA}\\    
\footnotesize{$^{5}$Google Research, Mountain View, CA 94043, USA}\\   
\footnotesize{$^{6}$Department of Atmospheric and Oceanic Sciences, University of California, Los Angeles,} \\    
\footnotesize{Los Angeles, CA 90095, USA}\\
\footnotesize{$^{7}$Scripps Institution of Oceanography, University of California, San Diego, La Jolla, CA 92037, USA}\\
\footnotesize{$^{8}$NVIDIA, Santa Clara, CA 95050, USA}\\
\footnotesize{$^\ast$To whom correspondence should be addressed; E-mail:  tom.beucler@gmail.com.}
}

\date{}

\begin{document}



\maketitle



\begin{sciabstract}
Projecting climate change is a generalization problem: we extrapolate the recent past using physical models across past, present, and future climates. Current climate models require representations of processes that occur at scales smaller than model grid size, which have been the main source of model projection uncertainty. Recent machine learning (ML) algorithms hold promise to improve such process representations, but tend to extrapolate poorly to climate regimes they were not trained on. To get the best of the physical and statistical worlds, we propose a new framework — termed ``climate-invariant'' ML —  incorporating knowledge of climate processes into ML algorithms, and show that it can maintain high offline accuracy across a wide range of climate conditions and configurations in three distinct atmospheric models. Our results suggest that explicitly incorporating physical knowledge into data-driven models of Earth system processes can improve their consistency, data efficiency, and generalizability across climate regimes.
\end{sciabstract}

Teaser: \textit{Physically-informed transformations aid the machine learning of Earth system model processes that generalize across climates.} 

\twocolumn

\section{Introduction}

\subsection{Background}

Following its success in computer vision and natural language processing, machine learning (ML) is rapidly percolating through climate science (e.g., reviews by \cite{reichstein2019deep,beucler2021machine,irrgang2021towards,de2023machine,molina2023review}). We use the term ML here to broadly describe algorithms that learn a task from data without being explicitly programmed for that task. Applications of ML in atmospheric science include the emulation of radiative transfer algorithms (e.g., \cite{ukkonen2020accelerating,belochitski2011tree,gristey2020relationship,lagerquist2021using}), momentum fluxes (e.g., \cite{chantry2021machine,espinosa2022machine,matsuoka2020application,yuval2021momentum}) and microphysical schemes (e.g., \cite{morrison2020bayesian,seifert2020potential,gettelman2021machine}), the bias correction of climate predictions (e.g., \cite{franccois2021adjusting,pan2021learning}), the detection and classification of clouds and storms (e.g., \cite{zantedeschi2019cumulo,rasp2020combining,watson2020large,denby2020discovering}), and the development of subgrid-scale ``closures" (i.e. representation based on coarse-scale processes only) from high-resolution simulation data 
(e.g., \cite{han2020moist,brenowitz2018prognostic,krasnopolsky2013using}), which is the main application discussed in this manuscript.  





ML algorithms typically optimize an objective on a training dataset and make implicit assumptions when extrapolating. Here, extrapolation refers to predictions outside of the training data range, henceforth referred to as \textit{out-of-distribution} predictions. As an example, multiple linear regressions (MLR) assume that the linear relationship that best describes the training set is valid outside of that training set. Alternatively, when confronted with out-of-distribution inputs, random forests (RF, \cite{breiman2001random}) find the closest inputs in their training sets and assign the corresponding outputs regardless of the out-of-distribution input values. Neural networks (NN), which are powerful nonlinear regression and classification tools, rely on nonlinear activation functions and fitted weights to extrapolate. Except in specific situations (e.g., samples in the close neighborhood of the training set or described by the same nonlinear mapping as the training set), there is no reason why NNs should generalize well far outside of their training sets. We show later that different NN training approaches on the same data can lead to drastically different out-of-distribution predictions, highlighting the uncertainty associated with such predictions.

In climate applications, this extrapolation issue means that ML algorithms typically fail when exposed to dynamic, thermodynamic, or radiative conditions that differ substantially from the range of conditions they were trained on. Examples include \cite{o2018using}, who showed that an RF-based moist convection scheme generalizes poorly in the tropics of a climate 6.5K warmer than the training climate; and \cite{hernanz2022evaluation}, who showed that NNs and support vector machines downscaling surface air temperature made substantial extrapolation errors when exposed to temperatures 2-3K warmer than in the training set. \cite{rasp2018deep} showed that an NN-based thermodynamic subgrid-scale closure generalizes well to climates 1-2K warmer than the training one, but makes large errors as soon as the test climate is 4K warmer than the training one. \cite{beucler2020towards} confirmed that these generalization errors remain even when the NN subgrid closure is modified to enforce conservation laws to within machine precision. This has led several studies to recommend training ML models in multiple climates if possible \cite{clark2022correcting,doury2022regional}. Interestingly, both Guillaumin et al. \cite{guillaumin2021stochastic} who trained an NN parameterization for subgrid oceanic momentum transport and Molina et al. \cite{molina2021benchmark} who trained convolutional NNs \cite{lecun1995convolutional} to classify thunderstorms in high-resolution model outputs found that their ML models generalized well to a warmer climate. While this may be because both models relied heavily on velocity inputs and their gradients, whose distributions changed only slightly when the climate warmed, Molina et al. noted that using two types of ML layers, namely batch normalization (BN, \cite{ioffe2015batch}) followed by dropout (DP, \cite{srivastava2014dropout}), was key to this successful generalization.

DP and BN are two examples of a larger set of methods that help NNs generalize and avoid overfitting, broadly referred to as ``regularizations'' \cite{ying2019overview}. Most empirical regularization methods (e.g., L1 regularization) rely on the parsimony principle, i.e. that simpler models, accurately describing the training set with fewer fitted parameters, are preferable to more complex models and generalize better to unseen conditions. More systematic approaches to regularization have been developed to use ML models in out-of-distribution situations that still require the same inputs/outputs, referred to as domain adaptation (e.g., \cite{wilson2020survey,patel2015visual,tuia2016domain}), a particular case of transfer learning (e.g., \cite{pan2009survey}). While not all domain adaptation approaches (sample-based, statistics-based, ensemble-based, domain-invariant feature learning, domain mapping, etc.) need supervision \cite{kouw2019review}, they usually require at least a few samples in the generalization domain.

Without dismissing existing domain adaptation methods, we here focus on physically-informed methods that do \textit{not} require samples in the generalization domain for three reasons: (1) one of the climate science community's long-term goals is to train ML models that rely on historical observations only as we cannot, by definition, observe the future climate; (2) as shown later, even if we have access to simulation data across climates, ML models that intrinsically generalize to climates they have not been trained on tend to be more data-efficient and robust to other changes (e.g., configuration changes); and (3) physically-informed methods can be readily combined with existing domain adaptation and regularization methods. Motivated by these challenges, we ask:


How can we enhance ML algorithms with physical knowledge to make accurate predictions in climate conditions that – in standard variables – lie far outside of the training set?

\subsection{Problem Definition}

\begin{figure*}
\centerline{\includegraphics[width=\textwidth]{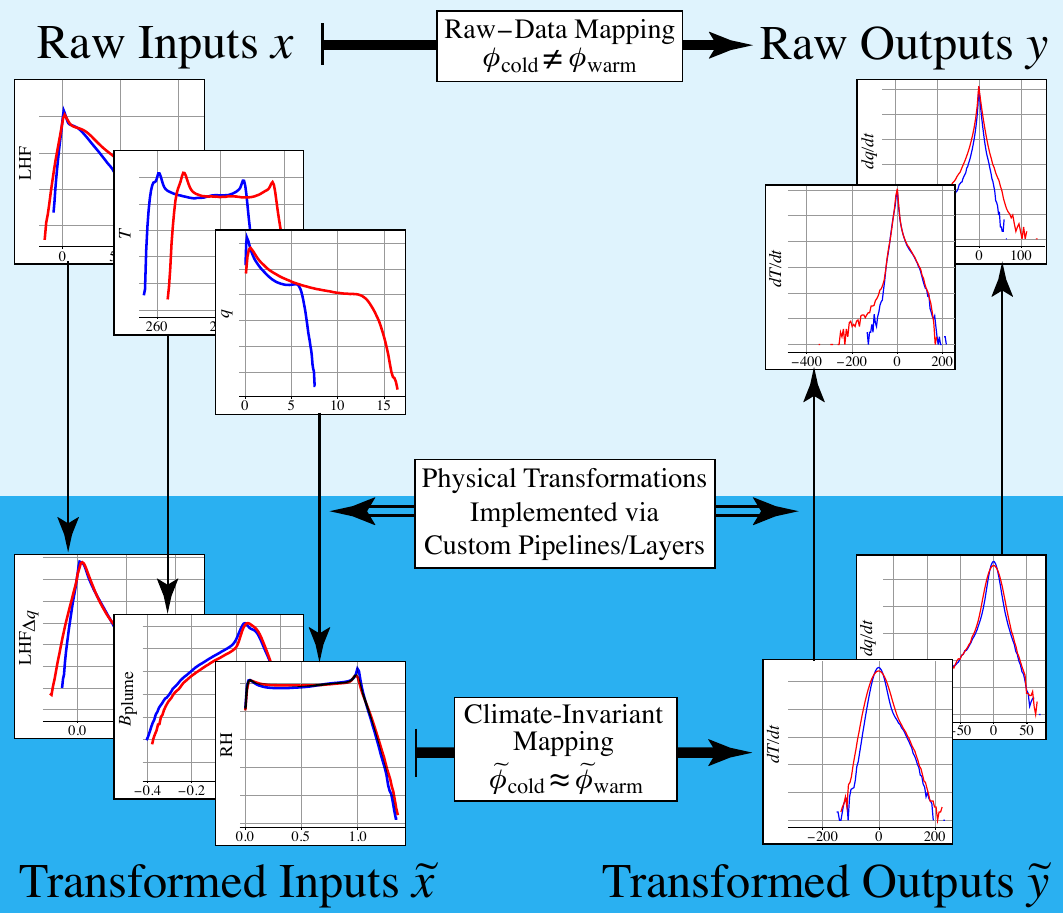}}
 \caption{\textbf{By transforming inputs $\boldsymbol{x} $ and outputs $\boldsymbol{y} $ to match their probability density functions across climates, the algorithms can learn a transformed mapping $\widetilde{\phi}$ that holds across climates.} To illustrate this, we show the marginal distributions of inputs and outputs in two different climates using blue and red lines, before (top) and after (bottom) the physical transformation.
  }\label{f1}
\end{figure*}

Our scientific contribution is to transform a mapping constructed using the original data's features, henceforth referred to as ``raw-data'' mapping, into a mapping that remains nearly constant across climates, here referred to as ``climate-invariant''. Inspired by invariants in physics and self-similarity in fluid mechanics \cite{ling2016machine}, we make the ML-emulated mapping climate-invariant by transforming the input and output vectors so that their distributions shift minimally across different climates (see Fig~\ref{f1}). 

We demonstrate this framework's utility by adapting ML closures of subgrid atmospheric thermodynamics (i.e., coarse-scale thermodynamic tendencies resulting from subgrid convection, radiation, gravity waves, and turbulence) so that they generalize better across climates. The motivation for this application is two-fold. First, purely physically-based \textit{subgrid closures} remain one of the largest sources of uncertainties in Earth system models \cite{randall2003,Bony2015, Schneider2017}. While ML-based closures have emerged as a promising alternative to traditional semi-empirical models \cite{gentine2021deep}, they lack robustness \cite{brenowitz2020interpreting,ott2020fortran} and, as discussed earlier, usually fail to generalize across climates \cite{o2018using,rasp2018deep,beucler2020towards}. Second, atmospheric thermodynamic processes are directly affected by global temperature changes, e.g., in response to anthropogenically-forced climate change \cite{forster2021earth}. Therefore, predicting subgrid thermodynamics in a warm climate with an ML model trained in a cold climate leads to very apparent failure modes \cite{beucler2020towards} that we can transparently tackle. 

In mathematical terms, our goal is to build a climate-invariant mapping between the input vector $\boldsymbol{x} $ representing the large-scale ($\approx$100km) climate state and the output vector $\boldsymbol{y} $ grouping large-scale thermodynamic tendencies due to explicitly-resolved convection and parameterized radiative transfer and turbulent mixing at the $\sim $1km-scale (see SM B1 for details). We keep the overall structure of the mapping $\boldsymbol{x}\mapsto \boldsymbol{y}$ fixed throughout the manuscript and aim to predict $\boldsymbol{y} $ as accurately as possible in training and generalization climates (out-of-distribution prediction). Note that this mapping makes some implicit assumptions based on successful past work \cite{rasp2018deep,mooers2021assessing}, including locality in horizontal space and time (outputs only depend on inputs in the same atmospheric column at the same time step) and determinism (only one possible output vector for a given input vector). We include cloud radiative effects in all heating terms (total heating $\boldsymbol{\dot{T}} $, longwave heating $\boldsymbol{\mathrm{lw}} $, and shortwave heating $\boldsymbol{\mathrm{sw}} $), but for simplicity do not predict changes in cloud liquid water and ice, and exclude cloud water and greenhouse gases other than water vapor $\boldsymbol{q_v} $ from the input vector $\boldsymbol{x}$. 

After introducing the climate simulations and training/validation/test split (Section~\ref{sec:Data}), we define the climate invariant mapping and feature transformations (Section~\ref{sec:Theory}), and demonstrate and explain their ability to generalize (Section~\ref{sec:results}) before concluding. We refer the reader to the Supplementary Materials (SM) for data availability (SM A), additional derivations and descriptions of the mapping and physical transformations (SM B), the implementation of our ML framework (SM C), and additional results (SM D).

\section{Data\label{sec:Data}}

\begin{figure*}
\centering
\centerline{\includegraphics[width=\textwidth]{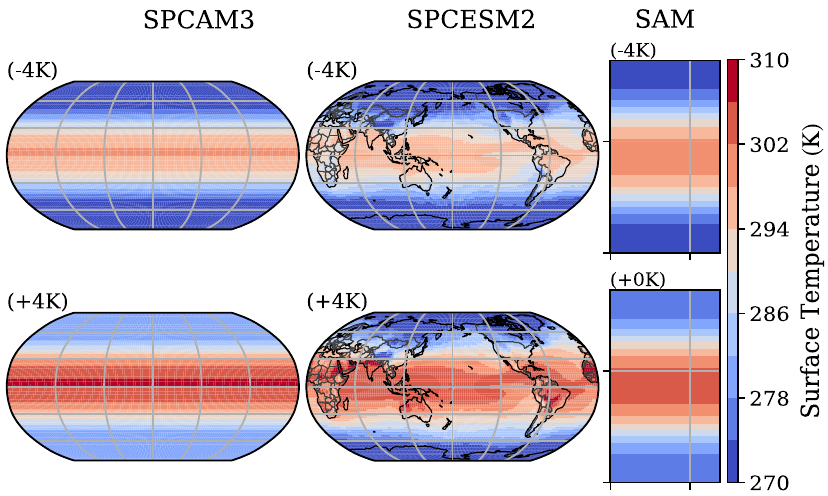}}
\caption{\textbf{Surface temperatures in the three utilized atmospheric models.} Prescribed surface temperature (in K) for (left) the aquaplanet SPCAM3 model and (right) the hypohydrostatic SAM model. (Center) Annual-mean, near-surface air temperatures in the Earth-like SPCESM2 model.\label{f_data}}
\end{figure*}

To test the robustness of our framework across model formulations and configurations, we use three distinct storm-resolving climate models and experimental set-ups: aquaplanet simulations using the Super-Parameterized Community Atmosphere Model version 3.0 (SPCAM3), Earth-like simulations (i.e., with continents) using the Super-Parameterized Community Earth System Model version 2 (SPCESM2), and quasi-global aquaplanet hypohydrostatic simulations using the System for Atmospheric Modeling version 6.3 (SAM). SPCAM3 and SPCESM2 assume a strict scale separation between the resolved coarse scales and subgrid processes, making them ideal testbeds to machine learn local subgrid closures \cite{gentine2018could,rasp2018deep}. In contrast, SAM does not assume scale separation as a global storm-resolving model. This improves realism but requires coarse-graining SAM's output for ML parameterization purposes \cite{yuval2020stable,yuval2021use}. For each climate model, we run three simulations with three different prescribed surface temperature distributions:
\begin{itemize}
    \item (+0K) A reference simulation with a temperature range analogous to the present climate
    \item (-4K) A cold simulation with surface temperatures 4K cooler than the (+0K) simulation
    \item (+4K) A warm simulation with surface temperatures 4K warmer than the (+0K) simulation,
\end{itemize}
with the exception of SAM for which only the (-4K) and (+0K) simulations are available. By prescribing surface temperature, we focus on ML's ability to consistently predict the atmospheric response to climate change across configurations. Projecting climate change involves a broader range of processes and is beyond this work's scope. We summarize the simulations and indicate their spatiotemporal resolutions in Tab~S1. Fig~\ref{f_data} gives a visualization of surface temperatures in each model and Fig~S1 provides snapshots of mid-tropospheric subgrid heating (Fig~\ref{f_data}b), which is one of our ML models' outputs.

\subsection{Super-Parameterized Aquaplanet Simulations\label{sub:SPCAM3}}

We use data from two-year SPCAM3 \cite{khairoutdinov2005simulations} climate simulations in an aquaplanet configuration \cite{pritchard2014causal}, with zonally-symmetric surface temperatures fixed to a realistic meridionally-asymmetric profile \cite{andersen2012moist}. The insolation is fixed to boreal summer conditions with a full diurnal cycle. A two-dimensional storm-resolving model is embedded in each grid cell of SPCAM3, namely 8 SAM atmospheric columns using a spatiotemporal resolution of 4km$\times 30 \mathrm{levels} \times$25s, and the default one-moment microphysical scheme \cite{khairoutdinov2001cloud}. SPCAM3 combines a spectral primitive equation solver with a semi-Lagrangian dynamical core for advection \cite{pritchard2014causal}. The (+0K) SPCAM3 simulation was first presented in \cite{gentine2018could} and subsequently used to train ML subgrid closures in \cite{rasp2018deep,ott2020fortran,beucler2021enforcing}. Inspired by the generalization experiment of \cite{o2018using}, the (+4K) simulation was introduced in \cite{rasp2018deep}, and we ran the (-4K) simulation for the work presented here to increase the surface temperature generalization gap from 4K to 8K.

\subsection{Super-Parameterized Earth-like Simulations\label{sub:SPCESM2}}

We run three two-year SPCESM2 \cite{wang2015multiscale} climate simulations in an Earth-like configuration with realistic surface boundary conditions, including a land surface model, seasonality, aerosol conditions representative of the year 2000, and a zonally-asymmetric annual climatology of sea surface temperatures derived from the ``HadOIB1'' dataset \cite{hurrell2008new}. We use CESM v2.1.3 to couple CAM v4.0 with the Community Land Model version 4.0, and similarly embed 32 SAM columns in each atmospheric grid cell to explicitly represent deep convection. Our (+0K) simulation is similar to \cite{mooers2021assessing}'s, which showed the potential of ML for subgrid closures in Earth-like conditions. 

\subsection{Quasi-Global Aquaplanet Hypohydrostatic Simulations\label{sub:SAM}}

While super-parameterization is well-adapted to statistically learning subgrid closures thanks to its explicit scale separation, this scale separation comes at the cost of distorted mesoscale systems and momentum fluxes \cite{Tulich2015}. Furthermore, most ML subgrid closures are based on coarse-graining high-resolution simulations (e.g., \cite{brenowitz2019spatially,bolton2019applications}). This motivates us to also test the climate-invariant framework in hypohydrostatic SAM simulations in which the dynamics are not affected by a prescribed scale separation. Computational expense is reduced through hypohydrostatic scaling, which multiplies the vertical acceleration in the equations of motion by a factor of 16 to increase the horizontal scale of convection without overly affecting the larger-scale flow \cite{garner2007resolving,boos2016convective}. While these simulations use idealized settings, such as aquaplanet configurations, an anelastic dynamical core, a quasi-global equatorial beta plane domain, and perpetual equinox without a diurnal cycle,  \cite{o2021response} showed that they produce tropical rainfall intensity and cluster-area distributions that are close to satellite observations. The prescribed surface temperature distribution in the control simulation of \cite{o2021response} is designed to be close to zonal-mean observations \cite{neale2000standard}, and its maximum value is roughly 2K colder than that of the distribution used for the (+0K) SPCAM3 simulation. To better match the SPCAM3 maxima of distributions of upper-level temperatures and humidities, we choose to treat this SAM control simulation as the (-4K) SAM simulation, and the warm simulation of \cite{o2021response} as the (+0K) SAM simulation. We refer the reader interested in the details of the simulations and the coarse-graining (here by a factor of 8) to \cite{yuval2020stable}. Differences in climate model formulation and ML parameterization design lead to key differences in the mappings learned for SAM as compared to SPCAM3/SPCESM2, which we summarize below:
\begin{itemize}
    \item the input vector does not contain specific humidity, surface pressure, sensible heat fluxes, or latent heat fluxes, but instead contains the total non-precipitating water concentration and uses distance to the equator as a proxy for solar insolation,
    \item the output vector includes the subgrid total non-precipitating water tendency instead of the subgrid specific humidity tendency, and the subgrid liquid/ice static energy tendency instead of the subgrid temperature tendency,
    \item the output vector does not contain subgrid longwave and shortwave heating, 
    \item SAM uses a height-based vertical coordinate rather than a pressure-based one, and
    \item the generalization experiment is from (-4K) to (+0K) (unavailable (+4K) simulation).
\end{itemize}

\subsection{Normalization and Training, Validation, and Test Split\label{sub:Training,Val,Test}}

Both generalization experiments expose ML models to out-of-distribution inputs they have not been trained on. Following best ML practices \cite{geron2019hands}, we use the \textit{training} set to optimize the ML model's trainable parameters, save the trainable parameters that led to the best performance on the \textit{validation} set to avoid overfitting the training set, and evaluate the final model on samples from a separate \textit{test} set. We split each of the 8 simulations into training/validation/test sets by using non-contiguous 3-month periods (reported in Tab~S1) to avoid high temporal correlations between training/validation/test set samples \cite{karpatne2018machine}. Following \cite{rasp2018deep}, the normalization procedure involves subtracting the mean value of each input variable at each vertical level and dividing by the maximum range of that variable across the entire atmospheric column.

To understand which solutions are most promising for helping ML algorithms generalize to unseen conditions, we design two generalization experiments: (1) Training and validating ML models on cold simulations (-4K) and testing them on warm simulations (+4K for SPCAM3/SPCESM2 and +0K for SAM); and (2) training and validating ML models on aquaplanet simulations (SPCAM3) and testing them on Earth-like simulations with continents (SPCESM2).

\section{Theory \label{sec:Theory}}

We formally define a \textit{climate-invariant mapping} as a mapping that is unchanged across climates. In practice, it is difficult to find mappings that are exactly invariant and we will use the terminology climate-invariant for any mapping that remains approximately constant across climates. To achieve climate invariance, we introduce \textit{physically-based feature transformations}, defined as physically-informed functions that map the inputs/outputs to different inputs/outputs whose distributions vary little across climates. We deem the physical transformation to be climate-invariant if it is successful at limiting  distributions variations of the inputs/outputs across climates. Note that climate-invariant transformations are distinct from nondimensionalization in dimensional analysis, as nondimensionalization does not necessarily alter distribution shape while climate-invariant transformations may yield variables that have physical units.

\begin{figure*}
\centerline{\includegraphics[width=\linewidth]{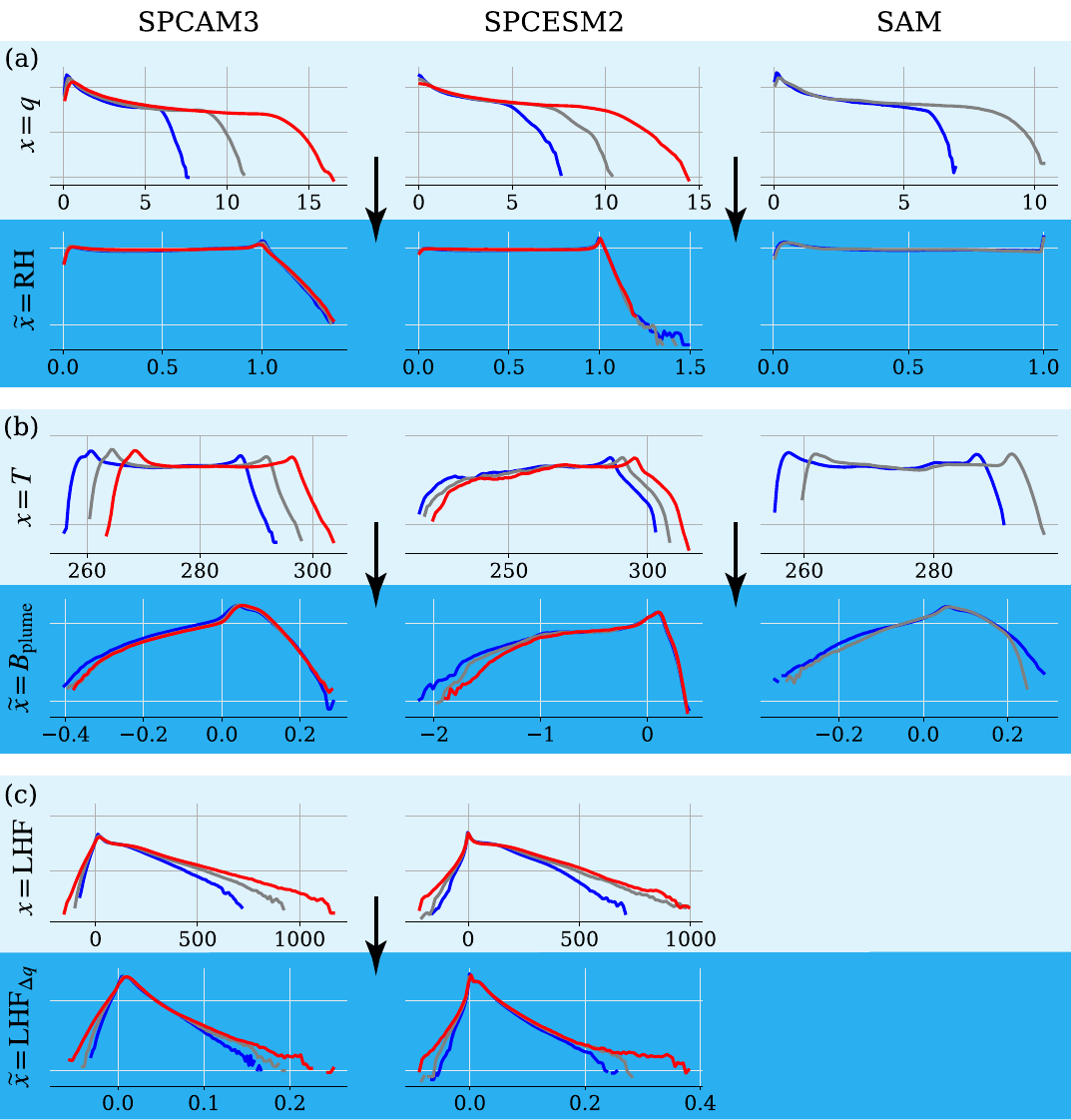}}
 \caption{\textbf{Physical transformations can align distributions across climates.} We show the univariate distributions of selected raw inputs $\boldsymbol{x} $: (a) 600hPa specific humidity, (b) 850hPa temperature, and (c) latent heat flux in the cold (blue), reference (gray), and warm (red) simulations of each model (SPCAM3, SPCESM2, and SAM). For each variable, we also show the PDFs of the transformed inputs $\boldsymbol{\widetilde{x}}$ as discussed in text. From top to bottom, the variables are $q$ (g/kg), RH, T (K), $B_{\mathrm{plume}}$ (m/s$^2$), LHF (W/m$^2$), and $\mathrm{LHF}_{\Delta q} $ ($\mathrm{kg\ m^{-2}s^{-1}} $). For a given variable and transformation, we use the same vertical logarithmic scale across models.  \label{f_PDFs}}
\end{figure*}

Throughout the following section, we compare two transformation options for each input, whose univariate PDFs are depicted for all three atmospheric models in Fig~\ref{f_PDFs}: No transformation (top) and our most successful transformation (bottom). All transformations are derived in SM B2. Our comparison relies on the Hellinger and Jensen-Shannon PDF distance metrics defined and calculated in Sec~\ref{sec:Materials_Methods} and SM~D1. To prevent information leaks from generalization test sets into the physically-informed ML framework, we take two precautions: (1) the physical transformations are fixed, meaning that their structure and parameters are non-trainable; and (2) transformations are ranked based on their generalization from (-4K) to (+0K) in SPCAM3. Our (+4K) results across models and configurations independently confirm this ranking. 

\paragraph{Specific Humidity\label{subsub:q}:} Without any transformation, the PDF of specific humidity $\boldsymbol{q} $ (Fig~\ref{f_PDFs}a, top) extends through a considerably larger range as the climate warms. This is because, barring supersaturation, $\boldsymbol{q} $ has a theoretical upper bound in a given climate, namely the saturation specific humidity,  which increases quasi-exponentially with temperature through the Clausius-Clapeyron relation (e.g., \cite{holton1973introduction,siebesma2020clouds}). The relative humidity (RH) transformation $ \boldsymbol{\tilde{q}_{\mathrm{RH}}} $ (Fig~\ref{f_PDFs}a, bottom) normalizes specific humidity by its saturation value. As a result, most of the RH PDF lies within $\left[0,1\right] $, except for a few atmospheric columns exhibiting super-saturation in SPCAM, and that PDF changes little as the climate warms \cite{manabe1967thermal}. In addition to capturing grid-scale saturation, $\boldsymbol{\tilde{q}_{\mathrm{RH}}} $ helps predict the subgrid effects of dry-air entrainment, known to regulate tropical convection \cite{BrownZhang1997,BrethertonEA2004,HollowayNeelin2009} (see SM B2b for details of RH calculations). 

\paragraph{Temperature\label{subsub:T}:}

The PDF of temperature $\boldsymbol{T} $(Fig~\ref{f_PDFs}b, top) shifts quasi-linearly as the climate warms. To address this shift without compromising the approximate invariance of tropopause temperatures with warming \cite{seeley2019fat,hartmann2002important,kuang2007testing}, we derive a temperature transformation directly relevant for moist convection: the buoyancy of a non-entraining, moist static energy-conserving plume $\boldsymbol{\tilde{T}_{\mathrm{buoyancy}}} $ (Fig~\ref{f_PDFs}b, bottom, see SM B2c for this buoyancy's derivation). This transformation is inspired by recently introduced lower-tropospheric buoyancy measures \cite{AhmedNeelin2018,AhmedEA2020}, but with an extension to the full troposphere \cite{SchiroEtal2018}. While $\boldsymbol{\tilde{T}_{\mathrm{buoyancy}}} $ does not explicitly include entrainment effects, the mapping of $\boldsymbol{\tilde{T}_{\mathrm{buoyancy}}\left(p\right)}$ and $\boldsymbol{\tilde{q}_{\mathrm{RH}}\left(p\right)}$ to heating and moisture sink will implicitly include these. This transformation captures leading order effects needed to yield approximate climate invariance (Fig~\ref{f_PDFs}b). $\boldsymbol{\tilde{T}_{\mathrm{buoyancy}}} $ increases physical interpretability by linking the vertical temperature structure and near-surface humidity changes to a metric that correlates well with deep convective activity \cite{AhmedEA2021}. $\boldsymbol{\tilde{T}_{\mathrm{buoyancy}}} $ also captures the role of near-surface humidity relative to the temperature structure aloft in contributing to moist convective instability in the tropics. 




\paragraph{Latent Heat Flux\label{subsub:LHF}:}

The last input whose distribution changes visibly with warming is the latent heat flux LHF (Fig~\ref{f_PDFs}c, top; the remaining inputs -- sensible heat flux and surface pressure -- change less with warming and are discussed in SM B2d). Similar to specific humidity, the increase of latent heat fluxes with warming is directly linked to the Clausius-Clapeyron relationship (e.g., \cite{hartmann2015global}). To address this shift, we leverage the bulk aerodynamic formula to represent surface fluxes and to provide a physics-motivated transformation of LHF using the near-surface saturation deficit (Fig~\ref{f_PDFs}c, bottom). This transforms LHF, a thermodynamic variable, into $\tilde{\mathrm{LHF}}_{\Delta q} $, approximately proportional to the magnitude of near-surface horizontal winds and density (e.g., \cite{hartmann2015global}), whose distributions vary less with warming. Note that this scaling is less effective over land (e.g., in SPCESM2) where evapotranspiration changes do not follow a Clausius-Clapeyron scaling.

\paragraph{} We now show that all three input transformations ($\boldsymbol{\tilde{q}_{\mathrm{RH}}\left(p\right)} $, $\boldsymbol{\tilde{T}_{\mathrm{buoyancy}}\left(p\right)} $, and $\tilde{\mathrm{LHF}}_{\Delta q} $) lead to statistically significant improvements in the ML models' ability to generalize. 

\section{Results \label{sec:results}}

The results are organized as follows. After demonstrating the benefits of progressively transforming the ML models’ inputs (Fig~\ref{f_Inp_Rescaling}), we show how climate-invariant models learn subgrid closures across climates and configurations \textit{during} training (Fig~\ref{tab:Results}, Fig~S7). We then discuss the global skill of different models \textit{after} training (Fig~\ref{f_R2}, Fig~S8\&9). Finally, we investigate the structure of climate-invariant mappings to understand \textit{why} they generalize better across climates (Fig~\ref{f_NN_visualization}), even when data from multiple climates are available (Fig~\ref{f_BothClimates}).

\subsection{Benefits of Incremental Input Transformations\label{sub:Input_Rescaling}} 

\begin{figure*}
\centering
\centerline{\includegraphics[width=\textwidth]{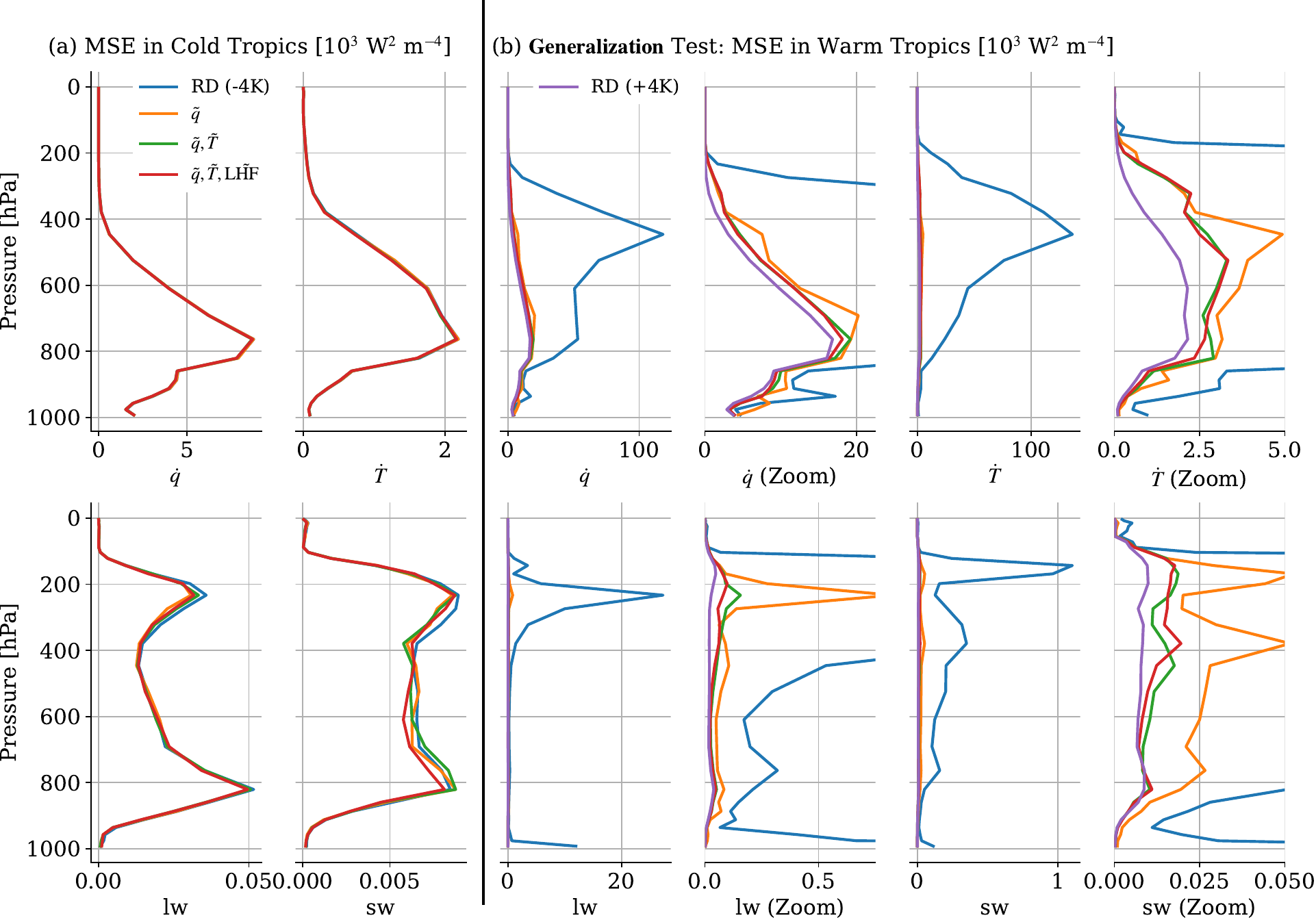}}
 \caption{\textbf{All neural networks, trained in the cold climate, exhibit low error in the cold climate's test set (a), but much larger error in the warm climate's test set (b).} This generalization error decreases as inputs are incrementally transformed: First no transformation (blue), then the vertical profile of specific humidity (orange), then the vertical profile of temperature (green), and finally latent heat fluxes (red). For reference, the purple line depicts an NN trained in the warm climate. We depict the tendencies' mean-squared error versus pressure, horizontally-averaged over the Tropics of SPCAM3 aquaplanet simulations, for the four model outputs: total moistening ($\dot{q}$), total heating ($\dot{T}$), longwave heating (lw) and shortwave heating (sw). Given that the raw-data NN's generalization error (blue line) greatly exceeds that of the transformed NNs, we zoom in on each panel to facilitate visualization.}\label{f_Inp_Rescaling}
\end{figure*}

In this section, we demonstrate that incrementally transforming the inputs of neural networks progressively improves their generalization abilities from the cold (-4K) aquaplanet (SPCAM3) simulation to the warm (+4K) aquaplanet simulation. The largest surface temperature jump tested in this study is between the cold aquaplanet simulation and the tropics of the warm aquaplanet simulation (``warm Tropics'' for short), defined as the regions with out-of-distribution surface temperatures, whose latitudes are between -15°S and 23°N (approximately the red regions in Fig~\ref{f_data}'s top-left subplots). To expose the failure modes of the ``brute force'' model and the benefits of progressively transforming the inputs, we first trained several NNs on the cold aquaplanet until they reached high accuracy (Fig~\ref{f_Inp_Rescaling}a) before testing their out-of-distribution generalizability in the warm Tropics (Fig~\ref{f_Inp_Rescaling}b). 

In the cold Tropics, the vertical profiles of the mean-squared error (MSE) are nearly indistinguishable for all types of NNs and roughly follow the vertical structure of subgrid variance, as discussed in \cite{gentine2018could,beucler2021enforcing}. When evaluated in the warm Tropics, the MSE of the ``brute force'' NN (blue line) increases by a factor of $\approx $10 and peaks above 100 W $^{2}$m$^{-4}$, underlining how \textit{raw-data NNs fail to generalize across climates}. As discussed in Section~\ref{sec:Theory}, we progressively transformed the inputs starting with specific humidity, which is transformed to relative humidity (orange line). This first transformation decreases the MSE so much (by a factor of 5-10) that we need to zoom in on each panel to distinguish the generalization abilities of additional NNs. Adding the transformations of temperature to plume buoyancy (green line) improves the generalization MSE for all variables. Adding the LHF to $\tilde{\mathrm{LHF}}_{\Delta q} $ transformation (red line) further decreases generalization MSE, except for shortwave heating where the MSE improves in the cold but not the warm climate. Impressively, transforming all three inputs decreases the MSE so much that \textit{the resulting climate-invariant NN’s MSE (red line) is within $\approx 25 \% $ of the MSE of a ``raw-data'' NN that was directly trained in the warm climate (purple line)}.

Hereafter, we use ``climate-invariant'' to refer to models for which all three inputs ($q$, $T$, LHF) but no outputs were transformed, solely based on physical principles. After demonstrating their success in the aquaplanet case, we are now ready to investigate how these climate-invariant models learn in the other climates and simulations introduced in Section~\ref{sec:Data}. 


\subsection{Learning across Climates and Configurations\label{sub:Learning_across_clim_geog}}

In this section, we show that climate-invariant models learn mappings that are valid across climates and configurations, and that their efficacy improves when used in conjunction with ML regularization techniques like BN and DP layers.

\begin{figure*}
\includegraphics[width=\linewidth]{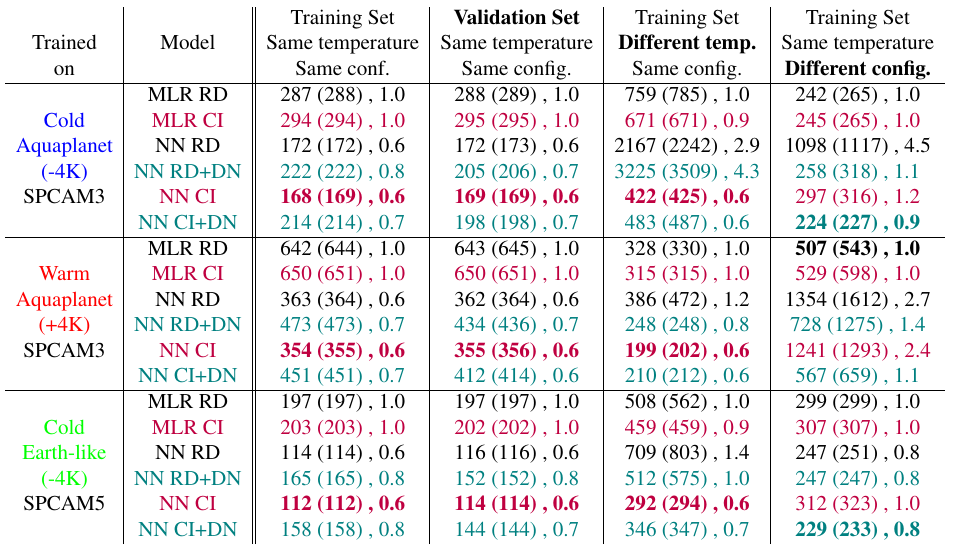}
\caption{\textbf{Model error across temperatures and configurations.} Mean-Squared Error (MSE, in units W$^{2}$ m$^{-4}$) of six models trained in three simulations (first column) and evaluated over the training or validation set of the same and two other simulations (last four columns). The models (second column) are raw-data (RD) or climate-invariant (CI), multiple linear regressions (MLR) or neural nets (NN), and sometimes include dropout layers preceded by a batch normalization layer (DN). The models are trained for 20 epochs. We first provide the MSE corresponding to the epoch of minimal validation loss, then the MSE averaged over the 5 epochs with lowest validation losses (in parentheses), and finally the MSE divided by the baseline MSE, where we use the raw-data multiple linear regression as baseline. Note that ``Different Temperature'' refers to (+4K) for (-4K) training sets and vice versa. In each application case, we highlight the best model's error using bold font.}
\label{tab:Results}
\end{figure*}

Fig~\ref{tab:Results} shows the MSE of ML models trained in three different datasets, and evaluated over their training and validation sets, and over test sets of different temperature and configuration. As discussed previously, climate-invariant NNs (NN CI) generalize better to warmer climates than raw-data NNs (NN RD). We go one step further by examining learning curves, defined as the MSE of an ML model at the end of each epoch during training (One epoch corresponds to the ML model being fed the entire training set once). Impressively, the learning curve of the climate-invariant NN trained in the cold aquaplanet but tested in the warm aquaplanet (starred blue line in Fig~S7a) is mostly decreasing, supporting this manuscript's key result: \textit{climate-invariant NNs continuously learn about subgrid thermodynamics in the warm aquaplanet as they are trained in the cold aquaplanet}. In contrast, the ``raw-data'' NN trained in the cold aquaplanet but tested in the warm aquaplanet makes extremely large generalization errors, which worsen as the model is trained in the cold aquaplanet (see SM D2 for details).  

\textit{Climate-invariant NNs also facilitate learning across configurations}, i.e., from the aquaplanet to the Earth-like simulations and vice-versa (see NN CI rows in Fig~\ref{tab:Results}). Climate-invariant transformations additionally improve the MLR baseline’s generalization ability, albeit less dramatically. This smaller improvement in MLR’s generalization abilities is linked to its relatively small number of trainable parameters, resulting in (1) ``raw-data'' MLRs generalizing better than ``raw-data'' NNs; and (2) MLRs having lower descriptiveness and fitting their training sets less well, limiting the maximal accuracy of climate-invariant MLRs on test sets.

There are a few cases in which transforming inputs does not fully solve the generalization problem, e.g., when trying to generalize from the aquaplanet to the Earth-like simulation. In that case, we leverage the fact that input transformations can easily be combined with standard techniques to improve generalization, such as DP layers before each activation function and a single BN layer before the first DP layer \cite{ioffe2015batch}. As DP layers randomly drop a fixed proportion of the trainable parameters during training, NNs with DP fit their training set less well (see NN CI+DN row of Fig~\ref{tab:Results}). However, they improve generalization in difficult cases (e.g., between cold aquaplanet and Earth-like simulations) and do not overly deteriorate generalization in cases where the input transformations work particularly well (e.g., from cold to warm aquaplanet). Our results suggest that \textit{combining physics-guided generalization methods (e.g., physical transformation of the inputs/outputs) with regularization methods (e.g., DP)} is advantageous and deserves further investigation. After analyzing the overall MSE during training, we now turn to the spatial characteristics of our ML models’ skill after training.

\subsection{Global Performance after Training\label{sub:Global_Perf}}

\begin{figure*}
\centering
 \includegraphics[width=\linewidth]{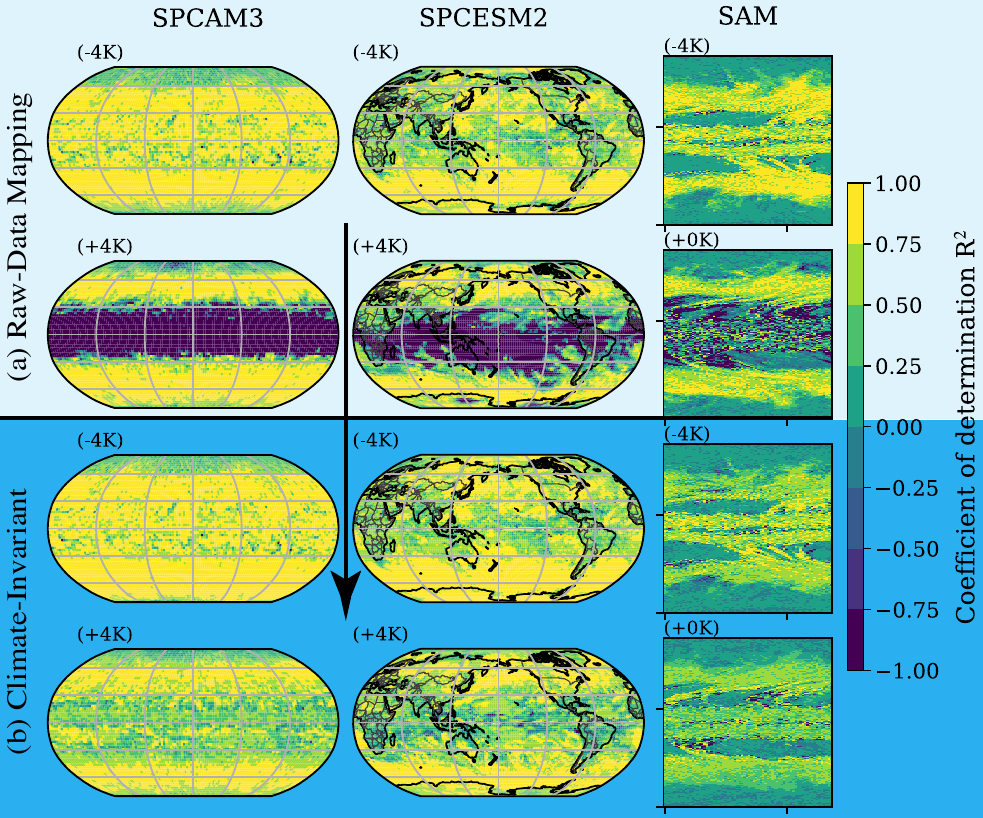}
 \caption{\textbf{Climate-invariant NNs (b) address the raw-data NNs' generalization problems in the ``warm Tropics'' (a).} This is demonstrated by the 500hPa subgrid heating's coefficient of determination $R^{2} $ calculated over the test set. We train NNs using the cold (-4K) training set of each model (SPCAM3, SPCESM2, and SAM). We note that these NNs do not use dropout nor batch normalization and refer readers to Fig~S8 for latitude-pressure cross-sections.}\label{f_R2}
\end{figure*}

In this section, we first compare the spatial characteristics of the brute force and climate-invariant NNs’ skill across climates of different temperatures to further establish the advantages of our climate-invariant transformation. These advantages are clearly visible in Fig~\ref{f_R2}, where the ``raw-data'' models struggle to generalize to the warm Tropics for all simulations despite fitting the cold training set well (Fig~\ref{f_R2}a). We can trace these generalization errors to warm temperature and moist atmospheric conditions the NNs were not exposed to during training, visible when comparing Fig~\ref{f_R2}a with Fig~\ref{f_data}a. In contrast, the climate-invariant models fit the warm climate almost as well as the cold climate they were trained in (Fig~\ref{f_R2}b). Note that Fig~\ref{f_R2} focuses on the horizontal map of a single output, i.e., the total subgrid heating at 500hPa, but that the horizontal map of other outputs, such as the near-surface subgrid heating (see Fig~S9) all exhibit the same pattern of ``raw-data'' models failing in the warm Tropics and the climate-invariant models mostly correcting these generalization errors. Finally, the spatial distribution of the skill in the training set (e.g., middle column of Fig~S8) is reassuringly consistent with the skill map of highly tuned NNs trained in similar conditions (e.g., \cite{mooers2021assessing}). This confirms that the ``raw-data'' models,  representative of state-of-the-art ML subgrid closures, fail to generalize. This failure is confirmed in the latitude-pressure map of the subgrid heating at all vertical levels shown in Fig~S8 and discussed in SM D3. 

To fully compare ML models across climate and configurations, we evaluate their overall MSEs in the training, validation, and both generalization test sets in Fig~\ref{tab:Results}. In addition to the MSE of minimal validation loss, we show the MSE averaged over the 5 epochs of minimal validation loss in parentheses to confirm that our models have converged. Consistent with Fig~S7's learning curves, climate-invariant NNs with DP and BN layers often demonstrate the highest level of generalizability (two rightmost columns of each row’s NN CI+DN models). While they fit their training sets less well, ``raw-data'' MLRs generalize better than ``raw-data'' NNs because they have fewer trainable parameters (see MLR RD and NN RD models). In Fig~\ref{tab:Results}, we also show that while DP and BN layers generally increase the generalization performance of ``raw-data'' NNs (NN RD+DN models), we can systematically improve these standard ML regularization methods by combining them with input transformations (NN CI+DN models). 






\subsection{Understanding Climate Invariance\label{sub:Understanding_clim_inv}}

\begin{figure*}
\centerline{\includegraphics[width=\linewidth]{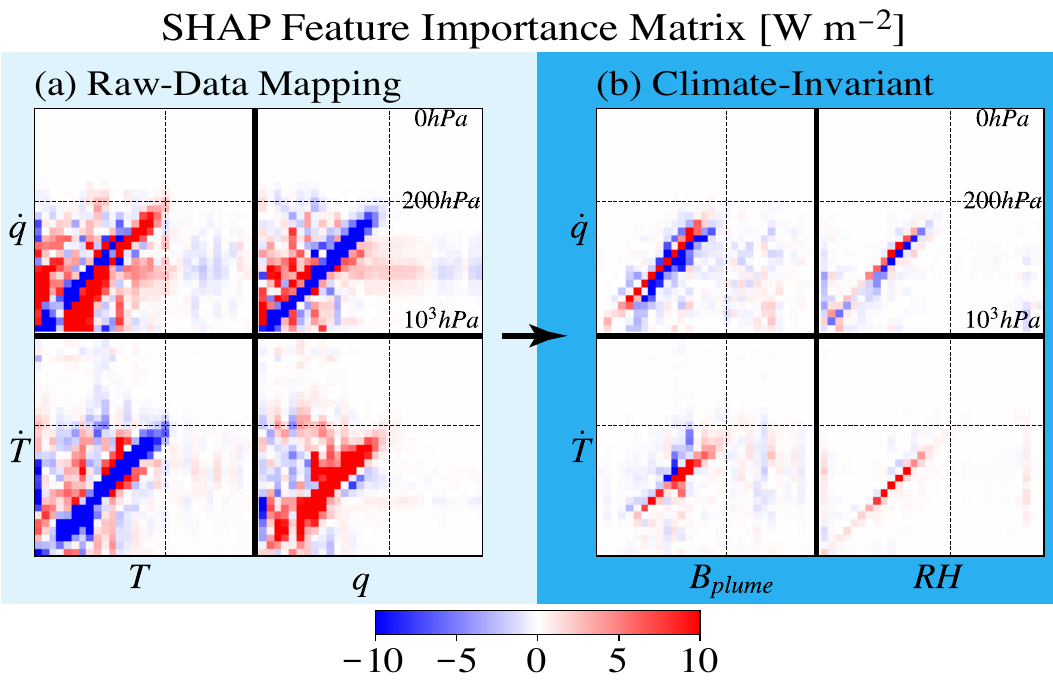}}
\caption{\textbf{Explainable artificial intelligence suggests that climate-invariant mappings are more spatially local.} We depict ${\cal M} $ for the (a) raw-data and (b) climate-invariant NNs trained in the SPCAM3 (+4K) warm aquaplanet simulation. The x-axes indicate the inputs' vertical levels, from the surface (left, 10$^{3}$hPa) to the top of the atmosphere (right, 0hPa), while the y-axes indicate the outputs' vertical levels, from the surface (bottom, 10$^{3}$hPa) to the top of the atmosphere (top, 0hPa). We additionally indicate the 200hPa vertical level with dotted black lines.}\label{f_NN_visualization}
\end{figure*}

To interpret our NN results, we use a game theory-based explainable ML approach, called SHAP \cite{lundberg2017unified,shapley201617}, to dissect climate-invariant mappings and provide insight into why they generalize better across climates and configurations. Note that MLRs are interpretable by construction, and we can draw preliminary conclusions by visualizing MLR weights without the need for explainable ML libraries (see SM D4). Similarly, we can directly visualize the NNs' linear responses \cite{kuang2007testing,herman2013linear,beucler2018linear,kuang2018linear} by calculating their Jacobians (gradients) via automatic differentiation \cite{brenowitz2020interpreting}. However, the difference between RD and CI MLRs is small and the Jacobians \cite{herman2013linear} cannot always be reliably used to explain nonlinear NN predictions as they are first-order derivatives calculated over a sample \cite{kindermans2019reliability}.

Therefore, as climate-invariant NNs have shown superior generalizability from cold to warm climates (see NN CI errors in Fig~\ref{tab:Results}), we employ SHAP's Kernel Explainer to elucidate the NNs' generalizability. We choose this attribution method for its versatility, as it can be used for any multi-input/output ML model. SHAP estimates the impact of a particular input value $x_{i} $ on each output $y_{j} $ of our model. It is designed with a local accuracy property, ensuring that the sum of the effects of individual inputs equals the difference between $y_{j} $ and its average value in the training set:

\begin{equation}
\sum_{i}\mathrm{SHAP}\left(x_{i},y_{j}\right)\overset{\mathrm{def}}{=}y_{j}^{\prime},
\end{equation}
where we have introduced the deviation $y^{\prime}_{j} $, defined as the difference between $y_{j} $ and its training set average $y_{i}^{\prime}\overset{\mathrm{def}}{=}y_{i}-\left\langle y_{i}\right\rangle _{{\cal E}} $. We use these ``Shapley values'' to build a nonlinear feature matrix $\cal M $ capturing the influence of an input $x_{i} $ on an output $y_{j} $:

\begin{equation}
{\cal M}_{\mathrm{ij}}\overset{\mathrm{def}}{=}\left\langle \mathrm{sign}\left(x_{i}^{\prime}\right)\times\mathrm{SHAP}\left(x_{i},y_{j}\right)\right\rangle _{{\cal E}},
\end{equation}
where we use the sign of the input deviation $x^{\prime}_{i} $ to make ${\cal M}_{\mathrm{ij}} $ positive if $x^{\prime}_{i} $ and $y^{\prime}_{j} $ have the same sign, e.g., if a positive input deviation leads to a positive output deviation. In the particular case of the MLR defined in SM Eq~25, the nonlinear feature matrix becomes the regression weight matrix multiplied by the absolute value of the input deviations: ${\cal M}_{\mathrm{ij}}=\left\langle A_{ij}\left|x_{i}^{\prime}\right|\right\rangle _{{\cal E}}$, confirming that the feature matrix ${\cal M} $ is a nonlinear extension of the Jacobian ($A$ in the MLR case). 

In Fig~\ref{f_NN_visualization}, each panel depicts the SHAP feature matrix $\cal{M} $ for a given model: raw-data (a) and climate-invariant (b). Each model's inputs (e.g., specific humidity $q $  and temperature $T $) are organized on the x-axis from the surface to the top of the atmosphere. Each model's outputs (subgrid moistening $\dot{q} $ and subgrid heating $\dot{T} $; see Fig~S14/15 for subgrid longwave heating $\mathrm{lw} $ and subgrid shortwave heating $\mathrm{sw} $) are organized on the y-axis, from the surface to the top of the atmosphere. Following a horizontal line shows how different inputs contribute to a given output, while following a vertical line shows how a given input influences different outputs.

Fig~\ref{f_NN_visualization} contains a wealth of information about subgrid closures trained in aquaplanet simulations; we focus here on visualizing how the climate-invariant NNs (b) operate in ways that generalize better than their raw-data mapping counterpart (a). Consider the row for subgrid heating $\dot{T} $. In the raw-data case (a), $\cal{M} $ has large coefficients in most of the troposphere (in the entire square below the dashed lines depicting the approximate tropopause level). This means that specific humidity and temperature deviations at \textit{all} levels impact subgrid heating at a given level, i.e. there are large non-local relations in the vertical. 
Some non-local relations are physically plausible for convection, since buoyant plumes tend to rise from the surface, and indeed near-surface $T$ and $q$ influence $\dot{T}$ through the entire troposphere. However, in this model, moisture at  higher altitudes appears to influence $\dot{q}$ at lower altitudes, raising suspicions that some of the raw-data NN's non-localities are not causal, but rather due to high auto-correlations within the input's vertical profile, as \cite{brenowitz2020interpreting} showed could happen. Temperature variations are observed to have strong vertical correlations \cite{HollowayNeelin2007} in part because of deep convective effects. Because temperature affects the saturation threshold for moisture, the RD NN will have to correctly capture the effects of both temperature and moisture wherever either has influence. 
In contrast, in the $B_{plume}$-RH climate-invariant case (b), leading non-local effects between the boundary layer and the free troposphere have already been taken into account in the buoyancy formulation, and the temperature-dependent saturation threshold is built into RH. Thus, $\cal{M}$ for $\dot{T}$ tends to be concentrated near the red diagonal, meaning that positive deviations of plume buoyancy and relative humidity increase subgrid heating near the same vertical level. The use of domain knowledge has effectively reduced the effects that must be estimated by the NN for the climate-invariant models. This tends to yield differences between the models trained in the cold and the warm climates that are much smaller than for the raw-data models (see last column of Fig~S15).


\subsection{Advantages of Climate Invariance with Multi-Climate Data}

\begin{figure*}
\centerline{\includegraphics[width=\linewidth]{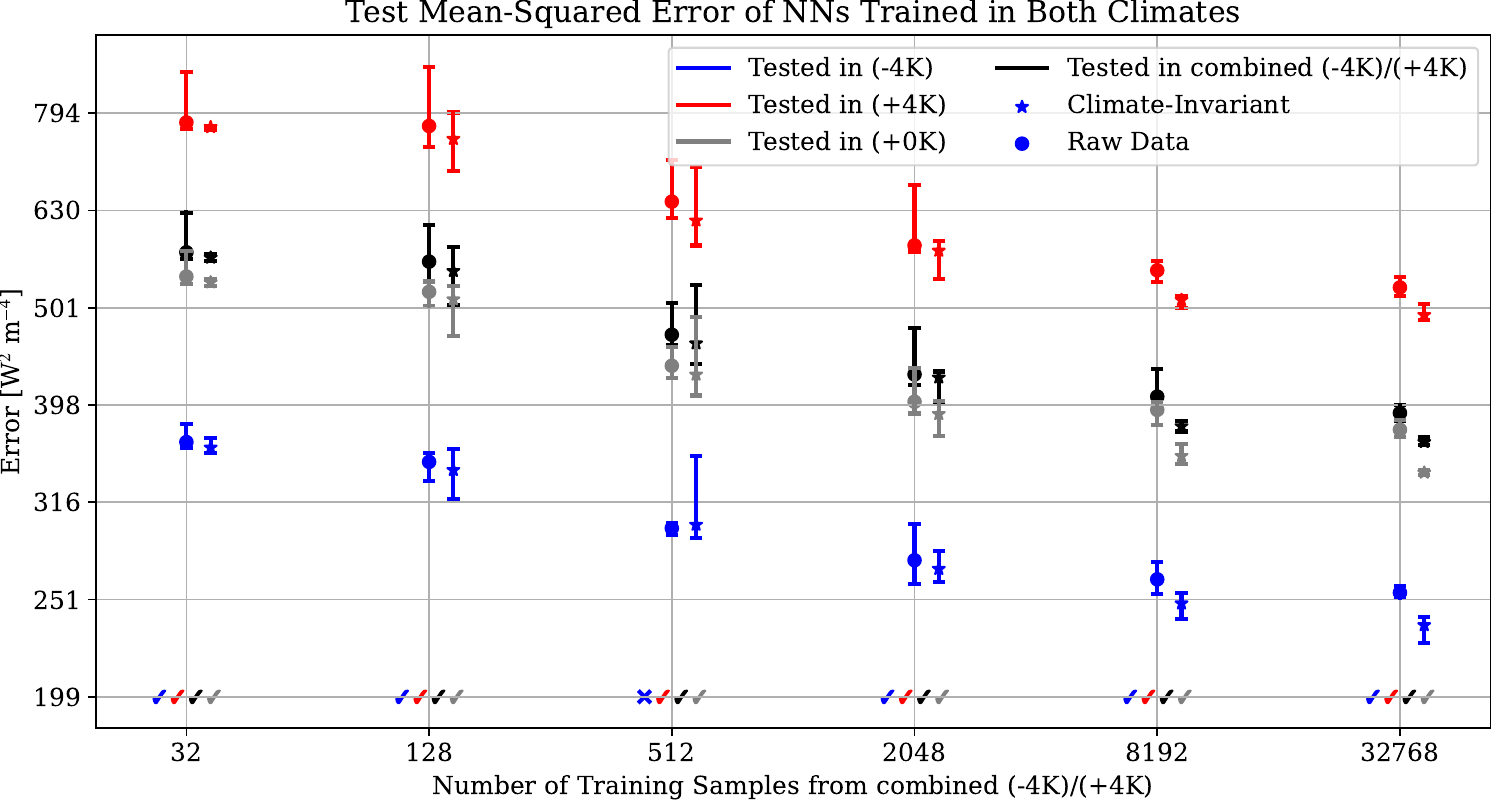}}
 \caption{\textbf{Climate-invariant (CI) neural networks trained on datasets containing both cold (-4K) and warm (+4K) samples outperform raw-data (RD) models offline in $\approx$95\% of cases, with less sensitivity to the data partition used for training.} Dots on the left represent the median RD error from a 10-fold cross-validation without replacement, with horizontal ticks indicating the 1st and 9th deciles. Stars on the right correspond to the median CI error. Ticks denote the majority of cases, for which CI models outperform RD models, even when data from both climates are available; crosses indicate the rare exceptions. We use a logarithmic scale for both axes. \label{f_BothClimates}}
\end{figure*}

It is natural to wonder if the benefits of climate invariance carry over to training scenarios that entail data from multiple simulations spanning diverse temperatures. In contrast, until now, we have only trained ML models on single climate simulations. To find out, we examine the benefits of our climate invariant transformation approach under the ideal scenario where we have access to data from multiple climates. We conduct experiments where we train NNs on both the cold (-4K) and warm (+4K) aquaplanet simulations. We progressively increase the amount of training data to assess data efficiency. Throughout the experiment, we use eight batches and systematically increase the batch size by powers of 2, starting with a batch size of 4. Note that we obtain similar results when increasing the number of batches while keeping the batch size fixed (not shown). To obtain well-defined uncertainty estimates, we employ a 10-fold cross-validation procedure via random sampling without replacement. Our findings, depicted in Fig~\ref{f_BothClimates}, demonstrate that CI NNs (1) consistently outperform RD NNs, particularly in data-rich scenarios; and (2) exhibit lower sensitivity to the training data partition, resulting in more reliable offline performance with reduced variability in test errors. This confirms that \textit{our climate-invariant mapping enhances data efficiency, performance, and fit reproducibility across different climates, even when training data from multiple climates are available}.


\section{Discussion \label{sec:Conclusion}}

In the context of climate change, we hypothesized that ML models emulating climate-invariant mappings (Fig~\ref{f1}), for which the inputs/outputs distributions change little across climates (Fig~\ref{f_PDFs}), generalize much better than ML models emulating raw-data mappings, for which the inputs/outputs distributions change substantially across climates. Tested on a suite of storm-resolving atmospheric simulations with different surface temperatures in three atmospheric models with distinct configurations (Fig~\ref{f_data}), physically-transformed NNs generalize better as their inputs are progressively transformed (Fig~\ref{f_Inp_Rescaling}). Climate-invariant NNs whose inputs have all been transformed learn mappings that are robust to temperature and configuration changes (Fig~\ref{tab:Results}), and hence exhibit superior generalization skill almost everywhere on the globe (Fig~\ref{f_R2}), including when data from multiple climates are available (Fig~\ref{f_BothClimates}). Finally, attribution maps reveal that in addition to providing control on the features' distributions, climate-invariant NNs learn more spatially local mappings that facilitate generalization across climates and configurations (Fig~\ref{f_NN_visualization}). 


From a computational perspective, incorporating physical knowledge, here of climate change, into an ML framework to improve its generalization skill is a successful example of using domain knowledge to extract more informative predictors, informally referred to as ``feature engineering'' (e.g., \cite{zheng2018feature}). This also aids interpretability of the mapping. From a climate science perspective, requiring that a nonlinear statistical model of the atmosphere generalize across climate is a stringent test that helped us discover new mappings. This climate-invariant mapping is more robust to climate and configuration changes, and is more advantageous than directly using model and observational outputs (e.g., specific humidity, temperature), even when data are available in various climate regimes. In the particular case of subgrid thermodynamics, our generalization results suggest the possibility of NN-powered closures that could work in Earth-like settings, even in vastly different climate conditions. Finally, the attribution maps suggest the possibility of new analytic representations of convection from data, facilitated by the more local climate-invariant representation of subgrid thermodynamics. Our strategy paves the way for the successful use of machine learning models for climate change studies. 

\section{Materials and Methods \label{sec:Materials_Methods}}

\begin{figure*}
\centerline{\includegraphics[width=\textwidth]{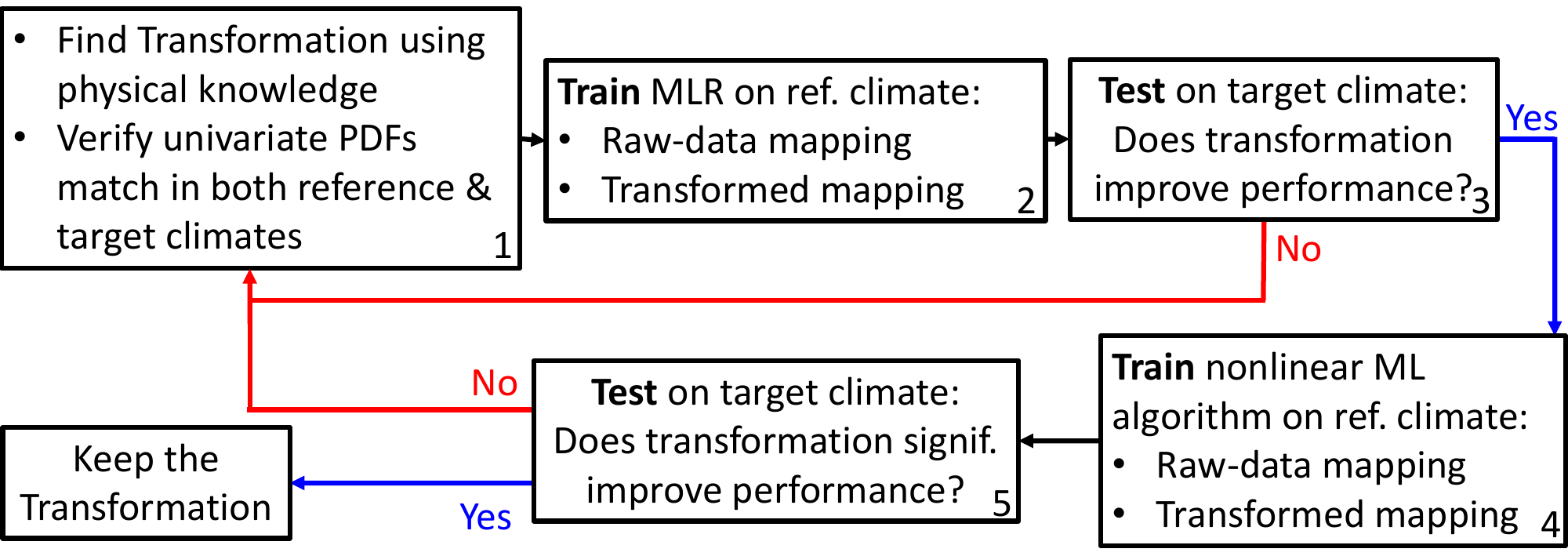}}
 \caption{\textbf{Proposed five-step workflow to find climate-invariant transformations.} The transformations help Machine Learning (ML) models generalize from a reference (ref.) climate to a target one, using (top) a baseline Multiple Linear Regression (MLR) as an initial guide.\label{f_rescaling}}
\end{figure*}

This section outlines how to find feature transformations yielding climate invariance. Fig~\ref{f_rescaling} illustrates our proposed workflow for finding robust inputs/outputs transformations that transform the initial ``raw-data'' mapping into a climate-invariant mapping when combined. Note that this workflow assumes that we cannot or do not want to retrain ML algorithms in the target climate, which excludes automatically finding a transformation by training a model. This limitation could arise because the data in the target climate are insufficient or less reliable, or because we seek to uncover new physical relations that hold across an even wider range of climates.

The first step is to propose a physical transformation to implement. We can do this through knowledge of robust physical or statistical relations that link and/or preserve distributions (e.g., state equations, self-similarities, conservation laws, accurate empirical relations, etc) as modeled in SM B2. These relations help derive invariants (e.g., \cite{shepherd1990symmetries}) under a change in thermodynamic conditions. Before taking the time to implement this transformation in the ML workflow, we can verify that the PDFs of the transformed inputs/outputs (approximately) match in the training and target climates. Ideally, the \textit{joint} PDFs of the transformed inputs/outputs would match. In practice, because it is easier to transform one variable at a time and the data are often insufficient in the target climate, we can fall back on the necessary (but not sufficient) condition that the univariate PDFs of the transformed inputs/outputs must match in the training and target climates. Mathematically, this match can be quantified using PDF distance metrics. 

An additional challenge is that the original and transformed variables may have different units and range, meaning that any nonlinear distance metric will complicate the PDF comparison. To address this, we normalize the PDFs and their support variables $X $ so that the PDFs' domains strictly lie within $\left[0,1\right] $. For a given variable, we use the same normalization factors across climates:

\begin{equation}
     X_{\mathrm{norm}}\overset{\mathrm{def}}{=}\frac{X-\min_{\mathrm{cl}}X}{\max_{\mathrm{cl}}X-\min_{\mathrm{cl}}X},
\end{equation}
\begin{equation}
    \mathrm{PDF_{norm}}\overset{\mathrm{def}}{=}\frac{\mathrm{PDF}}{\int_{0}^{1}dX_{\mathrm{norm}}\times\mathrm{PDF}\left(X_{\mathrm{norm}}\right)},
\end{equation}

where $ \mathrm{PDF_{norm}}$ is the transformed PDF and $X_{\mathrm{norm}} $ is its transformed support, and $\max_{\mathrm{cl}} $ and $\min_{\mathrm{cl}} $ respectively refer to the maximum and minimum operators over the variables' domains and across climates, i.e. over the (-4K), (+0K), and (+4K) simulations.

Once the PDFs of each variable are normalized, we may pick any informative PDF distance metric to quantify how PDFs match across climates. 
Here, we pick the commonly-used Hellinger distance between two PDFs $p $ and $q $, formally defined \cite{beran1977minimum} as:

\begin{equation}
    \mathrm{\cal H}\left(p,q\right)\overset{\mathrm{def}}{=}\sqrt{\int_{0}^{1}\frac{dx}{2}\left[\sqrt{p\left(x\right)}-\sqrt{q\left(x\right)}\right]^{2}}.
\end{equation}
This distance is symmetric (i.e., the arguments' order does not affect the outcome) and easy to interpret: $\mathrm{\cal H}\left(p,q\right) $ is bounded by 0 (when $p=q $) and $100\% $ (when $p $ is zero whenever $q $ is positive and vice-versa). In SM D1, we show that our results using the Hellinger distance (see Tab~S1) are consistent with those using the Jensen-Shannon distance \cite{endres2003new} (see Tab~S3), a PDF distance metric giving large weights to the PDFs' tails that tend to be particularly problematic for generalization purposes.

Once the univariate PDFs of the physically transformed variables match across climates, the second step is to train two inexpensive or ``baseline'' models on the reference climate to quickly check whether the transformation improves an ML model's generalization ability: (1) A ``raw-data'' model without the transformation; and (2) a ``climate-invariant'' model with the transformation. If the transformation does not improve the baseline model's generalization abilities (i.e., (2) performs worse than (1) in the target climate), then the transformation may not be appropriate. Note that we trained MLR baselines to create climate-invariant NNs, but the ML model used to define the baseline should be tailored to the desired final ML model

If the transformation improves the inexpensive baseline model's performances, the last step is to train the ``raw-data'' and ``climate-invariant'' versions of the desired ML model (usually nonlinear) on the reference climate. If the physical transformation improves the desired ML model's generalization abilities (i.e., the climate-invariant model beats the raw-data model in the target climate using the same performance metric calculated over a \textit{validation} set), then we may keep the transformation. This workflow may be repeated for the ML model's additional input/output variables until the emulated mapping is as climate-invariant as possible.

Before applying this workflow to subgrid thermodynamics closures, we underline one of its key challenges: Because some transformations are much more impactful than others, it is often not possible to develop each physical transformation independently. In our case, the specific humidity inputs vary the most across climates, meaning that transforming specific humidity affects the model's generalization abilities the most. As a result, initial experiments that independently tested the effect of transforming temperature suggested a negative impact of temperature transformation on generalization ability (not shown). This initial result was later invalidated by experiments that jointly transformed specific humidity and temperature. Following this, we adopt a progressive input transformation approach, where the most important inputs are transformed first: Specific humidity, then temperature, and finally surface energy fluxes.  

\section{Acknowledgments}

We thank William Boos for providing the SAM hypohydrostatic simulation output, Joyce Meyerson for helping us improve figures, and Andrea Jenney, Imme Ebert-Uphoff, and the CBRAIN discussion group for help and advice that greatly benefited the present manuscript.

\paragraph{Funding:} TB and MP acknowledge funding from NSF grants OAC1835863 and AGS-1734164. PG, MP, SY, and JL acknowledge NSF funding from the LEAP Science and Technology center grant (2019625-STC). MP acknowledges the DOE's Exascale Computing Project (17-SC-20-SC), Biological and Environmental Research (BER) program (DE-SC0022331), Advanced Scientific Computing Research (ASCR) program (DE-SC0022255). MP and SY also acknowledge the DOE EAGLES project (74358). JY acknowledges support from the EAPS Houghton-Lorenz postdoctoral fellowship. PG thanks NSF grant AGS-1734164 and USMILE European Research Council grant. PG, POG, and JY received M$^{2}$LInES research funding by the generosity of Eric and Wendy Schmidt by recommendation of the Schmidt Futures program. FA and JDN acknowledge NSF AGS-1936810; FA acknowledges NSF AGS-2225956. We also thank the Extreme Science and Engineering Discovery Environment supported by NSF grant number ACI-1548562 (charge numbers TG-ATM190002 and TG-ATM170029), the Texas Advanced Computing Center (TACC; ATM-20009), and HPC support from Cheyenne (doi:10.5065/D6RX99HX) provided by NCAR's Computational and Information Systems Laboratory (sponsored by the NSF) for computational resources.

\paragraph{Contributions:}\ \\    
Conceptualization: TB, MP, SR, and PG\\   
Methodology: All authors\\   
Investigation: TB, MP, JY, AG, LP, JL, and SY\\   
Visualization: TB, MP, PG, DN, and JY\\   
Supervision: TB, MP, and PG\\   
Writing—original draft: TB\\   
Writing—review \& editing: All authors\\

\paragraph{Competing interests:} The authors declare that they have no competing interests.

\paragraph{Data and materials availability:} All data needed to evaluate the conclusions in the paper are present in the paper and/or the Supplementary Materials. See SM A for details about code and data availability.

\section{Supplementary materials}     
Supplementary Text (SM A to SM D)\\   
Figs. S1 to S15\\       
Tables S1 to S3\\

\onecolumn
\small

\clearpage




\begin{thebibliography}{}

\bibitem{reichstein2019deep}
Reichstein, Markus, Gustau Camps-Valls, Bjorn Stevens, Martin Jung, Joachim Denzler, Nuno Carvalhais, and fnm Prabhat. "Deep learning and process understanding for data-driven Earth system science." \textit{Nature} 566, no. 7743 (2019): 195-204.

\bibitem{beucler2021machine}
Beucler, Tom, Imme Ebert‐Uphoff, Stephan Rasp, Michael Pritchard, and Pierre Gentine. "Machine learning for clouds and climate." \textit{Clouds and their Climatic Impacts: Radiation, Circulation, and Precipitation} (2023): 325-345.

\bibitem{irrgang2021towards}
Irrgang, Christopher, Niklas Boers, Maike Sonnewald, Elizabeth A. Barnes, Christopher Kadow, Joanna Staneva, and Jan Saynisch-Wagner. "Towards neural Earth system modelling by integrating artificial intelligence in Earth system science." \textit{Nature Machine Intelligence} 3, no. 8 (2021): 667-674.

\bibitem{de2023machine}
de Burgh-Day, Catherine O., and Tennessee Leeuwenburg. "Machine learning for numerical weather and climate modelling: a review." \textit{Geoscientific Model Development} 16, no. 22 (2023): 6433-6477.

\bibitem{molina2023review}
Molina, Maria J., Travis A. O’Brien, Gemma Anderson, Moetasim Ashfaq, Katrina E. Bennett, William D. Collins, Katherine Dagon, Juan M. Restrepo, and Paul A. Ullrich. "A Review of Recent and Emerging Machine Learning Applications for Climate Variability and Weather Phenomena." \textit{Artificial Intelligence for the Earth Systems} (2023): 1-46.

\bibitem{ukkonen2020accelerating}
Ukkonen, Peter, Robert Pincus, Robin J. Hogan, Kristian Pagh Nielsen, and Eigil Kaas. "Accelerating radiation computations for dynamical models with targeted machine learning and code optimization." \textit{Journal of Advances in Modeling Earth Systems} 12, no. 12 (2020): e2020MS002226.

\bibitem{belochitski2011tree}
Belochitski, Alexei, Peter Binev, Ronald DeVore, Michael Fox-Rabinovitz, Vladimir Krasnopolsky, and Philipp Lamby. "Tree approximation of the long wave radiation parameterization in the NCAR CAM global climate model." \textit{Journal of Computational and Applied Mathematics} 236, no. 4 (2011): 447-460.

\bibitem{gristey2020relationship}
Gristey, Jake J., Graham Feingold, Ian B. Glenn, K. Sebastian Schmidt, and Hong Chen. "On the relationship between shallow cumulus cloud field properties and surface solar irradiance." \textit{Geophysical Research Letters} 47, no. 22 (2020): e2020GL090152.

\bibitem{lagerquist2021using}
Lagerquist, Ryan, David Turner, Imme Ebert-Uphoff, Jebb Stewart, and Venita Hagerty. "Using deep learning to emulate and accelerate a radiative transfer model." \textit{Journal of Atmospheric and Oceanic Technology} 38, no. 10 (2021): 1673-1696.

\bibitem{chantry2021machine}
Chantry, Matthew, Sam Hatfield, Peter Dueben, Inna Polichtchouk, and Tim Palmer. "Machine learning emulation of gravity wave drag in numerical weather forecasting." \textit{Journal of Advances in Modeling Earth Systems} 13, no. 7 (2021): e2021MS002477.

\bibitem{espinosa2022machine}
Espinosa, Zachary I., Aditi Sheshadri, Gerald R. Cain, Edwin P. Gerber, and Kevin J. DallaSanta. "Machine learning gravity wave parameterization generalizes to capture the QBO and response to increased CO2." \textit{Geophysical Research Letters} 49, no. 8 (2022): e2022GL098174.

\bibitem{matsuoka2020application}
Matsuoka, Daisuke, Shingo Watanabe, Kaoru Sato, Sho Kawazoe, Wei Yu, and S. Easterbrook. "Application of deep learning to estimate atmospheric gravity wave parameters in reanalysis data sets." \textit{Geophysical Research Letters} 47, no. 19 (2020): e2020GL089436.

\bibitem{yuval2021momentum}
Yuval, Janni, and Paul A. O’Gorman. "Neural‐network parameterization of subgrid momentum transport in the atmosphere." \textit{Journal of Advances in Modeling Earth Systems} 15, no. 4 (2023): e2023MS003606.

\bibitem{morrison2020bayesian}
Morrison, Hugh, Marcus van Lier-Walqui, Matthew R. Kumjian, and Olivier P. Prat. "A Bayesian approach for statistical–physical bulk parameterization of rain microphysics. Part I: Scheme description." \textit{Journal of the Atmospheric Sciences} 77, no. 3 (2020): 1019-1041.

\bibitem{seifert2020potential}
Seifert, Axel, and Stephan Rasp. "Potential and limitations of machine learning for modeling warm‐rain cloud microphysical processes." \textit{Journal of Advances in Modeling Earth Systems} 12, no. 12 (2020): e2020MS002301.

\bibitem{gettelman2021machine}
Gettelman, Andrew, David John Gagne, C‐C. Chen, M. W. Christensen, Z. J. Lebo, Hugh Morrison, and Gabrielle Gantos. "Machine learning the warm rain process." \textit{Journal of Advances in Modeling Earth Systems} 13, no. 2 (2021): e2020MS002268.

\bibitem{franccois2021adjusting}
François, Bastien, Soulivanh Thao, and Mathieu Vrac. "Adjusting spatial dependence of climate model outputs with cycle-consistent adversarial networks." \textit{Climate dynamics} 57 (2021): 3323-3353.

\bibitem{pan2021learning}
Pan, Baoxiang, Gemma J. Anderson, André Goncalves, Donald D. Lucas, Céline JW Bonfils, Jiwoo Lee, Yang Tian, and Hsi‐Yen Ma. "Learning to correct climate projection biases." \textit{Journal of Advances in Modeling Earth Systems} 13, no. 10 (2021): e2021MS002509.

\bibitem{zantedeschi2019cumulo}
Zantedeschi, Valentina, Fabrizio Falasca, Alyson Douglas, Richard Strange, Matt J. Kusner, and Duncan Watson-Parris. "Cumulo: A dataset for learning cloud classes." \textit{arXiv preprint arXiv:1911.04227} (2019).

\bibitem{rasp2020combining}
Rasp, Stephan, Hauke Schulz, Sandrine Bony, and Bjorn Stevens. "Combining crowdsourcing and deep learning to explore the mesoscale organization of shallow convection." \textit{Bulletin of the American Meteorological Society} 101, no. 11 (2020): E1980-E1995.

\bibitem{watson2020large}
Watson‐Parris, Duncan, S. A. Sutherland, M. W. Christensen, Ryan Eastman, and Philip Stier. "A large‐scale analysis of pockets of open cells and their radiative impact." \textit{Geophysical Research Letters} 48, no. 6 (2021): e2020GL092213.

\bibitem{denby2020discovering}
Denby, L. "Discovering the importance of mesoscale cloud organization through unsupervised classification." \textit{Geophysical Research Letters} 47, no. 1 (2020): e2019GL085190.

\bibitem{han2020moist}
Han, Yilun, Guang J. Zhang, Xiaomeng Huang, and Yong Wang. "A moist physics parameterization based on deep learning." \textit{Journal of Advances in Modeling Earth Systems} 12, no. 9 (2020): e2020MS002076.

\bibitem{brenowitz2018prognostic}
Brenowitz, Noah D., and Christopher S. Bretherton. "Prognostic validation of a neural network unified physics parameterization." \textit{Geophysical Research Letters} 45, no. 12 (2018): 6289-6298.

\bibitem{krasnopolsky2013using}
Krasnopolsky, Vladimir M., Michael S. Fox-Rabinovitz, and Alexei A. Belochitski. "Using ensemble of neural networks to learn stochastic convection parameterizations for climate and numerical weather prediction models from data simulated by a cloud resolving model." \textit{Advances in Artificial Neural Systems} 2013 (2013): 5-5.

\bibitem{breiman2001random}
Breiman, Leo. "Random forests." \textit{Machine learning} 45 (2001): 5-32.

\bibitem{o2018using}
O'Gorman, Paul A., and John G. Dwyer. "Using machine learning to parameterize moist convection: Potential for modeling of climate, climate change, and extreme events." \textit{Journal of Advances in Modeling Earth Systems} 10, no. 10 (2018): 2548-2563.

\bibitem{hernanz2022evaluation}
Hernanz, Alfonso, Juan Andrés García‐Valero, Marta Domínguez, and Ernesto Rodríguez‐Camino. "A critical view on the suitability of machine learning techniques to downscale climate change projections: Illustration for temperature with a toy experiment." \textit{Atmospheric Science Letters} 23, no. 6 (2022): e1087.

\bibitem{rasp2018deep}
Rasp, Stephan, Michael S. Pritchard, and Pierre Gentine. "Deep learning to represent subgrid processes in climate models." \textit{Proceedings of the National Academy of Sciences} 115, no. 39 (2018): 9684-9689.

\bibitem{beucler2020towards}
Beucler, Tom, Michael Pritchard, Pierre Gentine, and Stephan Rasp. "Towards physically-consistent, data-driven models of convection." In \textit{Igarss 2020-2020 ieee international geoscience and remote sensing symposium}, pp. 3987-3990. IEEE, 2020.

\bibitem{clark2022correcting}
Clark, Spencer K., Noah D. Brenowitz, Brian Henn, Anna Kwa, Jeremy McGibbon, W. Andre Perkins, Oliver Watt‐Meyer, Christopher S. Bretherton, and Lucas M. Harris. "Correcting a 200 km resolution climate model in multiple climates by machine learning from 25 km resolution simulations." \textit{Journal of Advances in Modeling Earth Systems} 14, no. 9 (2022): e2022MS003219.

\bibitem{doury2022regional}
Doury, Antoine, Samuel Somot, Sebastien Gadat, Aurélien Ribes, and Lola Corre. "Regional climate model emulator based on deep learning: Concept and first evaluation of a novel hybrid downscaling approach." \textit{Climate Dynamics} 60, no. 5-6 (2023): 1751-1779.

\bibitem{guillaumin2021stochastic}
Guillaumin, Arthur P., and Laure Zanna. "Stochastic‐deep learning parameterization of ocean momentum forcing." \textit{Journal of Advances in Modeling Earth Systems} 13, no. 9 (2021): e2021MS002534.

\bibitem{molina2021benchmark}
Molina, Maria J., David John Gagne, and Andreas F. Prein. "A benchmark to test generalization capabilities of deep learning methods to classify severe convective storms in a changing climate." \textit{Earth and Space Science} 8, no. 9 (2021): e2020EA001490.

\bibitem{lecun1995convolutional}
LeCun, Yann, and Yoshua Bengio. "Convolutional networks for images, speech, and time series." \textit{The handbook of brain theory and neural networks} 3361, no. 10 (1995): 1995.

\bibitem{ioffe2015batch}
Ioffe, Sergey, and Christian Szegedy. "Batch normalization: Accelerating deep network training by reducing internal covariate shift." In \textit{International conference on machine learning}, pp. 448-456. pmlr, 2015.

\bibitem{srivastava2014dropout}
Srivastava, Nitish, Geoffrey Hinton, Alex Krizhevsky, Ilya Sutskever, and Ruslan Salakhutdinov. "Dropout: a simple way to prevent neural networks from overfitting." \textit{The journal of machine learning research} 15, no. 1 (2014): 1929-1958.

\bibitem{ying2019overview}
Ying, Xue. "An overview of overfitting and its solutions." In \textit{Journal of physics: Conference series}, vol. 1168, p. 022022. IOP Publishing, 2019.

\bibitem{wilson2020survey}
Wilson, Garrett, and Diane J. Cook. "A survey of unsupervised deep domain adaptation." \textit{ACM Transactions on Intelligent Systems and Technology (TIST)} 11, no. 5 (2020): 1-46.

\bibitem{patel2015visual}
Patel, Vishal M., Raghuraman Gopalan, Ruonan Li, and Rama Chellappa. "Visual domain adaptation: A survey of recent advances." \textit{IEEE signal processing magazine} 32, no. 3 (2015): 53-69.

\bibitem{tuia2016domain}
Tuia, Devis, Claudio Persello, and Lorenzo Bruzzone. "Domain adaptation for the classification of remote sensing data: An overview of recent advances." \textit{IEEE geoscience and remote sensing magazine} 4, no. 2 (2016): 41-57.

\bibitem{pan2009survey}
Pan, Sinno Jialin, and Qiang Yang. "A survey on transfer learning." \textit{IEEE Transactions on knowledge and data engineering} 22, no. 10 (2009): 1345-1359.

\bibitem{kouw2019review}
Kouw, Wouter M., and Marco Loog. "A review of domain adaptation without target labels." \textit{IEEE transactions on pattern analysis and machine intelligence} 43, no. 3 (2019): 766-785.

\bibitem{ling2016machine}
Ling, Julia, Andrew Kurzawski, and Jeremy Templeton. "Reynolds averaged turbulence modelling using deep neural networks with embedded invariance." \textit{Journal of Fluid Mechanics} 807 (2016): 155-166.

\bibitem{randall2003}
Randall, David, Marat Khairoutdinov, Akio Arakawa, and Wojciech Grabowski. "Breaking the cloud parameterization deadlock." \textit{Bulletin of the American Meteorological Society} 84, no. 11 (2003): 1547-1564.

\bibitem{Bony2015}
Bony, Sandrine, Bjorn Stevens, Dargan MW Frierson, Christian Jakob, Masa Kageyama, Robert Pincus, Theodore G. Shepherd et al. "Clouds, circulation and climate sensitivity." \textit{Nature Geoscience} 8, no. 4 (2015): 261-268.

\bibitem{Schneider2017}
Schneider, Tapio, João Teixeira, Christopher S. Bretherton, Florent Brient, Kyle G. Pressel, Christoph Schär, and A. Pier Siebesma. "Climate goals and computing the future of clouds." \textit{Nature Climate Change} 7, no. 1 (2017): 3-5.

\bibitem{gentine2021deep}
Gentine, Pierre, Veronika Eyring, and Tom Beucler. "Deep learning for the parametrization of subgrid processes in climate models." \textit{Deep Learning for the Earth Sciences: A Comprehensive Approach to Remote Sensing, Climate Science, and Geosciences} (2021): 307-314.

\bibitem{brenowitz2020interpreting}
Brenowitz, Noah D., Tom Beucler, Michael Pritchard, and Christopher S. Bretherton. "Interpreting and stabilizing machine-learning parametrizations of convection." \textit{Journal of the Atmospheric Sciences} 77, no. 12 (2020): 4357-4375.

\bibitem{ott2020fortran}
Ott, Jordan, Mike Pritchard, Natalie Best, Erik Linstead, Milan Curcic, and Pierre Baldi. "A Fortran-Keras deep learning bridge for scientific computing." \textit{Scientific Programming} 2020 (2020): 1-13.

\bibitem{forster2021earth}
Forster, P., T. Storelvmo, K. Armour, W. Collins, J.-L. Dufresne, D. Frame, D.J. Lunt, T. Mauritsen, M.D. Palmer, M. Watanabe, M. Wild, and H. Zhang, 2021: The Earth’s Energy Budget, Climate Feedbacks, and Climate Sensitivity. In \textit{Climate Change 2021: The Physical Science Basis. Contribution of Working Group I to the Sixth Assessment Report of the Intergovernmental Panel on Climate Change} [Masson-Delmotte, V., P. Zhai, A. Pirani, S.L. Connors, C. Péan, S. Berger, N. Caud, Y. Chen, L. Goldfarb, M.I. Gomis, M. Huang, K. Leitzell, E. Lonnoy, J.B.R. Matthews, T.K. Maycock, T. Waterfield, O. Yelekçi, R. Yu, and B. Zhou (eds.)]. Cambridge University Press, Cambridge, United Kingdom and New York, NY, USA, pp. 923–1054.

\bibitem{mooers2021assessing}
Mooers, Griffin, Michael Pritchard, Tom Beucler, Jordan Ott, Galen Yacalis, Pierre Baldi, and Pierre Gentine. "Assessing the potential of deep learning for emulating cloud superparameterization in climate models with real‐geography boundary conditions." \textit{Journal of Advances in Modeling Earth Systems} 13, no. 5 (2021): e2020MS002385.

\bibitem{gentine2018could}
Gentine, Pierre, Mike Pritchard, Stephan Rasp, Gael Reinaudi, and Galen Yacalis. "Could machine learning break the convection parameterization deadlock?." \textit{Geophysical Research Letters} 45, no. 11 (2018): 5742-5751.

\bibitem{yuval2020stable}
Yuval, Janni, and Paul A. O’Gorman. "Stable machine-learning parameterization of subgrid processes for climate modeling at a range of resolutions." \textit{Nature communications} 11, no. 1 (2020): 3295.

\bibitem{yuval2021use}
Yuval, Janni, Paul A. O'Gorman, and Chris N. Hill. "Use of neural networks for stable, accurate and physically consistent parameterization of subgrid atmospheric processes with good performance at reduced precision." \textit{Geophysical Research Letters} 48, no. 6 (2021): e2020GL091363.

\bibitem{khairoutdinov2005simulations}
Khairoutdinov, Marat, David Randall, and Charlotte DeMott. "Simulations of the atmospheric general circulation using a cloud-resolving model as a superparameterization of physical processes." \textit{Journal of the Atmospheric Sciences} 62, no. 7 (2005): 2136-2154.

\bibitem{pritchard2014causal}
Pritchard, Michael S., and Christopher S. Bretherton. "Causal evidence that rotational moisture advection is critical to the superparameterized Madden–Julian oscillation." \textit{Journal of the Atmospheric Sciences} 71, no. 2 (2014): 800-815.

\bibitem{andersen2012moist}
Andersen, Joseph Allan, and Zhiming Kuang. "Moist static energy budget of MJO-like disturbances in the atmosphere of a zonally symmetric aquaplanet." \textit{Journal of Climate} 25, no. 8 (2012): 2782-2804.

\bibitem{khairoutdinov2001cloud}
Khairoutdinov, Marat F., and David A. Randall. "A cloud resolving model as a cloud parameterization in the NCAR Community Climate System Model: Preliminary results." \textit{Geophysical Research Letters} 28, no. 18 (2001): 3617-3620.

\bibitem{beucler2021enforcing}
Beucler, Tom, Michael Pritchard, Stephan Rasp, Jordan Ott, Pierre Baldi, and Pierre Gentine. "Enforcing analytic constraints in neural networks emulating physical systems." \textit{Physical Review Letters} 126, no. 9 (2021): 098302.

\bibitem{wang2015multiscale}
Wang, Minghuai, Vincent E. Larson, Steven Ghan, Mikhail Ovchinnikov, David P. Schanen, Heng Xiao, Xiaohong Liu, Philip Rasch, and Zhun Guo. "A multiscale modeling framework model (superparameterized CAM5) with a higher‐order turbulence closure: Model description and low‐cloud simulations." \textit{Journal of Advances in Modeling Earth Systems} 7, no. 2 (2015): 484-509.

\bibitem{hurrell2008new}
Hurrell, James W., James J. Hack, Dennis Shea, Julie M. Caron, and James Rosinski. "A new sea surface temperature and sea ice boundary dataset for the Community Atmosphere Model." \textit{Journal of Climate} 21, no. 19 (2008): 5145-5153.

\bibitem{Tulich2015}
Tulich, S. N. "A strategy for representing the effects of convective momentum transport in multiscale models: Evaluation using a new superparameterized version of the Weather Research and Forecast model (SP‐WRF)." \textit{Journal of Advances in Modeling Earth Systems} 7, no. 2 (2015): 938-962.

\bibitem{brenowitz2019spatially}
Brenowitz, Noah D., and Christopher S. Bretherton. "Spatially extended tests of a neural network parametrization trained by coarse‐graining." \textit{Journal of Advances in Modeling Earth Systems} 11, no. 8 (2019): 2728-2744.

\bibitem{bolton2019applications}
Bolton, Thomas, and Laure Zanna. "Applications of deep learning to ocean data inference and subgrid parameterization." \textit{Journal of Advances in Modeling Earth Systems} 11, no. 1 (2019): 376-399.

\bibitem{garner2007resolving}
Garner, Steve T., D. M. W. Frierson, I. M. Held, O. Pauluis, and G. K. Vallis. "Resolving convection in a global hypohydrostatic model." \textit{Journal of the Atmospheric Sciences} 64, no. 6 (2007): 2061-2075.

\bibitem{boos2016convective}
Boos, William R., Alexey Fedorov, and Les Muir. "Convective self-aggregation and tropical cyclogenesis under the hypohydrostatic rescaling." \textit{Journal of the Atmospheric Sciences} 73, no. 2 (2016): 525-544.

\bibitem{o2021response}
O’Gorman, Paul Ambrose, Z. Li, William R. Boos, and Janni Yuval. "Response of extreme precipitation to uniform surface warming in quasi-global aquaplanet simulations at high resolution." \textit{Philosophical Transactions of the Royal Society A} 379, no. 2195 (2021): 20190543.

\bibitem{neale2000standard}
Neale, Richard B., and Brian J. Hoskins. "A standard test for AGCMs including their physical parametrizations: I: The proposal." \textit{Atmospheric Science Letters} 1, no. 2 (2000): 101-107.

\bibitem{geron2019hands}
Géron, Aurélien. \textit{Hands-on machine learning with Scikit-Learn, Keras, and TensorFlow.} " O'Reilly Media, Inc.", 2022.

\bibitem{karpatne2018machine}
Karpatne, Anuj, Imme Ebert-Uphoff, Sai Ravela, Hassan Ali Babaie, and Vipin Kumar. "Machine learning for the geosciences: Challenges and opportunities." \textit{IEEE Transactions on Knowledge and Data Engineering} 31, no. 8 (2018): 1544-1554.

\bibitem{holton1973introduction}
Holton, James R. "An introduction to dynamic meteorology." \textit{American Journal of Physics} 41, no. 5 (1973): 752-754.

\bibitem{siebesma2020clouds}
Siebesma, A. Pier, Sandrine Bony, Christian Jakob, and Bjorn Stevens, eds. \textit{Clouds and climate: Climate science's greatest challenge}. Cambridge University Press, 2020.

\bibitem{manabe1967thermal}
Manabe, Syukuro, and Richard T. Wetherald. "Thermal equilibrium of the atmosphere with a given distribution of relative humidity." \textit{Journal of Atmospheric Sciences}, 24, 241–259.

\bibitem{BrownZhang1997}
Brown, Randy G., and Chidong Zhang. "Variability of midtropospheric moisture and its effect on cloud-top height distribution during TOGA COARE." \textit{Journal of the atmospheric sciences} 54, no. 23 (1997): 2760-2774.

\bibitem{BrethertonEA2004}
Bretherton, Christopher S., Matthew E. Peters, and Larissa E. Back. "Relationships between water vapor path and precipitation over the tropical oceans." \textit{Journal of climate} 17, no. 7 (2004): 1517-1528.

\bibitem{HollowayNeelin2009}
Holloway, Christopher E., and J. David Neelin. "Temporal relations of column water vapor and tropical precipitation." \textit{Journal of the Atmospheric Sciences} 67, no. 4 (2010): 1091-1105.

\bibitem{seeley2019fat}
Seeley, Jacob T., Nadir Jeevanjee, and David M. Romps. "FAT or FiTT: Are anvil clouds or the tropopause temperature invariant?." \textit{Geophysical Research Letters} 46, no. 3 (2019): 1842-1850.

\bibitem{hartmann2002important}
Hartmann, Dennis L., and Kristin Larson. "An important constraint on tropical cloud‐climate feedback." \textit{Geophysical research letters} 29, no. 20 (2002): 12-1.

\bibitem{kuang2007testing}
Kuang, Zhiming, and Dennis L. Hartmann. "Testing the fixed anvil temperature hypothesis in a cloud-resolving model." \textit{Journal of Climate} 20, no. 10 (2007): 2051-2057.

\bibitem{AhmedNeelin2018}
Ahmed, Fiaz, and J. David Neelin. "Reverse engineering the tropical precipitation–buoyancy relationship." \textit{Journal of the Atmospheric Sciences} 75, no. 5 (2018): 1587-1608.

\bibitem{AhmedEA2020}
Ahmed, Fiaz, Angel F. Adames, and J. David Neelin. "Deep convective adjustment of temperature and moisture." \textit{Journal of the Atmospheric Sciences} 77, no. 6 (2020): 2163-2186.

\bibitem{SchiroEtal2018}
Schiro, Kathleen A., Fiaz Ahmed, Scott E. Giangrande, and J. David Neelin. "GoAmazon2014/5 campaign points to deep-inflow approach to deep convection across scales." \textit{Proceedings of the National Academy of Sciences} 115, no. 18 (2018): 4577-4582.

\bibitem{AhmedEA2021}
Ahmed, Fiaz, and J. David Neelin. "Protected convection as a metric of dry air influence on precipitation." \textit{Journal of Climate} 34, no. 10 (2021): 3821-3838.

\bibitem{hartmann2015global}
Hartmann, Dennis L. \textit{Global physical climatology}. Vol. 103. Newnes, 2015.

\bibitem{lundberg2017unified}
Lundberg, Scott M., and Su-In Lee. "A unified approach to interpreting model predictions." \textit{Advances in neural information processing systems} 30 (2017).

\bibitem{shapley201617}
Shapley, Lloyd S. \textit{A value for n-person games}. (1953): 307-317.

\bibitem{herman2013linear}
Herman, Michael J., and Zhiming Kuang. "Linear response functions of two convective parameterization schemes." \textit{Journal of Advances in Modeling Earth Systems} 5, no. 3 (2013): 510-541.

\bibitem{beucler2018linear}
Beucler, Tom, Timothy Cronin, and Kerry Emanuel. "A linear response framework for radiative‐convective instability." \textit{Journal of Advances in Modeling Earth Systems} 10, no. 8 (2018): 1924-1951.

\bibitem{kuang2018linear}
Kuang, Zhiming. "Linear stability of moist convecting atmospheres. Part I: From linear response functions to a simple model and applications to convectively coupled waves." \textit{Journal of the Atmospheric Sciences} 75, no. 9 (2018): 2889-2907.

\bibitem{kindermans2019reliability}
Kindermans, Pieter-Jan, Sara Hooker, Julius Adebayo, Maximilian Alber, Kristof T. Schütt, Sven Dähne, Dumitru Erhan, and Been Kim. "The (un) reliability of saliency methods." \textit{Explainable AI: Interpreting, explaining and visualizing deep learning} (2019): 267-280.

\bibitem{HollowayNeelin2007}
Holloway, Christopher E., and J. David Neelin. "The convective cold top and quasi equilibrium." \textit{Journal of the atmospheric sciences} 64, no. 5 (2007): 1467-1487.

\bibitem{zheng2018feature}
Zheng, Alice, and Amanda Casari. \textit{Feature engineering for machine learning: principles and techniques for data scientists}. " O'Reilly Media, Inc.", 2018.

\bibitem{shepherd1990symmetries}
Shepherd, Theodore G. "Symmetries, conservation laws, and Hamiltonian structure in geophysical fluid dynamics." In \textit{Advances in Geophysics}, vol. 32, pp. 287-338. Elsevier, 1990.

\bibitem{beran1977minimum}
Beran, Rudolf. "Minimum Hellinger distance estimates for parametric models." \textit{The annals of Statistics} (1977): 445-463.

\bibitem{endres2003new}
Endres, Dominik Maria, and Johannes E. Schindelin. "A new metric for probability distributions." \textit{IEEE Transactions on Information theory} 49, no. 7 (2003): 1858-1860.


















\end{thebibliography}
\end{document}


\let\oldthetable\thetable
\renewcommand{\thetable}{S\oldthetable}
\let\oldthefigure\thefigure
\renewcommand{\thefigure}{S\oldthefigure}

\begin{center}

\vspace{20pt}

{\Large Supplementary Materials for\\   
\textbf{Climate-Invariant Machine Learning}}\\   
\  \\    

Tom Beucler, Pierre Gentine, Janni Yuval, Ankitesh Gupta, Liran Peng, Jerry Lin, Sungduk Yu, Stephan Rasp, Fiaz Ahmed, Paul A. O'Gorman, J. David Neelin, Nicholas J. Lutsko, Michael Pritchard
\end{center}

\  \\    

\textbf{This PDF file includes:}\\    
\par Supplementary Text (SM A to SM D)
\par Figs. S1 to S15    
\par Tables S1 to S3    




\ \\   
The Supplementary Materials (SM) is organized as follows: First, we discuss code and data availability (SM A), including links to multiple repositories to reproduce the different ML-based closures and climate simulations discussed in the manuscript. Then, we present the characteristics of the emulated mapping (SM B1), derive the input transformations used in the manuscript (SM B2), the output transformations tested in this SM (SM B3), and discuss possible vertical coordinate transformations (SM B4). We provide a guide to find new climate-invariant transformations in SM B5. SM C details the practical implementation of our climate-invariant ML workflow. Finally, we present supplementary results in SM D, including the Hellinger and Jensen-Shannon distances between input distributions (SM D1), the learning curves of climate-invariant models across climates and geographies (SM D2), the generalization skill of climate-invariant NNs near the surface (SM D3), and three methods to visualize the ``raw-data'' and climate-invariant mappings and compare them in the cold (-4K) and warm (+4K) climates (SM D4).

\section*{A. Code and Data Availability}

The code used to process data, train models, and produce this manuscript's figure can be found in the following Github repository: \url{https://github.com/tbeucler/CBRAIN-CAM}, which is archived using Zenodo \url{https://zenodo.org/record/8140413} \cite{stephan_rasp_2023_8140413}. This repository includes a minimal reproducible example on how to train a climate-invariant neural network and verify its improved generalization ability: \url{https://colab.research.google.com/github/tbeucler/CBRAIN-CAM/blob/master/Climate_Invariant_Guide.ipynb} and a notebook to generate all figures in this manuscript requiring model data \url{https://github.com/tbeucler/CBRAIN-CAM/blob/master/notebooks/tbeucler_devlog/090_Climate_Invariant_Paper_Figures_v2.ipynb}. Both scripts rely on the manuscript's accompanying data, archived in the following Zenodo repository: \url{https://doi.org/10.5281/zenodo.8140536} \cite{tom_beucler_2023_8140536}. 

The above Github repository is forked from (and builds upon) Stephan Rasp's CBRAIN repository \url{https://github.com/raspstephan/CBRAIN-CAM}, also archived using Zenodo \url{https://zenodo.org/record/1402384#.YajSg9BKiUk} \cite{stephan_rasp_2018_1402384}. This repository contains a quickstart guide \url{https://github.com/raspstephan/CBRAIN-CAM/blob/master/quickstart.ipynb} to preprocess raw climate model output, train a neural network and benchmark it.

As described in Section~2, we use data from eight climate simulations using three climate models (SPCAM3, SPCESM2, and SAM) to form our training, validation, and test sets. We report the exact characteristics of the splits in Tab~S2 and information to re-generate the full simulation output below.

\paragraph{SPCAM3} 
The codebase for running the ``SPCAM3'' simulation is the same employed by ~\cite{rasp2018deep}, which is archived at ~\url{https://gitlab.com/mspritch/spcam3.0-neural-net/-/tree/sp-diagnostic} for the (+0K) simulation. The sea surface temperature is uniformly cooled by 4K to produce the (-4K) simulation and uniformly warmed by 4K to produce the (+4K) simulation. Raw output of the (+0K) simulation can be found at ~\url{https://zenodo.org/record/1402384#.YaUCsdDMI-w} \cite{stephan_rasp_2018_1402384}. The full simulations output, which is several TB, is archived on the GreenPlanet cluster at UC Irvine and available upon request.

\paragraph{SPCESM2}
The codebase for running the ``SPCESM2'' simulations is the same employed by~\cite{parishani2018insensitivity}, which is archived at~\url{https://github.com/mspritch/UltraCAM-spcam2_0_cesm1_1_1}; this code was in turn forked from a development version of the CESM1.1.1 located on the NCAR central subversion repository under tag~\url{spcam_cam5_2_00_forCESM1_1_1Rel_V09}, which dates to February 25, 2013. The full simulations output, which is several TB, is also archived on the GreenPlanet cluster at UC Irvine and available upon request. We additionally archived the input data and run scripts necesary to re-run all three simulations as part of the manuscript’s accompanying data using Zenodo \url{https://zenodo.org/record/5775541#.YbeMHNDMKUl} \cite{tom_beucler_2023_8140536}.

\paragraph{SAM}
The codebase for running the ``SAM'' simulations is the same employed by \cite{yuval2020stable}. %
The initial sounding, meridional surface temperature profile, and source code to re-run the simulation can be found at ~\url{https://zenodo.org/record/4118346#.YaT_WtBKg-w} \cite{janni_yuval_2020_4118346}. To produce the (+0K) simulation, the initial sounding and surface temperature profiles are both uniformly warmed by 4K. The output from the SAM simulations, which is several TB, is archived at MIT and is available upon request.

\section*{B. Feature Transformations}

\subsection*{B1. Mapping}

In this subsection, we present the characteristics of the emulated mapping, highlighting the differences between the superparameterized models (SPCAM3, SPCESM2) and the storm-resolving model (SAM). In both cases, the input vector $\boldsymbol{x} $ encodes the large-scale ($\approx$100km) climate state:  

\begin{equation}
\boldsymbol{x}=\begin{cases}
\left[\begin{array}{cccccc}
\boldsymbol{q\left(p\right)} & \boldsymbol{T\left(p\right)} & p_{s} & S_{0} & \mathrm{SHF} & \mathrm{LHF}\end{array}\right]^{T} & \left(\mathrm{SPCAM}3,\mathrm{SPCESM}2\right)\\
\left[\begin{array}{ccc}
\boldsymbol{q_{\mathrm{np}}\left(p\right)} & \boldsymbol{T\left(p\right)} & d_{\mathrm{Equator}}\end{array}\right]^{T} & \left(\mathrm{SAM}\right)
\end{cases}
\end{equation}
where $\boldsymbol{q \left(p\right)} $ is the vertical profile of specific humidity in units of kg/kg (written as a function of the background pressure coordinate $\boldsymbol{p} $ in units of Pa), $\boldsymbol{T \left(p\right)} $ is the temperature's vertical profile in units of K, $p_{s} $ is surface pressure in units Pa, $S_{0} $ is solar insolation in units of $\mathrm{W/m^{2}}$, SHF is surface sensible heat flux in units $\mathrm{W/m^{2}}$, LHF is surface latent heat flux in units of $\mathrm{W/m^{2}}$, $\boldsymbol{q_{\mathrm{np}}\left(p\right)} $ is the vertical profile of non-precipitating water concentration in units of kg/kg, and $d_{\mathrm{Equator}} $ is the distance to the Equator, which is used as a proxy for solar insolation in the mapping learned for SAM. The output vector $\boldsymbol{y} $ groups subgrid-scale thermodynamic tendencies: 

\begin{equation}
\boldsymbol{y}=\begin{cases}
\left[\begin{array}{cccc}
L_{v}\boldsymbol{\dot{q}\left(p\right)} & c_{p}\boldsymbol{\dot{T}\left(p\right)} & c_{p}\boldsymbol{\mathrm{lw}\left(p\right)} & c_{p}\boldsymbol{\mathrm{sw}\left(p\right)}\end{array}\right]^{T} & \left(\mathrm{SPCAM}3,\mathrm{SPCESM}2\right)\\
\left[\begin{array}{cc}
L_{v}\boldsymbol{\dot{q}_{\mathrm{np}}\left(p\right)} & \boldsymbol{\dot{H}_L\left(p\right)}\end{array}\right]^{T} & \left(\mathrm{SAM}\right)
\end{cases}
\end{equation}
where $L_{v} $ in units of $\mathrm{J\ kg^{-1}}$ is the latent heat of vaporization of water in standard conditions, $\boldsymbol{\dot{q} \left(p\right)} $ is the subgrid moistening vertical profile, $c_{p} $ in units of $\mathrm{J\ kg^{-1}\ K^{-1}}$ is the specific heat of dry air at constant pressure in standard atmospheric conditions, $\boldsymbol{\dot{T} \left(p\right)} $ is the total subgrid heating vertical profile (including subgrid  radiation), $\boldsymbol{\mathrm{lw} \left(p\right)} $ is the subgrid longwave radiative heating vertical profile, $\boldsymbol{\mathrm{sw} \left(p\right)} $ is the subgrid shortwave radiative heating vertical profile, $\boldsymbol{\dot{q}_{\mathrm{np}}\left(p\right)} $ is the non-precipitating water condensation vertical profile, and $\boldsymbol{\dot{H}_L\left(p\right)} $ is the subgrid time-tendency of the liquid/ice static energy vertical profile. Following \cite{beucler2021enforcing}, all components of the output vector $\boldsymbol{y} $ are mass-weighted and vertically integrated within each vertical layer to yield energy flux units ($\mathrm{W/m^{2}}$). Assuming vertical profiles have $N_{p} $ vertical levels, $\boldsymbol{x} $ is of length $\left(2N_{p}+4\right) $ for SPCAM3 and SPCESM2, and of length $\left(2N_{p}+1\right) $ for SAM. $\boldsymbol{y} $ is of length $4N_{p} $ for SPCAM3 and SPCESM2, and of length $2N_{p} $ for SAM.

\subsection*{B2. Physically-Based Input Transformations}

In Fig~S2, we compare three transformation options for each input, whose univariate PDFs are depicted in Fig~S2: No transformation (top), our most successful transformation (bottom), and our second best transformation (middle). After defining our second best input transformations (B2a), we delve into the details of our relative humidity (B2b) and plume buoyancy (B2c) transformations before discussing the other inputs' distribution shift (B2d). 

\subsubsection*{B2a. Second Best Input Transformations}

Along the way to our optimal feature transformations, we explored candidate options that proved second best. For completeness these are reviewed first in the SM.

\paragraph{Saturation Deficit\label{subsub:q}:} We explored saturation deficit but found it did not lead to climate invariance. Similar to $\boldsymbol{q} $, saturation deficit (Fig~S2a, middle) still has a corresponding expansion of the PDF with warming as a result of the Clausius-Clapeyron relation. It is defined as the amount by which the water vapor concentration must be increased to achieve saturation without changing the environmental temperature and pressure:

\begin{equation}
    \boldsymbol{\tilde{q}_{\mathrm{deficit}}\left(T,p\right)}\overset{\mathrm{def}}{=}\boldsymbol{q_{\mathrm{sat}}\left(T,p\right)}-\boldsymbol{q},
\end{equation}
where $\boldsymbol{q_{\mathrm{sat}}\left(T,p\right)} $ is the saturation specific humidity. 

In contrast, the relative humidity transformation $\boldsymbol{\tilde{q}_{\mathrm{RH}}\left(p\right)}\overset{\mathrm{def}}{=}\boldsymbol{\mathrm{RH}\left(q,T,p\right)} $ (Fig~S2a, bottom) results in a climate-invariant PDF, as evidenced by PDFs that mostly overlap across all three climates. 

\paragraph{Temperature minus Near-Surface Temperature\label{subsub:T}:} Assuming that the temperature's PDF shift with warming is almost uniform with height, we can derive an approximate invariant by subtracting the temperature at all levels $\boldsymbol{T\left(p\right)} $ from the near-surface temperature $T\left(p_{\mathrm{NS}}\right) $: 

\begin{equation}
    \boldsymbol{\tilde{T}_{\mathrm{from\ NS}}\left(p\right)}\overset{\mathrm{def}}{=}T\left(p_{\mathrm{NS}}\right)-\boldsymbol{T\left(p\right)},
\end{equation}
where $p_{\mathrm{NS}} $ is the lowest atmospheric pressure level see (Fig~S2b, middle). However, this linear transformation fails in the upper atmosphere, especially near the tropopause where temperatures are approximately invariant with warming \cite{seeley2019fat,hartmann2002important,kuang2007testing} and therefore decoupled from surface temperature changes. This is why the buoyancy of a moist static energy-conserving plume: 
\begin{equation}
\boldsymbol{\tilde{T}_{\mathrm{buoyancy}}\left(p\right)}\overset{\mathrm{def}}{=}\boldsymbol{B_{\mathrm{plume}}}\left(q_\mathrm{NS},\boldsymbol{T,p}\right),
\end{equation}
where $q_\mathrm{NS}=q(p_\mathrm{NS})$ is the near-surface specific humidity, yields approximate climate invariance (Fig~S2b, bottom), unlike the temperature minus near surface temperature transformation.

\paragraph{Scaling Latent Heat Fluxes by Near-Surface Specific Humidity\label{subsub:LHF}:} To address the increase of LHF with warming, we scale LHF by near-surface specific humidity $q\left(p_{\mathrm{NS}}\right) $ (Fig~S2c, middle):

\begin{equation}
    \tilde{\mathrm{LHF}}_{q}\overset{\mathrm{def}}{=}\frac{\mathrm{LHF}}{L_{v}\max\left\{ \epsilon_{q},q\left(p_{\mathrm{NS}}\right)\right\} },
\end{equation}
where $\epsilon_{q} $ is a user-chosen parameter that we  set to $10^{-4} $ to avoid division by zero. While better than directly using LHF, this transformation fails for very dry atmospheres when the latent heat flux is negative, e.g. in polar oceans where atmospheric water vapor may be condensing on the surface, or when the near-surface specific humidity is very small, e.g. in subtropical regions. This is why scaling LHF using the near-surface saturation deficit (Fig~S2c, bottom):

\begin{equation}
\tilde{\mathrm{LHF}}_{\Delta q}\overset{\mathrm{def}}{=}\frac{\mathrm{LHF}}{L_{v}\max\left\{ \epsilon_{q},q_{\mathrm{sat}}\left[T\left(p_{\mathrm{NS}}\right),p_{\mathrm{NS}}\right]-q\left(p_{\mathrm{NS}}\right)\right\} }.
\label{eq:LHF_nsdelQ}
\end{equation}
yields better generalizability. While both the Jensen-Shannon and Hellinger distances would suggest that $\tilde{\mathrm{LHF}}_{\Delta q} $ is a slightly less good transformation than $\tilde{\mathrm{LHF}}_{q} $, the $\tilde{\mathrm{LHF}}_{\Delta q} $ transformation leads to improved generalization performance compared to $\tilde{\mathrm{LHF}}_{q} $ (not shown). This confirms that only considering the PDF distances is not always sufficient to find the optimal transformations (discussed in SM B4).

\subsubsection*{B2b. Relative Humidity}

Relative humidity (RH) provides our optimal transformation for the specific humidity inputs. RH is defined as the ratio of the partial pressure of water vapor $\boldsymbol{e}\boldsymbol{\left(p,q\right)}\ $to its saturation value $\boldsymbol{e_{\mathrm{sat}}\left(T\right)}$, and can be expressed analytically: 
\begin{equation}
\boldsymbol{\mathrm{RH}}\overset{\mathrm{def}}{=}\frac{\boldsymbol{e}\boldsymbol{\left(p,q\right)}}{\boldsymbol{e_{\mathrm{sat}}\left(T\right)}}
\overset{q\ll1}{\approx}
\frac{R_{v}}{R_{d}}\frac{\boldsymbol{p}\boldsymbol{q}}{\boldsymbol{e_{\mathrm{sat}}\left(T\right)}},\label{eq:RH_def}
\end{equation}
where $R_{v}\approx461\mathrm{J\ kg^{-1}\ K^{-1}}\ $is the specific gas constant for water vapor, $R_{d}\approx287\mathrm{J\ kg^{-1}\ K^{-1}}\ $is the specific gas constant for dry air, $\boldsymbol{p}\ $(in units Pa) is the total atmospheric pressure, $\boldsymbol{q}\ $(in
units kg/kg) is specific humidity, and $\boldsymbol{e_{\mathrm{sat}}\left(T\right)}\ $(in units Pa) is the saturation pressure of water vapor, whose analytic expression in our case is given below. Consistent with Eq~\ref{eq:RH_def}, the saturation specific humidity $\boldsymbol{q_{\mathrm{sat}}} $ corresponding to $\mathrm{RH}=1$, is
\begin{equation}
    \boldsymbol{q_{\mathrm{sat}}\left(T,p\right)}=\frac{R_{d}\boldsymbol{e_{\mathrm{sat}}\left(T\right)}}{R_{v}\boldsymbol{p}}.
\end{equation}

SAM's single-moment microphysics scheme \cite{khairoutdinov2003cloud}, which is also used in the SPCAM3 and SPCESM2 simulations, partitions water between the liquid and ice phases using a weight $\omega\ $ that is a linear function of the absolute temperature:
\begin{equation}  
\boldsymbol{\omega}\overset{\mathrm{def}}{=}\frac{\boldsymbol{T}-T_{00}}{T_{0}-T_{00}}.
\end{equation} 

Under the assumptions of this microphysics scheme, the saturation pressure of water vapor can then be found by integrating the Clausius-Clapeyron equation with respect to temperature,  expressed analytically as:

\begin{equation}
\boldsymbol{e_{\mathrm{sat}}\left(T\right)}=\begin{cases}
\boldsymbol{e_{\mathrm{liq}}\left(T\right)} & \boldsymbol{T}>T_{0}\\
\boldsymbol{e_{\mathrm{ice}}\left(T\right)} & \boldsymbol{T}<T_{00}\\
\boldsymbol{\omega e_{\mathrm{liq}}\left(T\right)}+\boldsymbol{\left(1-\omega\right)e_{\mathrm{ice}}\left(T\right)} & \boldsymbol{T}\in\left[T_{00},T_{0}\right]
\end{cases},\label{eq:Sat_wat_vap_pressure}
\end{equation}
where $T_{0}=273.16\mathrm{K} $ and $T_{00}=253.16\mathrm{K} $. In Eq~\ref{eq:Sat_wat_vap_pressure}, as temperature increases, the saturation pressure of water vapor goes from the saturation vapor pressure with respect to liquid $\boldsymbol{e_{\mathrm{liq}}}$, to the saturation vapor pressure with respect to ice $\boldsymbol{e_{\mathrm{ice}}}$.
These are given by the following polynomial approximations:
\begin{equation}
\boldsymbol{e_{\mathrm{liq}}\left(T\right)}=100\mathrm{Pa}\times\sum_{i=0}^{8}a_{\mathrm{liq,i}}\left[\max\left(-193.15\mathrm{K},T-T_{0}\right)\right]^{i},
\end{equation}
where $\boldsymbol{a_{\mathrm{liq}}}\ $is a vector of length 9 containing nonzero polynomial coefficients.   The polynomial approximation for $\boldsymbol{e_{\mathrm{ice}}}$, with the same temperature switches as Eq~\ref{eq:Sat_wat_vap_pressure}, is:
\begin{equation}
\boldsymbol{e_{\mathrm{ice}}\left(T\right)}=\begin{cases}
\boldsymbol{e_{\mathrm{liq}}\left(T\right)}\\
100\mathrm{Pa}\times\left\{ c_{\mathrm{ice},1}+\boldsymbol{{\cal C}\left(T\right)}\left[c_{\mathrm{ice},4}+c_{\mathrm{ice},5}\boldsymbol{{\cal C}\left(T\right)}\right]\right\}\\
100\mathrm{Pa}\times\sum_{i=0}^{8}a_{\mathrm{ice,i}}\left(\boldsymbol{T}-T_{0}\right)^{i}
\end{cases},
\end{equation}  
where $\boldsymbol{{\cal C}\left(T\right)}\ $is a ramp function of temperature given by:
\begin{equation}  
\boldsymbol{{\cal C}\left(T\right)}\overset{\mathrm{def}}{=}\max\left(c_{\mathrm{ice},2},\boldsymbol{T}-T_{0}\right),
\end{equation}  
and $\left(\boldsymbol{a_{\mathrm{ice}}},\boldsymbol{c_{\mathrm{ice}}}\right)\ $are vectors of length 9 and 5 containing nonzero elements, respectively. Between temperatures of $T_{00}\ $and $T_{0}$, the saturation pressure of water vapor is a weighted mean of $\boldsymbol{e_{\mathrm{liq}}}\ $and $\boldsymbol{e_{\mathrm{ice}}}$. The reader interested in the numerical details of this transformation is referred to our implementation of relative humidity at \url{https://colab.research.google.com/github/tbeucler/CBRAIN-CAM/blob/master/Climate_Invariant_Guide.ipynb}.

\subsubsection*{B2c. Plume Buoyancy}

Plume Buoyancy ($B_{\mathrm{plume}} $) is our most successful transformation for the temperature inputs. Buoyancy is defined as the upward acceleration exerted upon parcels by virtue of the density difference between the parcel and the surrounding air of the atmospheric column (e.g., \cite{emanuel1994atmospheric}). Because our ML model's inputs represent the large-scale thermodynamic state, the ML model does not have information about the storm-scale buoyancy field, and we must rely on idealized approximations to estimate the buoyancy that a plume would have for given specific humidity and temperature profiles. To be consistent with the model's conserved quantities \cite{khairoutdinov2003cloud}, we derive a simple buoyancy metric based on a moist static energy ($h $) conserving plume below following similar derivations in \cite{AhmedEA2020} and \cite{AdamesEA2021}. We refer the reader interested in the numerical details of this transformation to \url{https://colab.research.google.com/github/tbeucler/CBRAIN-CAM/blob/master/Climate_Invariant_Guide.ipynb}.

For purposes of this transformation, we omit virtual temperature effects and condensate loading (effects of the environmental water vapor on heating/moisture sink are being estimated separately). Thus the parcel buoyancy is simply proportional to the relative difference between its temperature $T_{\mathrm{par}} $ and the environmental temperature $T $:

\begin{equation}
\boldsymbol{B_{\mathrm{plume}}}\approx g\frac{T_{\mathrm{par}}-\boldsymbol{T}}{\boldsymbol{T}},
\label{eq:bpar}
\end{equation}
where $g $ is the gravity constant. Further assuming that the plume is non-entraining, obeys hydrostatic balance, and lifts parcels from the near-surface, the lifted parcel's moist static energy is conserved and equal to its near-surface value (at pressure $p_{\mathrm{NS}}$:

\begin{equation}
\begin{aligned}
h_\mathrm{{par}} & \approx h\left(p_{\mathrm{NS}}\right)\\
 & \overset{\mathrm{def}}{=}L_{v}q\left(p_{\mathrm{NS}}\right)+c_{p}T\left(p_{\mathrm{NS}}\right),
\end{aligned}
\label{eq:MSE_par}
\end{equation}
where we have used the environmental moist static energy's definition:

\begin{equation}
\boldsymbol{h}\overset{\mathrm{def}}{=}L_{v}\boldsymbol{q}+c_{p}\boldsymbol{T}+g\boldsymbol{z},    
\end{equation}
where $\boldsymbol{z} $ is geopotential height, and we neglected $z\left(p_{\mathrm{NS}}\right) $ as the near-surface is close to the surface by definition. To express the parcel's buoyancy as a function of the environmental thermodynamic state, we finally assume that the parcel is saturated (not necessarily true close to the surface), and that the thermodynamic differences between the parcel and the environment are small, which allows us to linearize the Clausius-Clapeyron equation about the environmental temperature:

\begin{equation}
\begin{aligned}
T_{\mathrm{par}}-\boldsymbol{T} & \overset{\mathrm{Claus.-Clap.}}{\approx}\left(\frac{\partial T}{\partial q^{*}}\right)_{\boldsymbol{T,p}}\left(q_{\mathrm{sat,par}}-\boldsymbol{q_{\mathrm{sat}}}\right)\\
 & =\frac{R_{v}\boldsymbol{T}^{2}}{L_{v}\boldsymbol{q_{\mathrm{sat}}}}\left(q_{\mathrm{sat,par}}-\boldsymbol{q_{\mathrm{sat}}}\right)\\
 & =\frac{R_{v}\boldsymbol{T}^{2}}{L_{v}^{2}\boldsymbol{q_{\mathrm{sat}}}}\left[h_{\mathrm{par}}-\boldsymbol{h_{\mathrm{sat}}}-c_{p}\left(T_{\mathrm{par}}-\boldsymbol{T}\right)\right].\label{eq:Tpar-T}
\end{aligned}
\end{equation}

Using Eq~\ref{eq:Tpar-T} to write $T_{\mathrm{par}}-\boldsymbol{T} $ as a function of the environmental thermodynamic state and substituting the resulting expression into Eq~\ref{eq:bpar} yields an estimation of plume buoyancy from $\boldsymbol{\left(q,T,p\right)} $:

\begin{equation}
    \boldsymbol{B_{\mathrm{plume}}}\boldsymbol{\left(q,T,p\right)}=\frac{g\left[h_{\mathrm{par}}-\boldsymbol{h_{\mathrm{sat}}} \left(\boldsymbol{q,T,p}\right)\right]}{\boldsymbol{\kappa\left(T,p\right)}\times c_{p}\boldsymbol{T}},
\end{equation}
where the parcel's moist static energy is expressed as a function of near-surface $\left(q,T\right) $ in Eq~\ref{eq:MSE_par}, the environmental saturated moist static energy in pressure coordinates is defined as:

\begin{equation}
    \boldsymbol{h_{\mathrm{sat}}\left(q,T,p\right)} \overset{\mathrm{def}}{=}L_{v}
\boldsymbol{q_{\mathrm{sat}}\left(T,p\right)}+c_{p}\boldsymbol{T}+g\boldsymbol{z\left(q,T,p\right)},
\end{equation}

and we have introduced the dimensionless factor:

\begin{equation}
    \boldsymbol{\kappa\left(T,p\right)}=1+\frac{L_{v}^{2}\boldsymbol{q_{\mathrm{sat}}\left(T,p\right)}}{R_{v}c_{p}\boldsymbol{T}^{2}}.
\end{equation}

Note that in pressure coordinates, we calculate the geopotential height by vertically integrating the hydrostatic equation after using the ideal gas law:

\begin{equation}
    \boldsymbol{z\left(q,T,p\right)}=\int_{p}^{p_{\mathrm{NS}}}dp^{\prime}\frac{T\left(p^{\prime}\right)}{gp^{\prime}}\left\{ R_{d}+\left[R_{v}-R_{d}\right]q\left(p^{\prime}\right)\right\} .
\end{equation}

\subsubsection*{B2d. Sensible Heat Fluxes and Surface Pressure \label{ap:SHF-pS}}

In this appendix, we discuss the univariate PDFs of the two inputs we did \textit{not} transform in the main manuscript (see Section~3) in Fig~S3 for both super-parameterized models and all three surface temperatures (-4K, +0K, and +4K). The PDF of sensible heat fluxes changes very little with warming. There is a slight expansion of the left tail of the surface pressure PDF with warming as the most extreme low-pressure systems become more intense, but we hypothesize that these changes are small enough not to require a dedicated input transformation.

\subsection*{B3. Output Transformation}

\subsubsection*{B3a. Theory}

In contrast to input transformation, transforming our ML models' outputs, namely subgrid thermodynamics, only marginally improves the models' ability to generalize. In the absence of physical theory on how the full vertical profile of subgrid thermodynamics changes with warming, we place ourselves in an idealized scenario: 

Assuming we know how the outputs' marginal PDF changes with warming, can we help our ML models generalize via output transformation?

We note that assuming knowledge of how the \textit{marginal} (univariate) PDFs (or equivalently the CDF) of convective heating and moistening change with warming is more realistic than assuming full knowledge of how their \textit{joint} PDFs change with warming. This knowledge could come from e.g. convection theory (e.g., \cite{siebesma2020clouds}) or shorter simulations than those required to train a subgrid closure. Under this assumption, a natural transformation is the outputs' cumulative distribution function (CDF):

\begin{equation}
    \boldsymbol{\tilde{y}}=\boldsymbol{\mathrm{CDF}\left(y\right)}.
\end{equation}

In essence, we are assuming that the mapping is more likely to be invariant in quantile than in physical space, which is a common practice when debiasing the outputs of climate models referred to as \textit{quantile mapping} (e.g., review by \cite{maraun2016bias}). In practice, we test two distinct methods for transforming the outputs using their CDFs and report the results for SPCAM3 aquaplanet simulations in SM B2b. 

\paragraph{Quantile mapping after training:} The first method is to transform the ML model's input \textit{during} training, and then transform the ML model's output \textit{after} training. This is akin to standard, post-hoc, quantile mapping. In the particular case of trying to generalize from a (-4K) cold simulation to a (+4K) warm simulation, the entire transformation to yield outputs in physical units can be mathematically written as:

\begin{equation}
    y\mapsto\mathrm{CDF}_{+4\mathrm{K}}^{-1}\left[\mathrm{CDF}_{\mathrm{-4K}}\left(y\right)\right],
\end{equation}
where for simplicity but without loss of generality, we have considered a singular input $y $ whose CDF is $\mathrm{CDF}_{\mathrm{-4K}} $ in the (-4K) cold simulation and $\mathrm{CDF}_{\mathrm{+4K}} $ in the (+4K) warm simulation. 

\paragraph{Quantile mapping before training:} The second method is to transform the ML model's output \textit{before} training. In that case, we directly train the ML model to predict $\tilde{y}=\mathrm{CDF}\left(y\right) $ as accurately as possible. We then map the output back to physical units using $\mathrm{CDF}^{-1} $ \textit{after} training. 

\subsubsection*{B3b. Results}

The two methods to transform outputs presented above are depicted in Fig~S4 and the generalization results presented in Fig~S5. Transforming outputs \textit{after} training slightly improves generalization skill (from an overall $R^{2}$ of 0.58 to 0.62 for the generalization (+4K) set). In contrast, transforming outputs \textit{before} training leads to equally bad results both on the training and generalization sets, which is a negative result underlining the challenges of designing the appropriate loss function in probability space. A possible solution would be to convert back the outputs to physical space before feeding them to the loss function during training, and further investigation is required to fully assess the potential and limitations of training these ML models in probability space.

\subsection*{B4. Spatial Coordinate Transformation \label{sub:Vert_coord_rescaling}}

Another transformation to consider when input/output variables are functions of spatiotemporal coordinates is \textit{coordinate transformation}, resulting in a coordinate change.  In our specific example, it is possible to transform the vertical coordinate, i.e. the hybrid pressure coordinate $p $. Possible transformations include:
\begin{enumerate}
    \item the temperature ($\widetilde{p}=T $), e.g. for radiative heating, which tends to vary less in temperature coordinates \cite{jeevanjee2018mean},
    \item the saturation specific humidity ($\widetilde{p}=q_{\mathrm{sat}}$) which is consistent with a transformation of the primitive equations that captures an upward shift of the circulation as the climate warms \cite{singh2012upward},
    \item the geopotential height or the altitude ($\widetilde{p}=z $), which could more consistently capture gravity wave propagation,
    \item or a coordinate with fixed values for characteristic vertical levels in the atmosphere, such as the top of the planetary boundary layer or the tropopause \cite{seeley2019fat}.
\end{enumerate}

While we were able to transform the vertical coordinate using interpolation functions (not shown), the benefits were not visible in our particular case. This could be because input transformation already addresses some of the upward shift of convective activity warming, as shown by the influence of humidity inputs in Fig~S14.

\section*{C. Implementation\label{sec:Implementation}}

For reproducibility purposes \cite{irving2016minimum}, we now detail the practical implementation of a climate-invariant ML workflow (see Fig~9), from its overall structure (SM C1) to its benchmarking (SM C4) via the characteristics of the multiple linear regressions (MLR, SM C2) and neural networks (NN, SM C3) presented in this manuscript.

\subsection*{C1. Overall Workflow\label{sub:Overall_workflow}}

We present three ways to implement physical transformations. The first way is to physically transform the inputs/outputs \textit{before} training. While this option is easiest to implement and debug, it usually comes at the cost of disk space: Every time we try a new transformation, we need to duplicate our training/validation/test datasets for all the climates/geographies we are interested in, which can quickly be prohibitive when trying multiple transformation combinations.

Therefore, it can be advantageous to transform the input/output variables within the ML framework, so that the transformations occur \textit{during} training. In essence, we are trading disk space for computational time. In that spirit, the second method is to transform the inputs/outputs via custom layers (e.g., Ch~12 of \cite{geron2019hands}) in the ML algorithm itself. Since this second method tends to substantially slow down training as it adds sequential operations on the GPU, we take advantage of the fact that the transformations occur before and after the emulated mapping, and propose a third method that can happel in parallel on the CPU: Transforming inputs/outputs by customizing the pipeline or ``data generator'', which is the algorithm responsible for feeding numbers to the ML model after reading the training data files. For each batch, the \textit{custom data generator} then transforms inputs \textit{before} feeding them to the ML algorithm. In our case, note that we transform outputs independently via quantile mapping (see SM B2).

For the rest of this manuscript, we will train our ML models using custom data generators: For ``raw-data'' models, the transformations are set to None (no transformation), while for ``climate-invariant'' models, the $q $ transformation is set to $\tilde{q}_{\mathrm{RH}} $, the $T $ transformation is set to $\tilde{T}_{\mathrm{buoyancy}} $, and the LHF transformation is set to $\tilde{\mathrm{LHF}}_{\Delta q} $. For all models, we additionally subtract the mean from each input before dividing it by its range to feed the ML algorithm floating-point numbers between (-1) and 1. Note that for each transformation, numbers are ``de-normalized'' before the transformation and ``re-normalized'' after following the normalization procedure described in Sec~2.4. Therefore, all transformations are done in physical units while the ML algorithm is always fed single-precision floating-point numbers in $\left[-1,1\right]$.

For simplicity and building upon previous ML-powered subgrid closures \cite{rasp2018deep,brenowitz2019spatially,yuval2020stable}, we use the mean-squared error (MSE) of the prediction \textit{in physical units} (here W$^{2}$m$^{-4}$) as our loss function. Motivated by the framework presented in Fig~9, we first train MLRs (SM C2) before training NNs (SM C3) and benchmarking our ML models to quantify their accuracy and ability to generalize (SM C4).

\subsection*{C2. Multiple Linear Regressions\label{sub:MLR}}

To use the same data generator for both MLRs and NNs, we implement our MLRs in Tensorflow 2.0 \cite{tensorflow2015-whitepaper} and train them using the Adam optimizer, which builds on stochastic gradient descent \cite{kiefer1952stochastic}. Training a climate-invariant MLR results in a weight matrix $\boldsymbol{A} $ of size $4N_{p}\times\left(2N_{p}+4\right) $ and a bias vector $\boldsymbol{b} $ of length $4N_{p} $ such that:

\begin{equation}
    \boldsymbol{y}\approx\boldsymbol{A}\boldsymbol{\tilde{x}}+\boldsymbol{b},
    \label{eq:MLR}
\end{equation}
where stochastic optimization means that there is no unique optimal solution for $\boldsymbol{A} $ and $\boldsymbol{b} $. We train MLRs for 20 epochs using the default Keras learning rate of 0.001 and save the weights and biases corresponding to the minimal loss over the validation set.

\subsection*{C3. Neural Network Design\label{sub:NN}}

To isolate the effects of physically transforming the NN's inputs, we fix the hyperparameters of all NNs trained in this study, and leave the joint investigation of hyperparameter tuning and physical transformations for future work. Informed by \cite{ott2020fortran} and \cite{mooers2021assessing}, we fix the architecture to a multilayer perceptron of 7 layers of 128 neurons separated by Leaky Rectified Linear Unit activation functions of slope 0.3, resulting in 122,872 trainable parameters for each NN. We implement the SPCAM NNs using Tensorflow 2.0 \cite{tensorflow2015-whitepaper}, train them for 20 epochs using the Adam optimizer with the default Keras learning rate of 0.001 and a default batch size of 1024, and save the parameters corresponding to the minimal validation loss. 

Following the supplemental material (Sec~2) of \cite{yuval2021use}, some of the hyperparameters used for the NNs trained on SAM data are different. The SAM NNs are implemented using PyTorch 1.4.0 \cite{NEURIPS2019_9015}, have 5 dense layers of 128 neurons each, and use cyclic learning rate \cite{smith2017cyclical}: Starting with an initial learning rate in $\left[2\ \times 10^{-4},2\ \times 10^{-3}\right]$ for the first epoch out of 10, we then reduce the minimal and maximal learning rates by 10$\% $ for the next 6 epochs before further reducing them by a factor 10 for the last 3 epochs. 

For SPCAM and following \cite{molina2021benchmark}, we augment some of our NNs with BN and DP layers, more specifically one DP layer before each activation function and a single BN layer before the first DP layer. Following \cite{srivastava2014dropout}, we use the default DP rate of 30\% and the default parameters of the Keras BN layer that normalize each feature using its mean and standard deviation in a given batch \cite{ioffe2015batch}. Note that we do not adjust the default parameters of DP and BN to optimize \textit{generalization} skill as this would require misusing the generalization test set as a validation test.

\subsection*{C4. Benchmarking\label{sub:Benchmarking}}

We benchmark our ML models using two different metrics: their MSEs and their coefficient of determination $R^{2} $, defined for a singular output $y_{k} $ as:
\begin{equation}
    R^{2}\overset{\mathrm{def}}{=}1-\frac{\left\langle y_{\mathrm{Err,k}}^{2}\right\rangle _{\mathrm{samp}}}{\left\langle \left(y_{\mathrm{Truth,k}}-\left\langle y_{\mathrm{Truth,k}}\right\rangle _{\mathrm{samp}}\right)^{2}\right\rangle _{\mathrm{samp}}},
\end{equation}
where $\left\langle \cdot\right\rangle _{\mathrm{samp}} $ is the averaging operator over the samples of interest. For instance, if we want a horizontal map of $R^{2} $, we average samples at a given location over time, while we average over time and horizontal space if we want a single $R^{2} $ value for $y_{k} $. Similarly, if we want one value of MSE per output $y_{k} $, we only average the MSE over time and horizontal space rather than over all outputs, as when calculating the loss function.

While comparing MSE and $R^{2} $ in the reference and target generalization climates is enough to assess generalization skill \textit{after} training, we are also interested in how a given ML model learns to generalize \textit{during} training. To address that question, we augment our SPCAM ML models with a function (technically a ``Keras callback'' \cite{chollet2018keras}) that calculates the MSE over two datasets that correspond to the two generalization experiments at the end of each epoch: (1) a dataset of different temperature (warm when training on cold, and vice-versa); and (2) a dataset of different geography (Earth-like when training on Aquaplanet, and vice-versa). At the end of training, we hence obtain three learning curves for each ML model: the validation loss, and the loss in the two generalization sets as a function of number of epochs. Note that these callbacks are computationally expensive as they require evaluating the ML model over $\approx 100M$ samples at the end of each epoch, which means they should be avoided when purely seeking performance, e.g. during hyperparameter tuning.

\section*{D. Supplementary Results}

\subsection*{D1. Jensen-Shannon Distance between PDFs across Climates}

As an alternative to the Hellinger PDF distance, we pick the Jensen-Shannon distance \cite{endres2003new} because it is a symmetric distance (i.e., the arguments' order does not affect the outcome) that uses the logarithms of the PDFs, hence giving large weights to the PDFs' tails that tend to be particularly problematic for generalization purposes:

\begin{equation}
    \mathrm{JS}\left(p,q\right)\overset{\mathrm{def}}{=}\sqrt{\frac{\mathrm{KL}\left(p,q\right)+\mathrm{KL}\left(q,p\right)}{2}},
\end{equation}

where $p $ and $q $ are the normalized PDFs to compare and KL is the Kullback–Leibler divergence, defined for continuous PDFs as:

\begin{equation}
    \mathrm{KL}\left(p,q\right)\overset{\mathrm{def}}{=}\int_{0}^{1}dx\times p\left(x\right)\ln\left[\frac{p\left(x\right)}{q\left(x\right)}\right].
\end{equation}

\subsection*{D2. Learning across Climates and Geographies}

This section complements Sec~4.2 and confirms that climate-invariant models learn mappings that are valid across climates and geographies \textit{during} training. For this purpose, we track the models' generalizability throughout the training process as explained below.

Fig~S7 shows learning curves; the color of each line indicates the dataset the model was \textit{trained} in, while the color of the row indicates the dataset the model was \textit{tested} in. To gain intuition, we can start by looking at lines that have the same color as their axes: These are the ``standard'' learning curve showing that each model’s validation loss in the same climate/geography monotonically decreases as the model is trained, confirming that we are not overfitting the training set. 

We are now ready to zoom in on a key result of this manuscript: The learning curve of the ``climate-invariant'' NN trained in the cold aquaplanet but tested in the warm aquaplanet (starred blue line in the red box (a)). Impressively, this learning curve is mostly decreasing, confirming that \textit{``climate-invariant'' NNs are able to continuously learn about subgrid thermodynamics in the warm aquaplanet as they are trained in the cold aquaplanet}. In contrast, the ``raw-data'' NN trained in the cold aquaplanet but tested in the warm aquaplanet (circled blue line in the red box (a)) makes extremely large generalization errors, which worsen as the model is trained in the cold aquaplanet. 

\textit{``Climate-invariant'' NNs also facilitate learning across geographies}, i.e., from the aquaplanet to the Earth-like simulations (starred blue line in green box (b) is consistently below circled blue line) and vice-versa (starred green line in blue box (c) is consistently below circled green line). ``Climate-invariant'' transformations additionally improve the MLR baseline’s generalization ability (see right column, e.g., starred blue line in red box (a) and starred green line in blue box (c)), albeit less dramatically. This smaller improvement in MLR’s generalization abilities is linked to its relatively small number of free parameters, resulting in (1) ``raw-data'' MLRs generalizing better than ``raw-data'' NNs; and (2) MLRs having lower representation power and fitting their training sets less well, limiting the maximal accuracy of ``climate-invariant'' MLRs on the test set.

There are a few cases in which transforming inputs does not fully solve the generalization problem, e.g., when trying to generalize from the aquaplanet to the Earth-like simulation (starred blue line in green box (b)). NNs with DP fit their training set less well (squared lines that have the same color as their boxes are above corresponding circled/starred lines). However, they improve generalization in difficult cases (e.g., squared blue line in green box (b)) and do not overly deteriorate generalization in cases where the input transformations work particularly well (e.g., squared green line in blue box (a)). This confirms that \textit{combining physics-guided generalization methods (e.g., physical transformation of the inputs/outputs) with standard ML generalization methods (e.g., DP)} is advantageous.

\subsection*{D3. Geographic Skill\label{ap:R2-NS}}

This section complements Sec~4.3 by presenting different cross-sections of NN skill \textit{after} training. Our results confirm that while raw-data NN trained in the cold climate struggle to generalize to the warm climate's Tropics, the climate-invariant mapping alleviates this limitation.

Fig~S8a, which shows cross-section of the coefficient of determination $R^2$ (1 or yellow for perfect predictions, and -1 or blue for errors larger or equal to two standard deviations) exposes the raw-data NN's poor generalization skill in the warm (+4K) Tropics. In contrast, Fig~S8b underlines how climate-invariant NNs improve generalization throughout the atmosphere in the warm Tropics without deteriorating skill in the mid-latitudes and poles of the warm simulation. This consideration helped us choose our final input transformation, as the $\boldsymbol{\tilde{T}_\mathrm{from\ NS}}$ temperature transformation significantly deteriorated generalization in the mid-latitudes, while the $\boldsymbol{\tilde{T}_\mathrm{buoyancy}}$ transformation helps generalization in the Tropics without overly compromising skills at other latitudes. There is a slight skill compromise at high latitudes, as can be seen by comparing the second rows of Fig~S8a and Fig~S8b, which is especially apparent in the SAM case and can be partially traced back to challenges in generalizing subgrid ice sedimentation (not shown here, see \cite{yuval2020stable} for details).

To show that the improved generalization skill of climate-invariant NNs for subgrid heating is not unique to the mid-troposphere (see Fig~5), in Fig~S9 we also show the generalization skill of climate-invariant NNs near the surface. Consistent with \cite{mooers2021assessing,han2020moist}, the highest skill for the training climate is over land for all NNs as most of the variability comes from the diurnal cycle, which is easy to predict for NNs. Similarly to Fig~5, the generalization error is apparent for the raw-data NN (a) and mostly solved by making the NN climate-invariant (b). 

\subsection*{D4. Visualizing Climate-Invariant Mappings\label{ap:Cl-inv-MLR}}

Before using SHAP in Section~4.4 to visualize the difference between raw-data and climate-invariant mappings, we test simple linear methods to analyze ML models. First, we directly plot the weights $\boldsymbol{A} $ (see Eq~\ref{eq:MLR}) of our multi-linear regressions in Fig~S10/S11. Second, we plot the mean Jacobian of our NN calculated via automatic differentiation in Fig~S12/S13. Unlike SHAP, the MLR weights and the Jacobian matrices both suggest that the climate-invariant mapping is non-local in the vertical. Fig~S10/S11 is consistent with the climate-invariant MLR generalizing only slightly better than the raw-data MLR (see top-right panel of Fig~S7). Meanwhile, comparing Fig~S12/S13 to the full SHAP feature importance matrix (Fig~S14/S15) suggests that while the linear sensitivity of subgrid heating/moistening with respect to lower-tropospheric plume buoyancy is high (top panels of Fig~S12b), which is expected, subgrid heating/moistening can be well-predicted using mostly local plume buoyancy information (top panels of Fig~S14b).

\newpage

\begin{table*}
\begin{centering}
\begin{tabular}{c|c|c|c|c}
Row & Input & SPCAM3 & SPCESM2 & SAM\tabularnewline
\hline
\hline
1 & $q_{600\mathrm{hPa}}$ & \textcolor{gray}{20.3}, \textcolor{red}{35.1} & \textcolor{gray}{17.1}, \textcolor{red}{29.5} & \textcolor{gray}{22.1} \tabularnewline2 & $q_{\mathrm{deficit,}600\mathrm{hPa}}$ & \textcolor{gray}{24.9}, \textcolor{red}{36.5} & \textcolor{gray}{18.1}, \textcolor{red}{31.0} & \textcolor{gray}{30.0} \tabularnewline3 & $\mathrm{RH}_{600\mathrm{hPa}}$ & \textcolor{gray}{3.6}, \textcolor{red}{8.2} & \textcolor{gray}{3.2}, \textcolor{red}{5.3} & \textcolor{gray}{4.3} \tabularnewline\hline
4 & $T_{850\mathrm{hPa}}$ & \textcolor{gray}{53.2}, \textcolor{red}{64.3} & \textcolor{gray}{25.7}, \textcolor{red}{37.3} & \textcolor{gray}{51.9} \tabularnewline5 & $T_{\mathrm{from\ NS},850\mathrm{hPa}}$ & \textcolor{gray}{5.1}, \textcolor{red}{6.2} & \textcolor{gray}{3.3}, \textcolor{red}{6.4} & \textcolor{gray}{10.5} \tabularnewline6 & $B_{\mathrm{plume},850\mathrm{hPa}}$ & \textcolor{gray}{9.4}, \textcolor{red}{14.7} & \textcolor{gray}{3.6}, \textcolor{red}{7.6} & \textcolor{gray}{5.8} \tabularnewline\hline
7 & $T_{150\mathrm{hPa}}$ & \textcolor{gray}{30.2}, \textcolor{red}{33.5} & \textcolor{gray}{31.6}, \textcolor{red}{34.4} & \textcolor{gray}{65.6} \tabularnewline8 & $T_{\mathrm{from\ NS},150\mathrm{hPa}}$ & \textcolor{gray}{38.0}, \textcolor{red}{53.7} & \textcolor{gray}{14.5}, \textcolor{red}{28.7} & \textcolor{gray}{51.0} \tabularnewline9 & $B_{\mathrm{plume},150\mathrm{hPa}}$ & \textcolor{gray}{35.1}, \textcolor{red}{42.2} & \textcolor{gray}{10.4}, \textcolor{red}{20.9} & \textcolor{gray}{21.1} \tabularnewline\hline
10 & $\mathrm{LHF}$ & \textcolor{gray}{8.6}, \textcolor{red}{14.5} & \textcolor{gray}{9.7}, \textcolor{red}{12.5} & \textcolor{gray}{} \tabularnewline11 & $\mathrm{LHF}_{q}$ & \textcolor{gray}{4.7}, \textcolor{red}{9.5} & \textcolor{gray}{10.0}, \textcolor{red}{10.7} & \textcolor{gray}{} \tabularnewline12 & $\mathrm{LHF}_{\Delta q}$ & \textcolor{gray}{6.3}, \textcolor{red}{9.9} & \textcolor{gray}{9.9}, \textcolor{red}{14.0} & \textcolor{gray}{} \tabularnewline\end{tabular}
\par\end{centering}{\small \par}

\caption{\textbf{Hellinger distance (in $\%$) away from the (-4K) simulation for the PDFs of key inputs $\left(q_{600\mathrm{hPa}},T_{850\mathrm{hPa}},T_{150\mathrm{hPa}},\mathrm{LHF}\right)$ and their transformations.} (+0K) distance in gray and (+4K) distance in red.}
\end{table*}

\begin{table*}
\begin{center}
    
\resizebox{\textwidth}{!}{%
\begin{tabular}{c|c|c|c|c}
Model & Spatiotemporal Resolution & Training Set & Validation Set & Test Set\tabularnewline
\hline 
SPCAM3 & $(2.8$°$\times2.8$°$)_{\mathrm{T42}}\times$30lev$\times30\mathrm{min}$ & Yr2, Mo1-4$\rightarrow$47M  & Yr2, Mo5-8$\rightarrow$48M & Yr1, Mo6-9$\rightarrow$48M\tabularnewline
\hline 
SPCESM2  & 2.5°$\times$1.9°$\times$30lev$\times15\mathrm{min}$ & Yr1, Day0-9/Mo$\rightarrow$143M & Yr2, Day0-9/Mo$\rightarrow$143M & Yr2, Day20-28/Mo $\rightarrow$118M\tabularnewline
\hline 
SAM \textcolor{blue}{(-4K)} & 96km$\times$96kmx48lev$\times$180min & \textcolor{blue}{day 225-545 $\rightarrow$ 13.8M} & \textcolor{blue}{day 545-562 $\rightarrow$ 0.7M} & \textcolor{blue}{day 562-587 $\rightarrow$ 2.6M}\tabularnewline
\cline{3-5} \cline{4-5} \cline{5-5} 
\textcolor{gray}{(+0K)} & 96kmx96km$\times$48lev$\times$180min & \textcolor{gray}{-} & \textcolor{gray}{-} & \textcolor{gray}{day 380-405 $\rightarrow$ 2.6M}\tabularnewline
\hline 
\end{tabular}
}
\caption{\textbf{Characteristics of the training/validation/test sets used in this manuscript.} The spatiotemporal resolution uses the format longitude $\times$ latitude $\times$ vertical levels $\times$ time. For SPCAM3, which uses a T42 spectral truncation, we use months 1 to 4 of the second simulation year to build the training set, resulting in $\approx47M$ samples. For SPCESM2, we use the first 9 days of every month of the first simulation year to build the training set, resulting in $\approx143M$ samples.}

\end{center}
\end{table*}

\begin{table}
\begin{centering}
\begin{tabular}{c|c|c|c|c}
Row & Input & SPCAM3 & SPCESM2 & SAM\tabularnewline
\hline
\hline
1 & $q_{600\mathrm{hPa}}$ & \textcolor{gray}{0.5}, \textcolor{red}{0.8} & \textcolor{gray}{0.4}, \textcolor{red}{0.7} & \textcolor{gray}{0.5} \tabularnewline2 & $q_{\mathrm{deficit,}600\mathrm{hPa}}$ & \textcolor{gray}{0.7}, \textcolor{red}{1.0} & \textcolor{gray}{0.4}, \textcolor{red}{0.8} & \textcolor{gray}{0.8} \tabularnewline3 & $\mathrm{RH}_{600\mathrm{hPa}}$ & \textcolor{gray}{0.1}, \textcolor{red}{0.2} & \textcolor{gray}{0.1}, \textcolor{red}{0.1} & \textcolor{gray}{0.1} \tabularnewline\hline
4 & $T_{850\mathrm{hPa}}$ & \textcolor{gray}{1.4}, \textcolor{red}{1.9} & \textcolor{gray}{0.5}, \textcolor{red}{0.8} & \textcolor{gray}{1.3} \tabularnewline5 & $T_{\mathrm{from\ NS},850\mathrm{hPa}}$ & \textcolor{gray}{0.1}, \textcolor{red}{0.1} & \textcolor{gray}{0.1}, \textcolor{red}{0.1} & \textcolor{gray}{0.2} \tabularnewline6 & $B_{\mathrm{plume},850\mathrm{hPa}}$ & \textcolor{gray}{0.2}, \textcolor{red}{0.3} & \textcolor{gray}{0.1}, \textcolor{red}{0.2} & \textcolor{gray}{0.1} \tabularnewline\hline
7 & $T_{150\mathrm{hPa}}$ & \textcolor{gray}{0.6}, \textcolor{red}{0.7} & \textcolor{gray}{0.7}, \textcolor{red}{0.7} & \textcolor{gray}{1.5} \tabularnewline8 & $T_{\mathrm{from\ NS},150\mathrm{hPa}}$ & \textcolor{gray}{0.9}, \textcolor{red}{1.4} & \textcolor{gray}{0.3}, \textcolor{red}{0.6} & \textcolor{gray}{1.4} \tabularnewline9 & $B_{\mathrm{plume},150\mathrm{hPa}}$ & \textcolor{gray}{1.0}, \textcolor{red}{1.2} & \textcolor{gray}{0.2}, \textcolor{red}{0.4} & \textcolor{gray}{0.5} \tabularnewline\hline
10 & $\mathrm{LHF}$ & \textcolor{gray}{0.2}, \textcolor{red}{0.3} & \textcolor{gray}{0.2}, \textcolor{red}{0.3} & \textcolor{gray}{} \tabularnewline11 & $\mathrm{LHF}_{q}$ & \textcolor{gray}{0.1}, \textcolor{red}{0.2} & \textcolor{gray}{0.2}, \textcolor{red}{0.2} & \textcolor{gray}{} \tabularnewline12 & $\mathrm{LHF}_{\Delta q}$ & \textcolor{gray}{0.1}, \textcolor{red}{0.2} & \textcolor{gray}{0.2}, \textcolor{red}{0.3} & \textcolor{gray}{} \tabularnewline\end{tabular}
\par\end{centering}{\small \par}

\caption{\textbf{Jensen-Shannon distance away from the (-4K) simulation for the PDFs of key inputs $\left(q_{600\mathrm{hPa}},T_{850\mathrm{hPa}},T_{150\mathrm{hPa}},\mathrm{LHF}\right)$ and their transformations.} (+0K) distance in gray and (+4K) distance in red.}

\end{table}

\begin{figure}
    \centerline{\includegraphics[width=\linewidth]{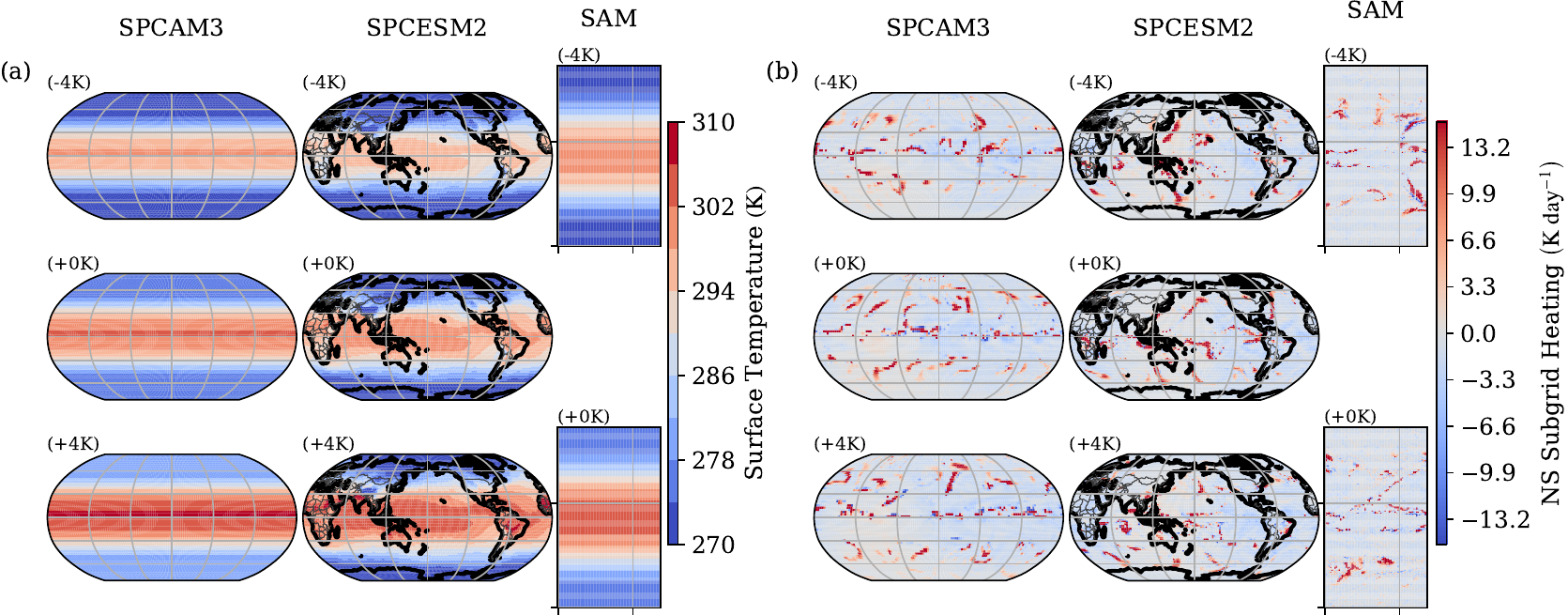}}
    \caption{\textbf{Surface temperatures and subgrid heating rate in the three utilized atmospheric models.} (a) Prescribed surface temperature (in K) for (left) the aquaplanet SPCAM3 model and (right) the hypohydrostatic SAM model. (center) Annual-mean, near-surface air temperatures in the Earth-like SPCESM2 model. (b) Snapshots of near-surface subgrid heating rate (in K/day). For each model, we show the cold (-4K), reference (+0K), and warm (+4K) simulations.}
    \label{fig:Data_Figure}
\end{figure}

\begin{figure}
\centerline{\includegraphics[width=\linewidth]{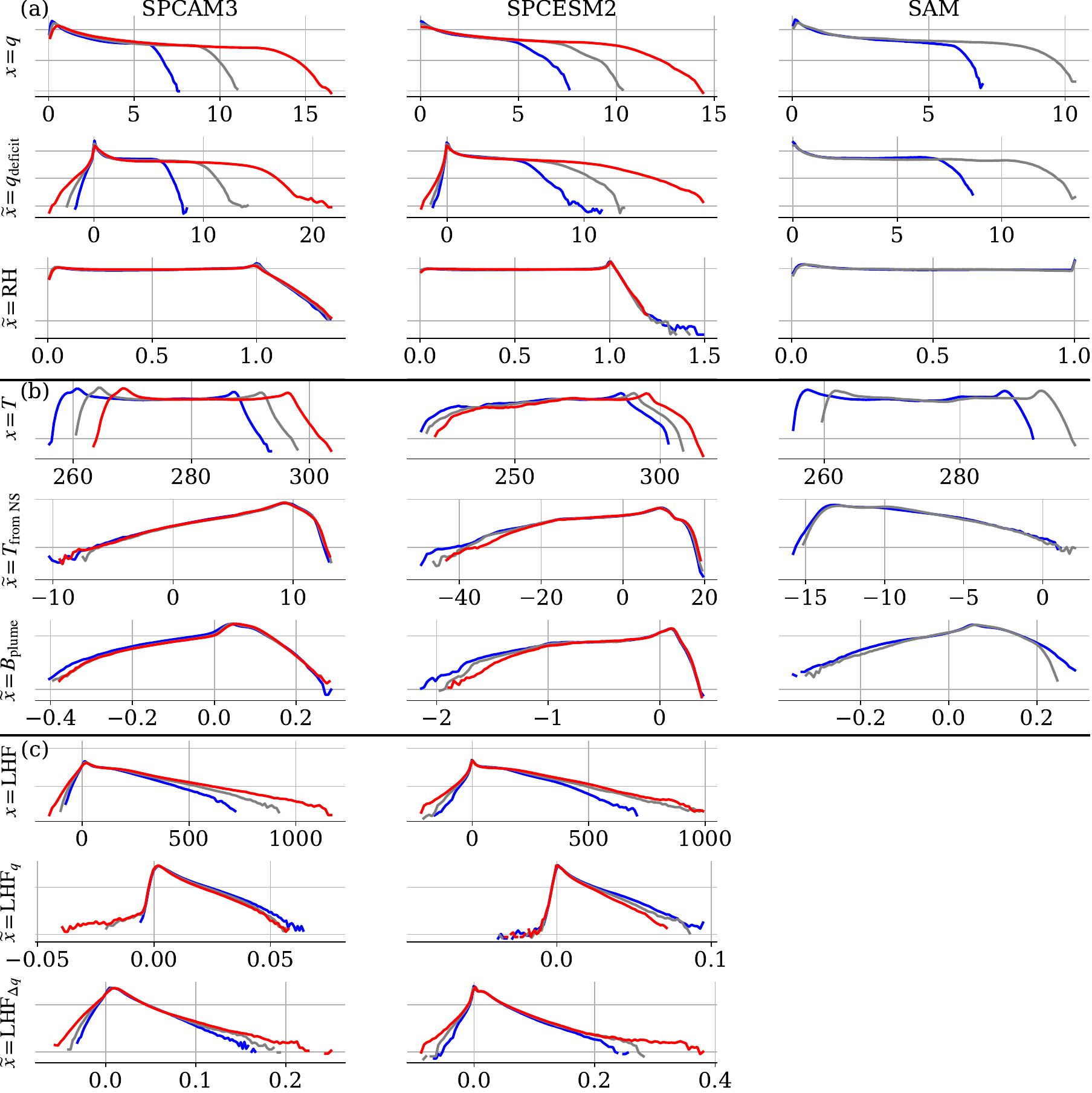}}
 \caption{\textbf{Univariate PDFs of the (a) 600hPa specific humidity, (b) 850hPa temperature, and (c) latent heat flux in the cold (blue), reference (gray), and warm (red) simulations of each model (SPCAM3, SPCESM2, and SAM).} For each variable, we also show the PDFs of the two transformations discussed in SM SB.2. From top to bottom, the variables are $q$ (g/kg), $q_{\mathrm{deficit}}$ (g/kg), RH, T (K), $T_{\mathrm{from\ NS}}$ (K), $B_{\mathrm{plume}}$ (m/s$^2$), LHF (W/m$^2$), $\mathrm{LHF}_{q} $ ($\mathrm{kg\ m^{-2}s^{-1}} $), and $\mathrm{LHF}_{\Delta q} $ ($\mathrm{kg\ m^{-2}s^{-1}} $). For a given variable and transformation, we use the same vertical logarithmic scale across models. Note that unlike for $q $, the best options for $T $ and $\mathrm{LHF} $ do not decrease distribution distance more than the second best options, which is discussed in text. }
\end{figure}

\begin{figure}
\centerline{\includegraphics[width=\columnwidth]{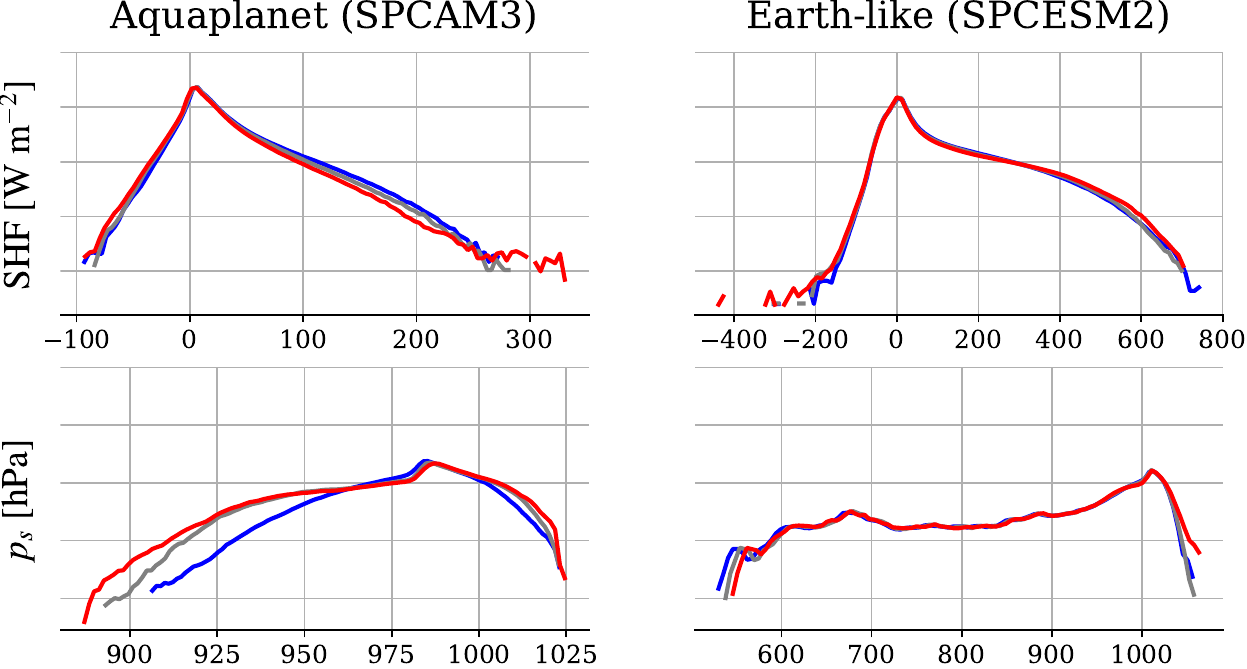}}
 \caption{\textbf{Univariate PDFs of the sensible heat flux and surface pressure in the cold (blue), reference (gray), and warm (red) simulations of SPCAM3 and SPCESM2.} For a given variable, we use the same vertical logarithmic scale across models.\label{f_SHF}}
\end{figure}

\begin{figure}
\centerline{\includegraphics[width=0.5\columnwidth]{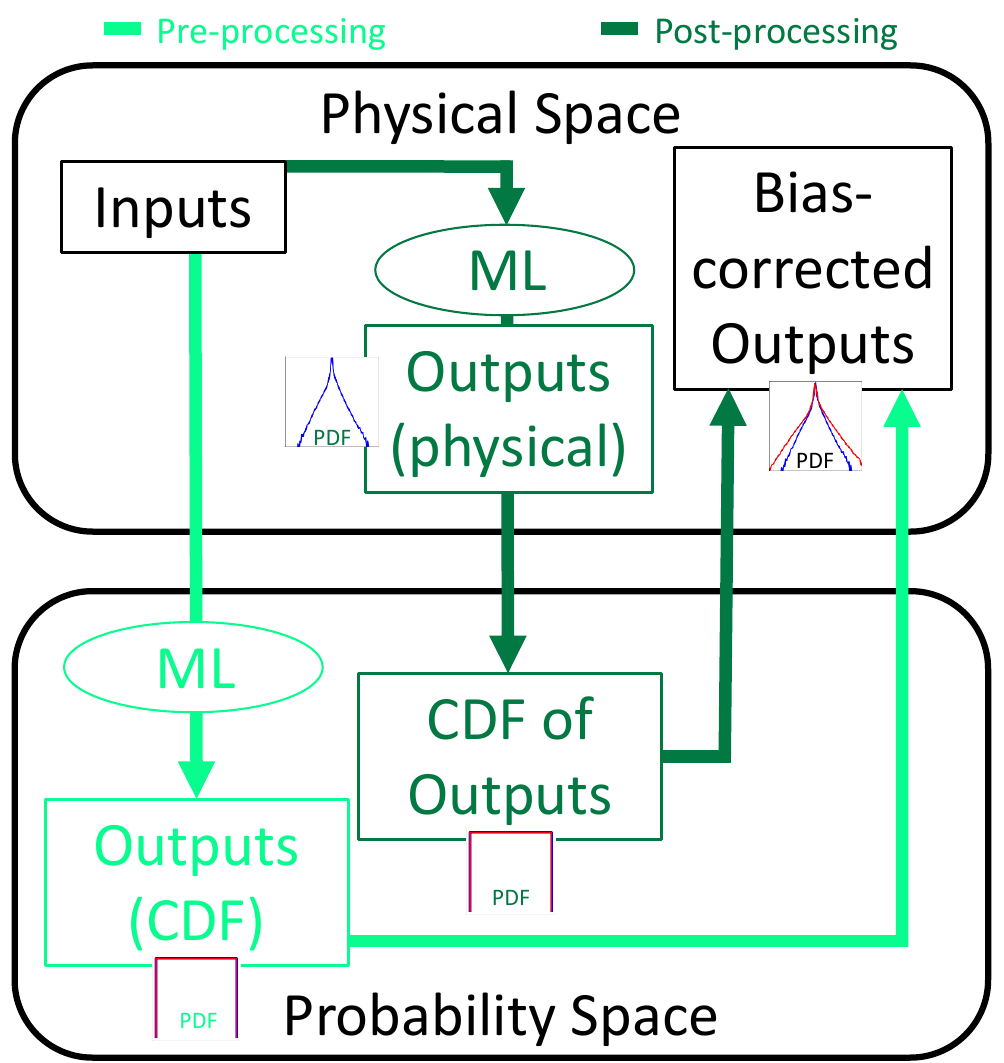}}
 \caption{\textbf{Two types of bias-correction methods used to transform outputs in SM B2.} (1, dark green, ``post-processing'' method) Quantile mapping is typically done \textit{after} training the model to bias-correct the outputs, and (2, light green, ``pre-processing'' method) we additionally test directly making predictions in probability space by converting the outputs to their CDF values \textit{before} training. Note that this usually changes the loss function.}\label{f_post_preprocessing}
\end{figure} 

\begin{figure}
\centerline{\includegraphics[width=\columnwidth]{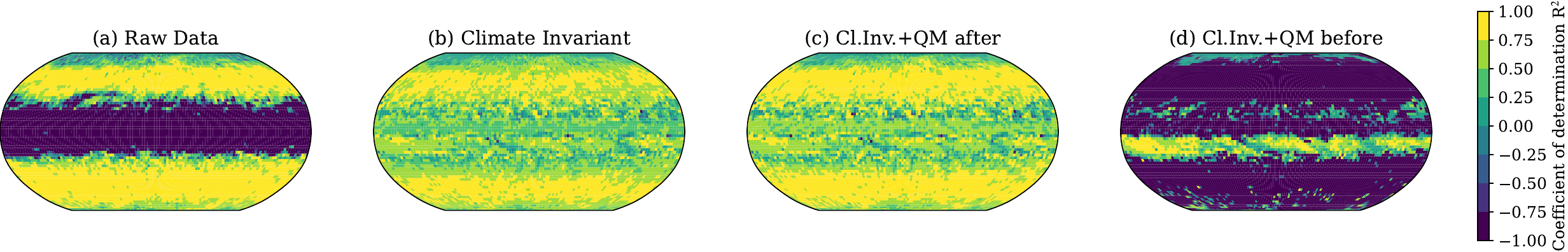}}
 \caption{\textbf{Transforming outputs via quantile mapping \textit{after} training slightly improves the climate invariant model's ability to generalize from a cold to a warm climate.} Coefficient of determination $R^{2}$ for 500-hPa subgrid heating of raw-data (a), climate-invariant (b), climate-invariant with outputs transformed \textit{after} training (c), climate-invariant with outputs transformed \textit{before} training (d) NNs trained using the cold (-4K) training set of SPCAM3 and calculated over the warm (+4K) training set of SPCAM3.}\label{f_R2_post_preprocessing}
\end{figure}

\begin{figure}
\centerline{\includegraphics[width=0.5\columnwidth]{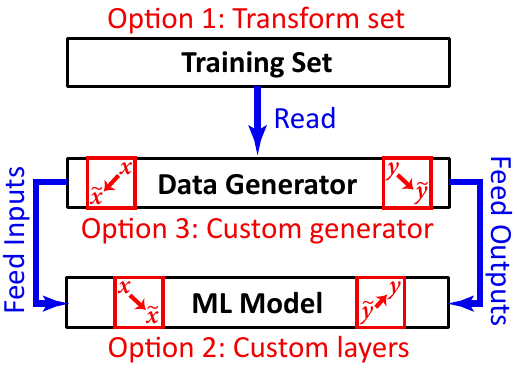}}
 \caption{\textbf{Implementation of the climate-invariant ML framework.} The physical transformations can be implemented by (Option~1) transforming the training set, (Option~2) adding custom layers to the ML model, or (Option~3) customizing the data generator so that it automatically transforms the model inputs/outputs.}
\end{figure} 

\begin{figure*}
\centerline{\includegraphics[width=\textwidth]{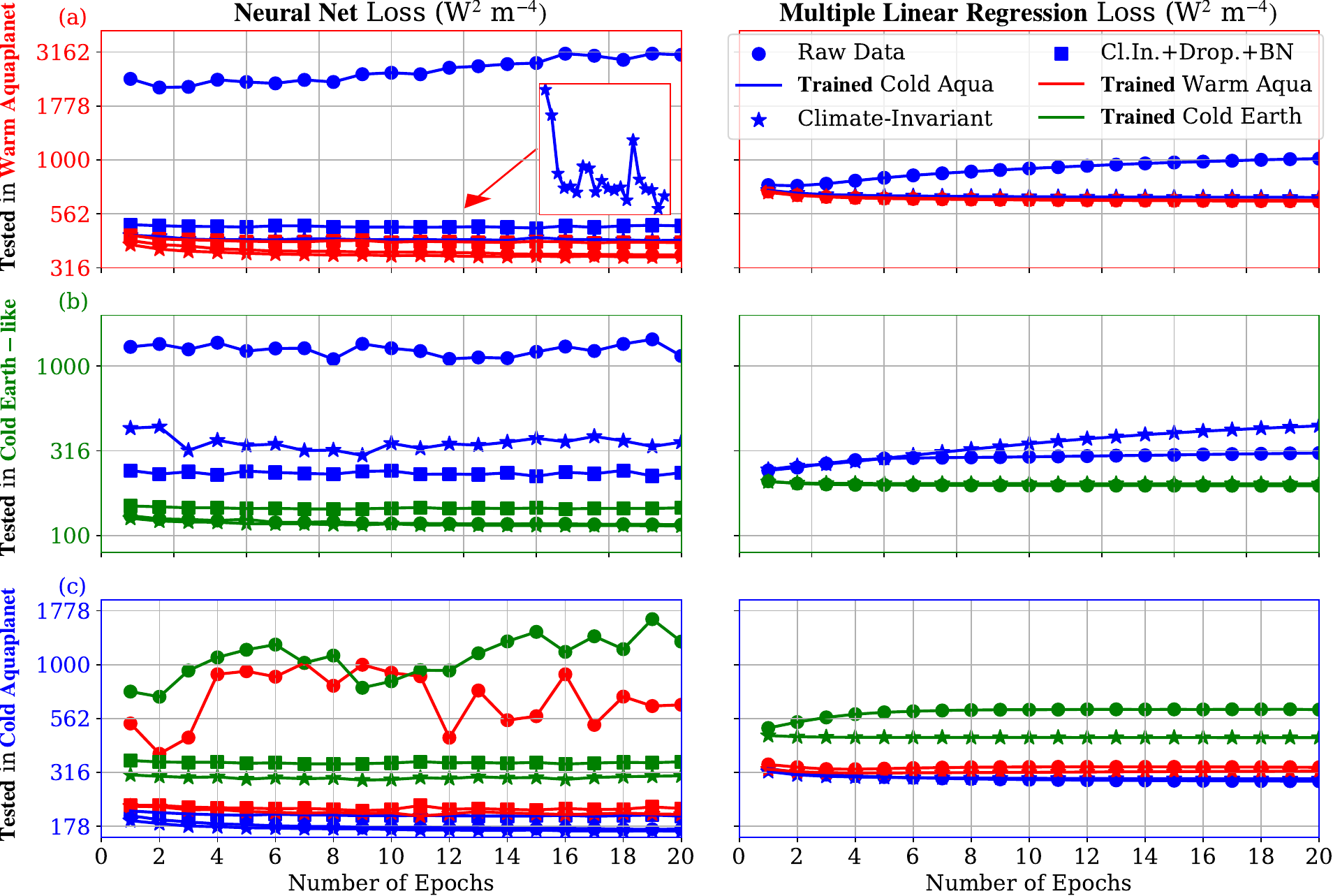}}
 \caption{\textbf{Unlike raw-data models, climate-invariant models continuously learn about subgrid thermodynamics in the warm aquaplanet as they are trained in the cold aquaplanet.} More generally, they can learn information about configurations that differ from the one they were trained in. Learning curves of neural nets (left) and multiple linear regressions (right) \textbf{tested} in the (-4K) cold aquaplanet simulation (a, top row), the (+4K) warm aquaplanet simulation (b, middle row), and the (-4K) cold Earth-like simulation (c, bottom row). The lines' colors indicate the \textbf{training} dataset, while their symbols refer to whether the ML model is raw-data (circle), climate-invariant (star), or  climate-invariant with dropout layers before each activation function and batch normalization (square). (b) We additionally zoom in on the climate-invariant neural network's learning curve in the (+4K) simulation.}
\end{figure*}

\begin{figure}
    \centering
    \includegraphics{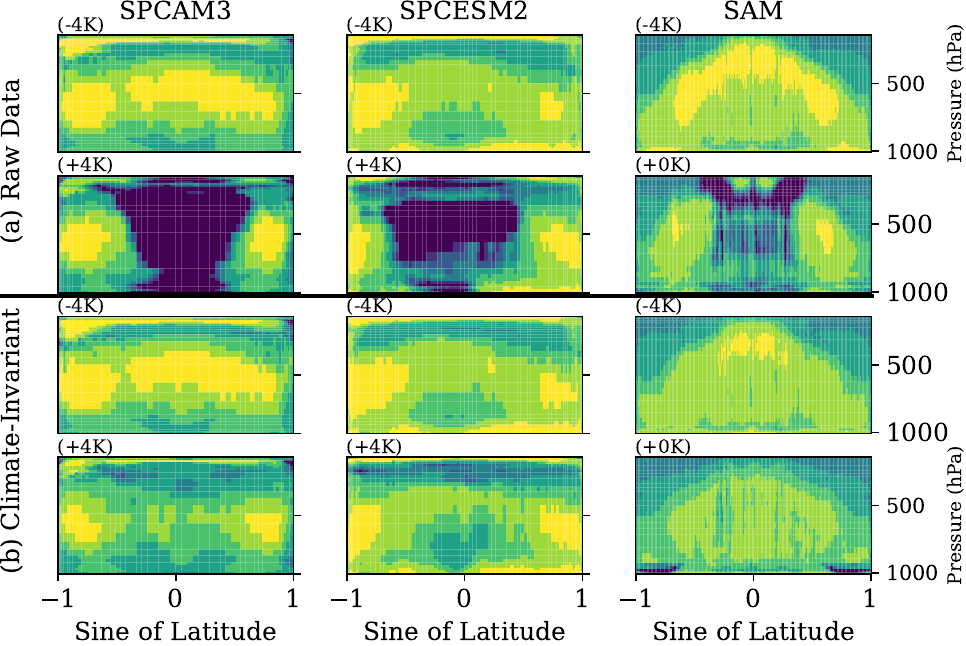}
    \caption{\textbf{Latitude-Pressure cross-section of subgrid heating's coefficient of determination $R^2$.} We train (a) raw-data and (b) climate-invariant NNs using the cold (-4K) training set of each model (SPCAM3, SPCESM2, and SAM) and test them in the cold (-4K) and warm (+4K) climates. See Fig~5 or Fig~S9 for the colorbar.}
    \label{fig:Lat-P_Cross_Section}
\end{figure}

\begin{figure*}
\centerline{\includegraphics[width=\textwidth]{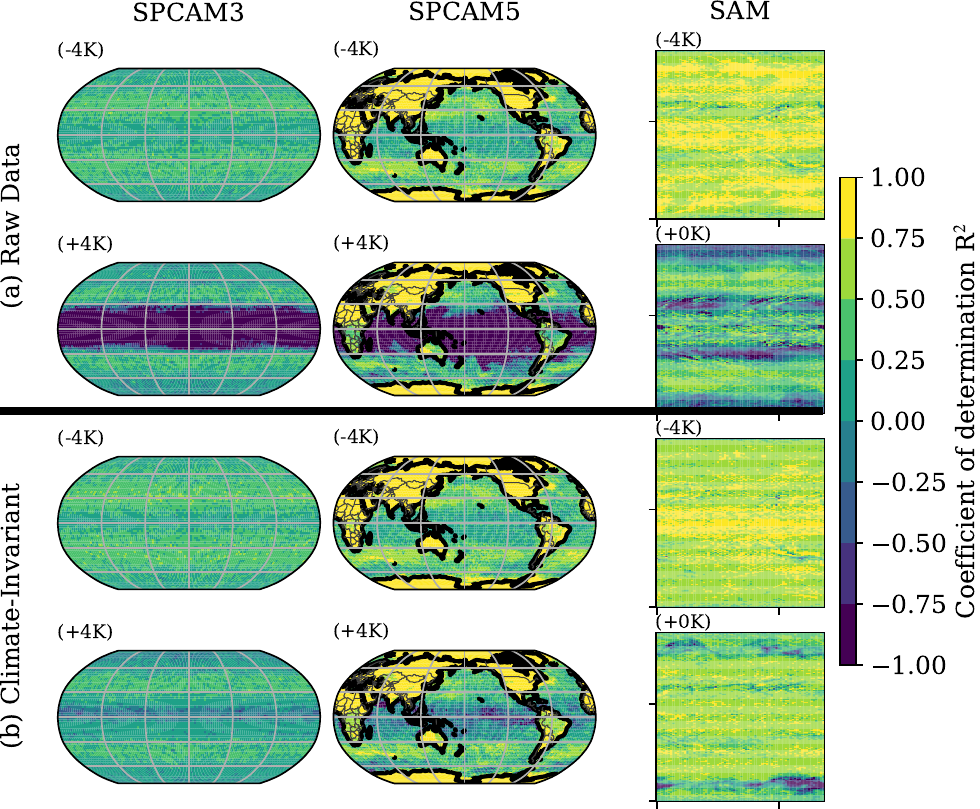}}
 \caption{\textbf{Climate-invariant NNs mitigate generalization issues in the ``Warm Tropics'' for near-surface subgrid heating.} Same as Fig~6 for near-surface subgrid heating.}
\end{figure*}

\begin{figure*}
\centerline{\includegraphics[width=\textwidth]{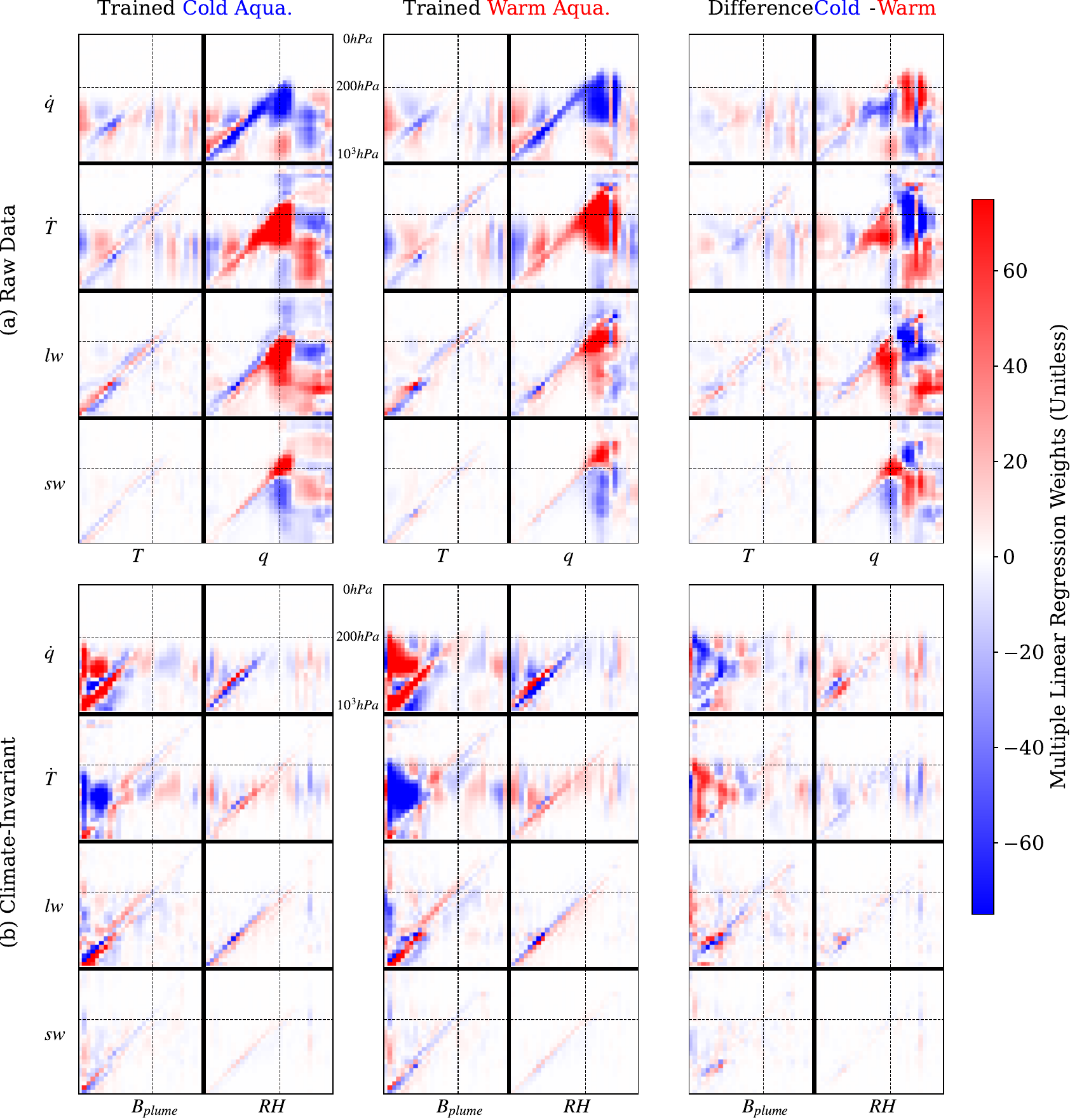}}
 \caption{\textbf{Weights of the (a) raw-data and (b) climate-invariant multi-linear regressions trained in the cold (-4K) aquaplanet simulation (left), the warm (+4) warm aquaplanet simulation (middle), and their difference (right).} The x-axes indicate the vertical levels of the inputs, from the surface (left, 10$^{3}$hPa) to the top of the atmosphere (right, 0hPa), while the y-axes indicate the vertical levels of the outputs, from the surface (bottom, 10$^{3}$hPa) to the top of the atmosphere (top, 0hPa). We additionally indicate the 200hPa vertical level with dotted black lines.}
\end{figure*}

\begin{figure*}
\centerline{\includegraphics[width=\textwidth]{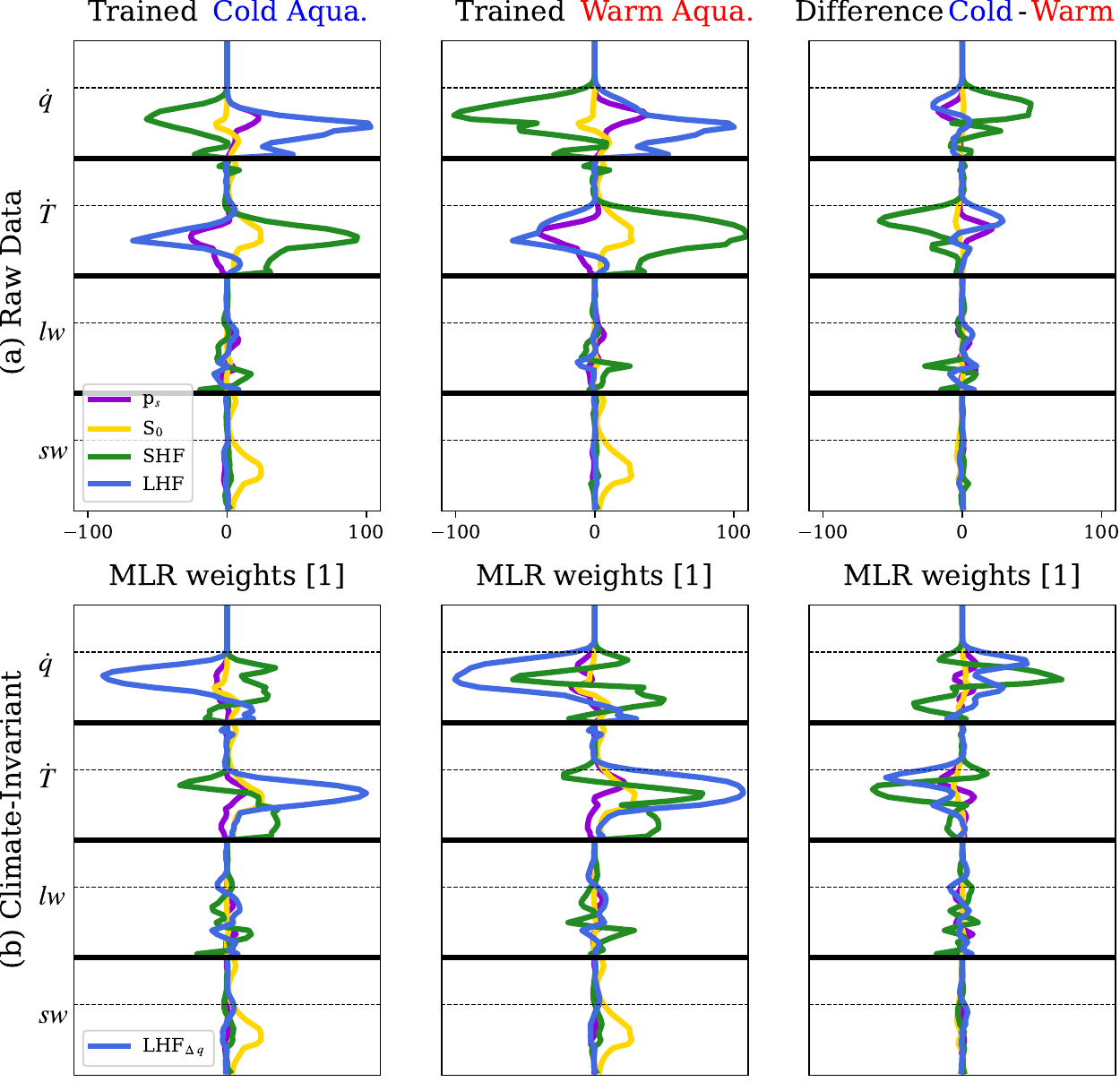}}
 \caption{\textbf{Same as Fig~S10, but for the four scalar inputs used in addition to the temperature and specific humidity inputs.} The scalar inputs are surface pressure ($p_{s} $, purple line), solar insolation ($S_{0} $, yellow line), surface sensible heat flux (SHF, green line), and surface latent heat flux (LHF, blue line). For the climate-invariant mapping (b), LHF is transformed to $\mathrm{LHF}_{\Delta q} $ as described in Sec~3, which in conjunction with the temperature and humidity transformations, changes the multi-linear regression weights for all input variables.}
\end{figure*}

\begin{figure*}
\centerline{\includegraphics[width=\textwidth]{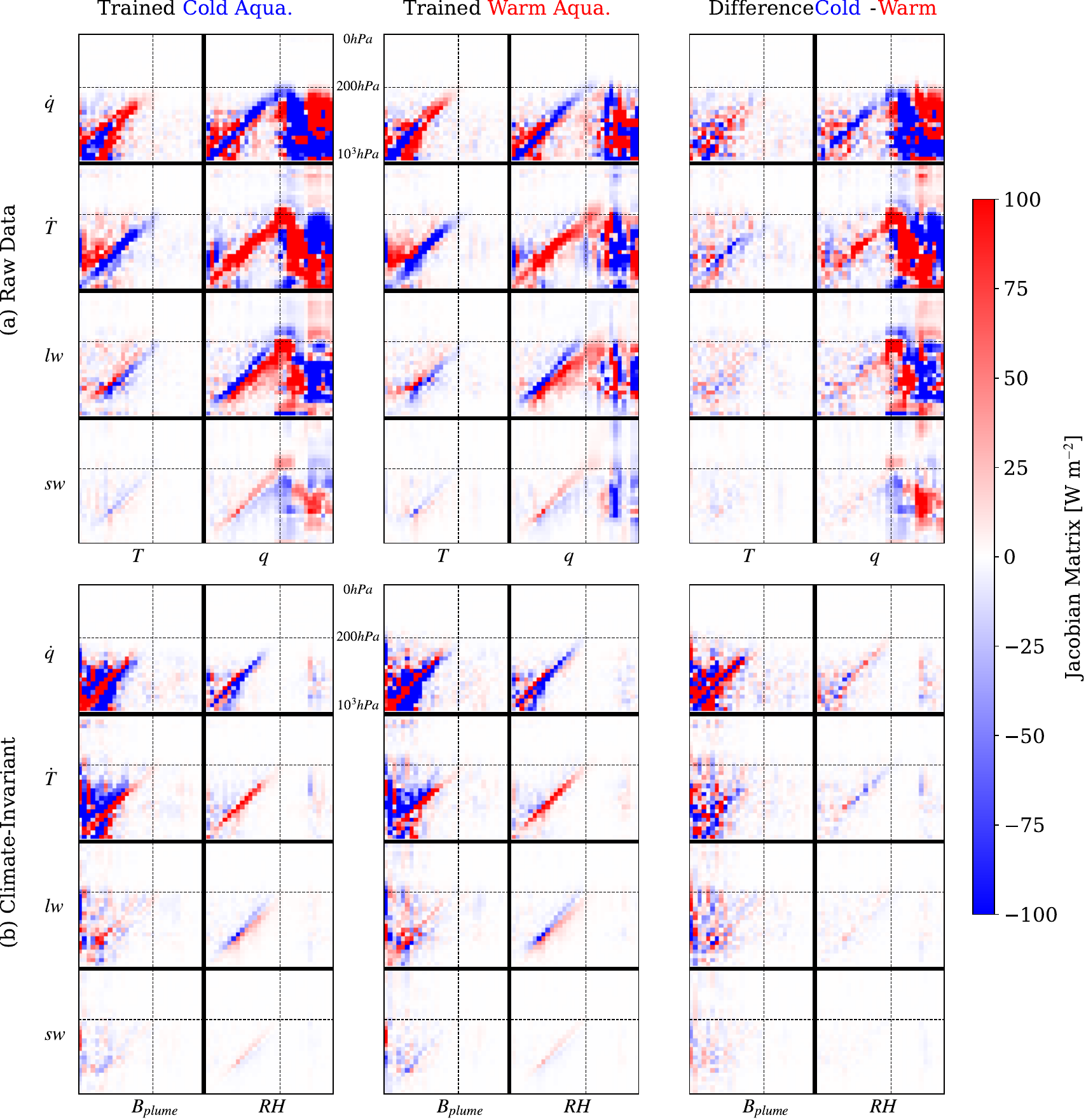}}
 \caption{\textbf{Jacobian matrices of the (a) raw-data and (b) climate-invariant neural networks trained in the cold (-4K) aquaplanet simulation (left), the warm (+4) warm aquaplanet simulation (middle), and their difference (right).} The x-axes indicate the vertical levels of the inputs, from the surface (left, 10$^{3}$hPa) to the top of the atmosphere (right, 0hPa), while the y-axes indicate the vertical levels of the outputs, from the surface (bottom, 10$^{3}$hPa) to the top of the atmosphere (top, 0hPa). We additionally indicate the 200hPa vertical level with dotted black lines.}
\end{figure*}

\begin{figure*}
\centerline{\includegraphics[width=\textwidth]{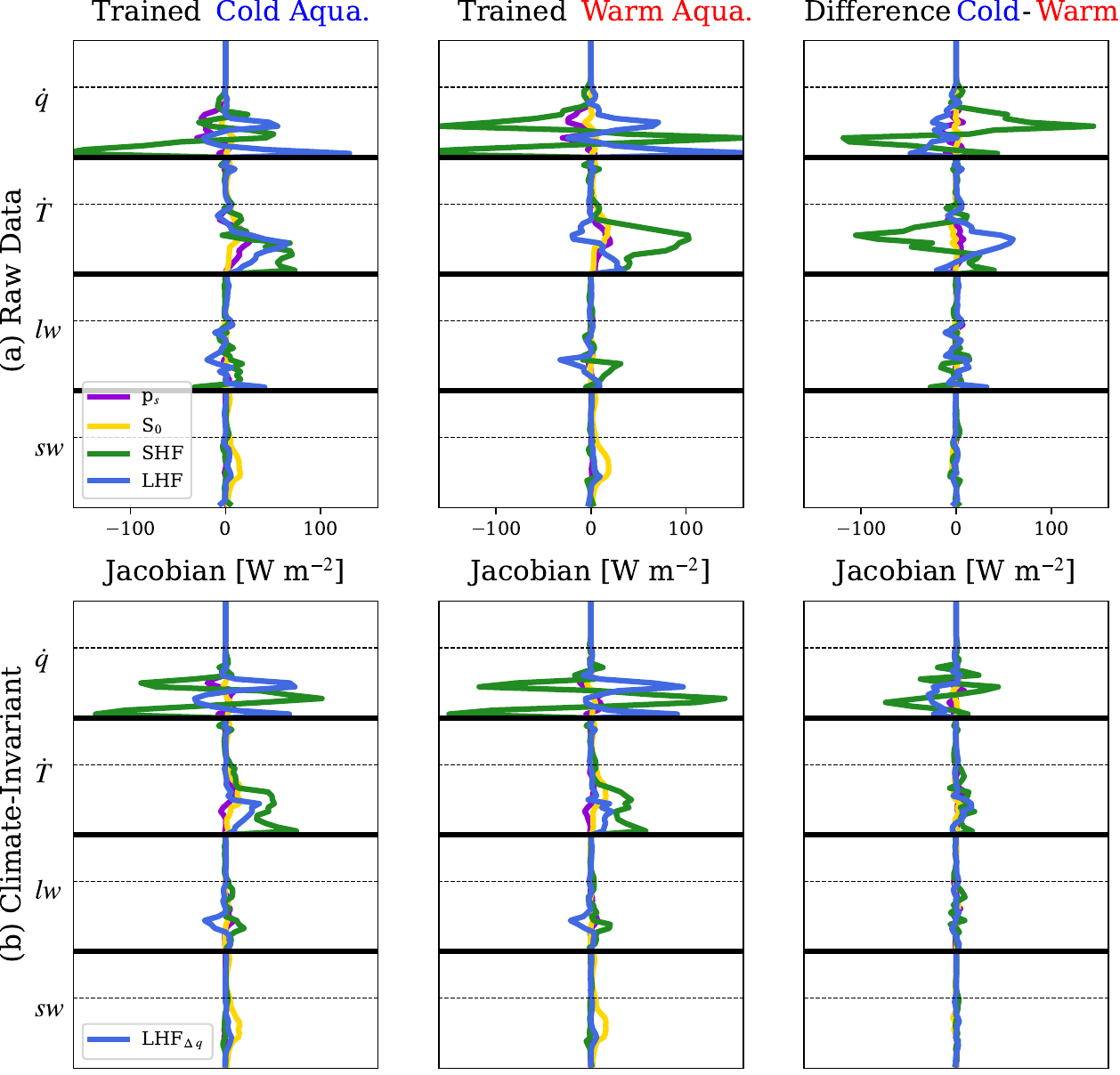}}
 \caption{\textbf{Same as Fig~S12, but for the four scalar inputs used in addition to the temperature and specific humidity inputs.} The scalar inputs are surface pressure ($p_{s} $, purple line), solar insolation ($S_{0} $, yellow line), surface sensible heat flux (SHF, green line), and surface latent heat flux (LHF, blue line). For the climate-invariant mapping (b), LHF is transformed to $\mathrm{LHF}_{\Delta q} $ as described in Sec~3, which in conjunction with the temperature and humidity transformations, changes the Jacobian matrices for all input variables.}
\end{figure*}

\begin{figure*}
\centerline{\includegraphics[width=\textwidth]{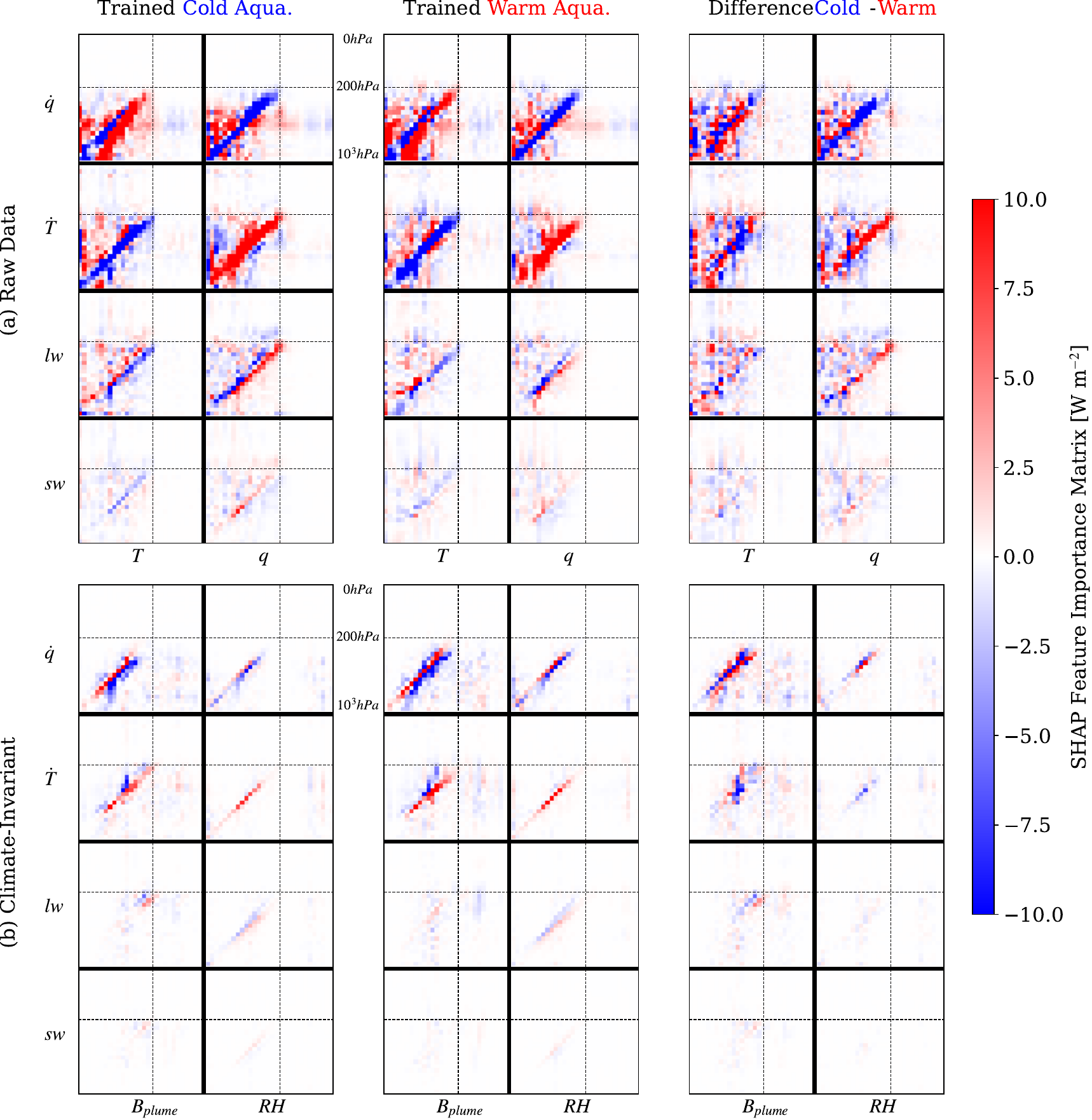}}
\caption{\textbf{SHAP feature importance matrix for the (a) raw-data and (b) climate-invariant neural nets trained in the cold (-4K) aquaplanet simulation (left), the warm (+4) warm aquaplanet simulation (middle), and their difference (right).} To calculate these matrices, we sample inputs from the (+4K) warm aquaplanet simulation for all ML models to facilitate inter-model comparison. The x-axes indicate the vertical levels of the inputs, from the surface (left, 10$^{3}$hPa) to the top of the atmosphere (right, 0hPa), while the y-axes indicate the vertical levels of the outputs, from the surface (bottom, 10$^{3}$hPa) to the top of the atmosphere (top, 0hPa). We additionally indicate the 200hPa vertical level with dotted black lines.}
\end{figure*}

\begin{figure*}
\centerline{\includegraphics[width=\textwidth]{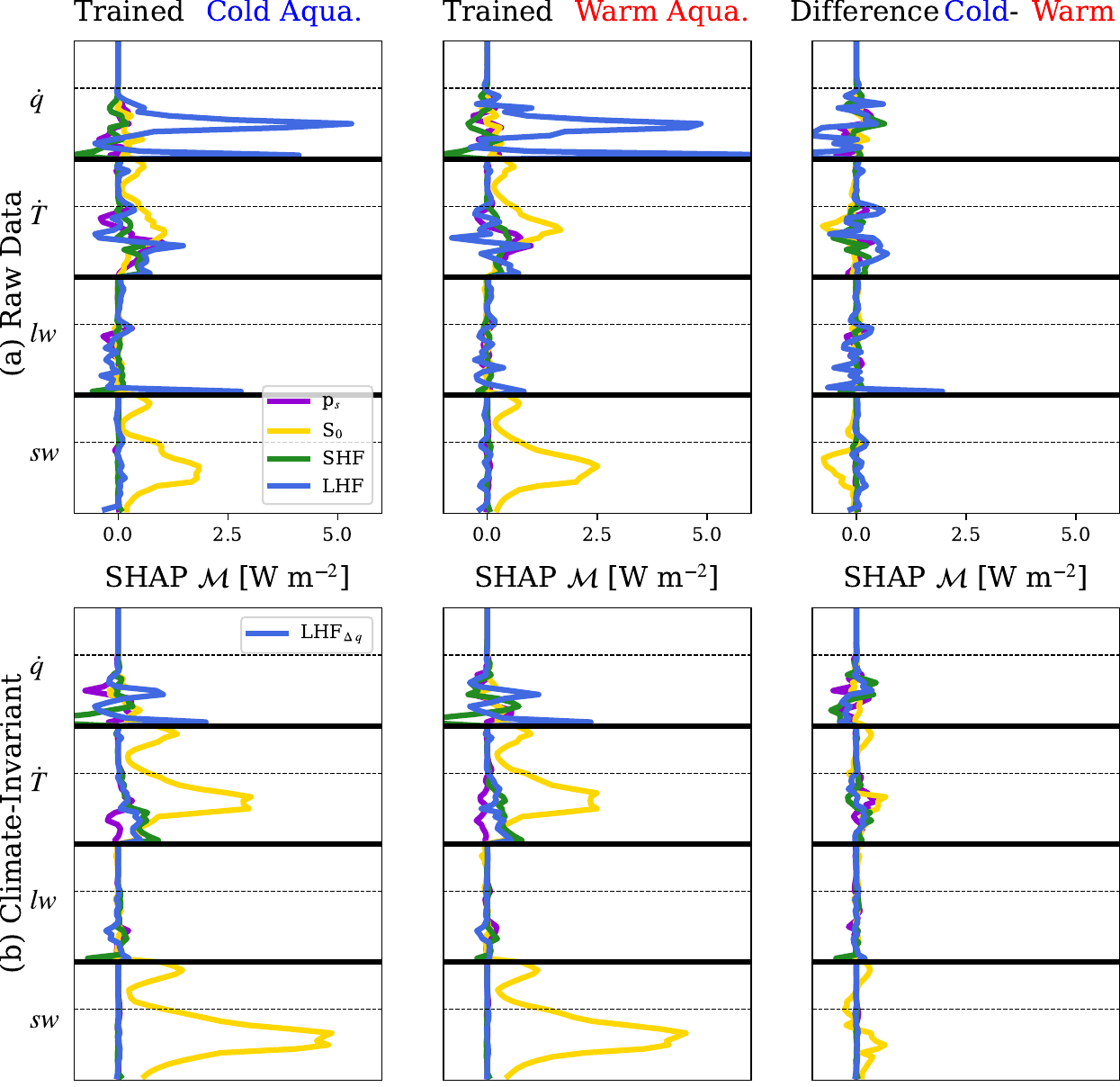}}
 \caption{\textbf{Same as Fig~S14, but for the four scalar inputs used in addition to the temperature and specific humidity inputs.} The scalar inputs are surface pressure ($p_{s} $, purple line), solar insolation ($S_{0} $, yellow line), surface sensible heat flux (SHF, green line), and surface latent heat flux (LHF, blue line). For the climate-invariant mapping (b), LHF is transformed to $\mathrm{LHF}_{\Delta q} $ as described in the ``Theory'' section, which in conjunction with the temperature and humidity transformations, changes the SHAP feature importance matrix for all input variables.}
\end{figure*}

\clearpage
\small
